\pdfoutput=1
\documentclass[10pt,twocolumn,letterpaper]{article}
\usepackage[OT1]{fontenc}
\usepackage[pagenumbers]{cvpr}
\usepackage{times}
\usepackage{amsmath,amssymb}
\usepackage{booktabs}
\usepackage{multirow}
\usepackage{array}
\usepackage{placeins}
\usepackage{float}
\usepackage{xcolor}
\usepackage{colortbl}
\usepackage{graphicx}
\ifdefined\pdfsuppressptexinfo\pdfsuppressptexinfo=-1\relax\fi
\usepackage{pgfplots}
\pgfplotsset{compat=1.17}
\usetikzlibrary{arrows.meta,positioning,calc}
\usepackage{microtype}
\definecolor{linkblue}{rgb}{0.21,0.49,0.74}
\usepackage[breaklinks,colorlinks,allcolors=linkblue]{hyperref}
\hypersetup{
  pdftitle={GWMDriveVLA: A Comprehensive Benchmark for Driving Vision--Language--Action Models},
  pdfauthor={Pinhan Fu, Xianda Guo, Qian Zhou, Ruijie Luo, Jian Liang, Yanlun Peng, Mang Ye},
  pdfsubject={Preprint},
  bookmarksnumbered=true
}
\usepackage[capitalize,nameinlink]{cleveref}
\crefname{section}{Sec.}{Secs.}
\Crefname{section}{Section}{Sections}
\crefname{table}{Tab.}{Tabs.}
\Crefname{table}{Table}{Tables}
\crefname{figure}{Fig.}{Figs.}
\Crefname{figure}{Figure}{Figures}

\newcommand{\ci}[2]{\,{\scriptsize[$#1$,\,$#2$]}}
\newcommand{\vci}[3]{\begin{tabular}[c]{@{}c@{}}$#1$\\[-2pt]{\tiny[$#2$,\,$#3$]}\end{tabular}}
\newcommand{\tname}[1]{\texttt{#1}}
\newcommand{\fw}{GWMDriveVLA}
\newcommand{\dbplus}{\mbox{DriveBase+}}
\newcommand{\sig}{$^\ast$}
\newcommand{\hl}{\rowcolor{gray!12}}
\newcommand{\best}[1]{{\boldmath\bfseries #1}}
\newcommand{\second}[1]{\underline{#1}}
\newcommand{\yes}{$\checkmark$}
\newcommand{\no}{$\times$}
\newcommand{\dn}[1]{\,{\scriptsize\color{red!70!black}$#1$}}
\newcommand{\up}[1]{\,{\scriptsize\color{green!45!black}$#1$}}
\setlist{nosep,leftmargin=*}

\title{\fw{}: A Comprehensive Benchmark for Driving\\Vision--Language--Action Models}
\author{
Pinhan Fu$^{1,2,}$\thanks{Equal contribution\quad$^{\dagger}$Project leader\quad$^{\ddagger}$Corresponding author}, Xianda Guo$^{1,*,\dagger}$, Qian Zhou$^{1,2,*}$, Ruijie Luo$^{1,2}$, Jian Liang$^{1,2}$, Yanlun Peng$^{2,\ddagger}$, Mang Ye$^{1,\ddagger}$\\[0.3em]
$^{1}$Wuhan University \quad $^{2}$Great Wall Motor
}
\date{}

\AddToHook{cmd/@maketitle/after}{
  \begin{center}
    \captionsetup{type=figure}
    \includegraphics[width=\textwidth,pagebox=cropbox]{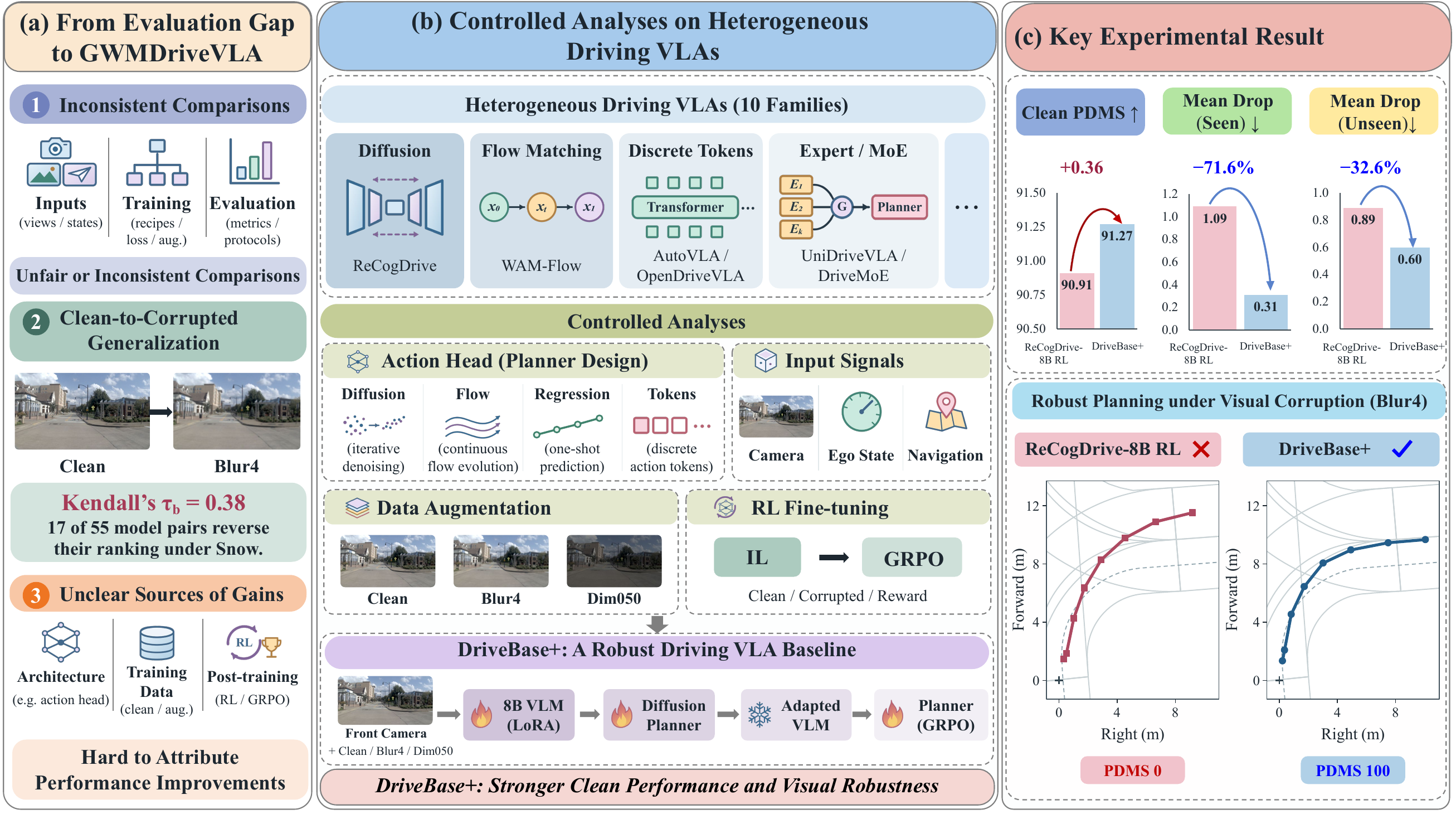}
    \captionof{figure}{\textbf{Overview of \fw{} and \dbplus{}.} (a) Evaluation gaps motivate unified comparison and corruption testing. (b) Controlled analyses of heterogeneous driving VLAs guide \dbplus{}. (c) \dbplus{} against ReCogDrive-8B RL on NAVSIM navtest: clean PDMS, mean drops under seen and unseen corruptions, and one planning example under \tname{blur4}.}
    \label{fig:teaser}
  \end{center}
}

\begin{document}
\maketitle

\begin{abstract}
  Vision--language--action (VLA) models have emerged as a promising paradigm for autonomous driving, bringing the world knowledge and reasoning of large vision--language models to safety-critical trajectory planning, yet their reliability when cameras blur, darken, or compress remains poorly understood. It is unclear which design choices yield both accuracy and robustness, as methods differ in inputs, training recipes, and evaluation code, while clean benchmark scores reveal little about performance under corruption. We introduce \fw{}, a unified codebase and benchmark for driving VLAs that natively runs ten VLA families and a world-action reference on NAVSIM, nuScenes, and Bench2Drive, reproduces their reported results, and evaluates all methods with a common image-corruption layer and protocol. Our analysis reveals that ground-truth selection of the best sampled trajectory inflates PDMS by 0.90--1.34 points, exceeding many method gaps; clean accuracy does not predict robustness; and RL improves clean scores but has method-dependent effects on robustness. These findings motivate \dbplus{}, a strong, robust baseline that adapts an 8B VLM with corruption augmentation before standard GRPO. On NAVSIM navtest, \dbplus{} ranks first among all compared methods on clean inputs (91.27 PDMS) and under every primary corruption (90.78 mean PDMS), while halving the degradation of its 2B predecessor. Code is available at \url{https://github.com/GWM-Drive-Research/GWMDriveVLA}.
\end{abstract}

\section{Introduction}
\label{sec:intro}
Autonomous driving requires planners that act safely in an open world. Vision--language--action (VLA) models have emerged as a promising paradigm toward this goal: built on large vision--language models (VLMs), they bring world knowledge and reasoning to trajectory planning and promise to generalize beyond the scenes of their training logs. Recent systems explore diffusion planners, action tokens, flow matching and expert-based architectures, trained with imitation learning (IL) alone or with an additional reinforcement-learning (RL) stage~\cite{recogdrive,autovla,wamflow,opendrivevla,unidrivevla,drivemoe,minddrive}, and progress on public benchmarks has been rapid.

Whether this progress carries over to the road depends on robustness and on measurement. Deployed cameras face motion blur, low light, compression and bad weather, so a planner must stay reliable when its images degrade, not only on the clean frames of a benchmark; and the five best prior methods on NAVSIM lie within 0.83 PDMS of each other, so which design is better depends on the details of training and evaluation. It therefore remains unclear which design choices make a driving VLA both accurate and robust.

Answering this is hard because of three gaps, summarized in \cref{fig:teaser}(a). \emph{Inconsistent comparisons:} methods differ in their inputs, such as camera views and ego states, in their training recipes and in their evaluation metrics and protocols, and some report the best of several samples chosen with ground truth, which can add more than the difference between two models~\cite{autovla,drivevlaw0}. \emph{Clean-to-corrupted generalization:} a clean score gives an incomplete picture of behavior under degraded camera inputs. Corruption benchmarks expose this sensitivity~\cite{que2026drivedegrade,zhang2026bench2driverobust} but test released models; in our comparison, the Kendall $\tau_b$ between the clean and the snow ranking is only 0.38, and 17 of 55 model pairs reverse their order. \emph{Unclear sources of gains:} a final score mixes the architecture, such as the action head, the training data and post-training such as RL, so an improvement is hard to attribute to any of them.

\begin{figure*}[t]
  \centering
  \includegraphics[width=0.93\textwidth,pagebox=cropbox]{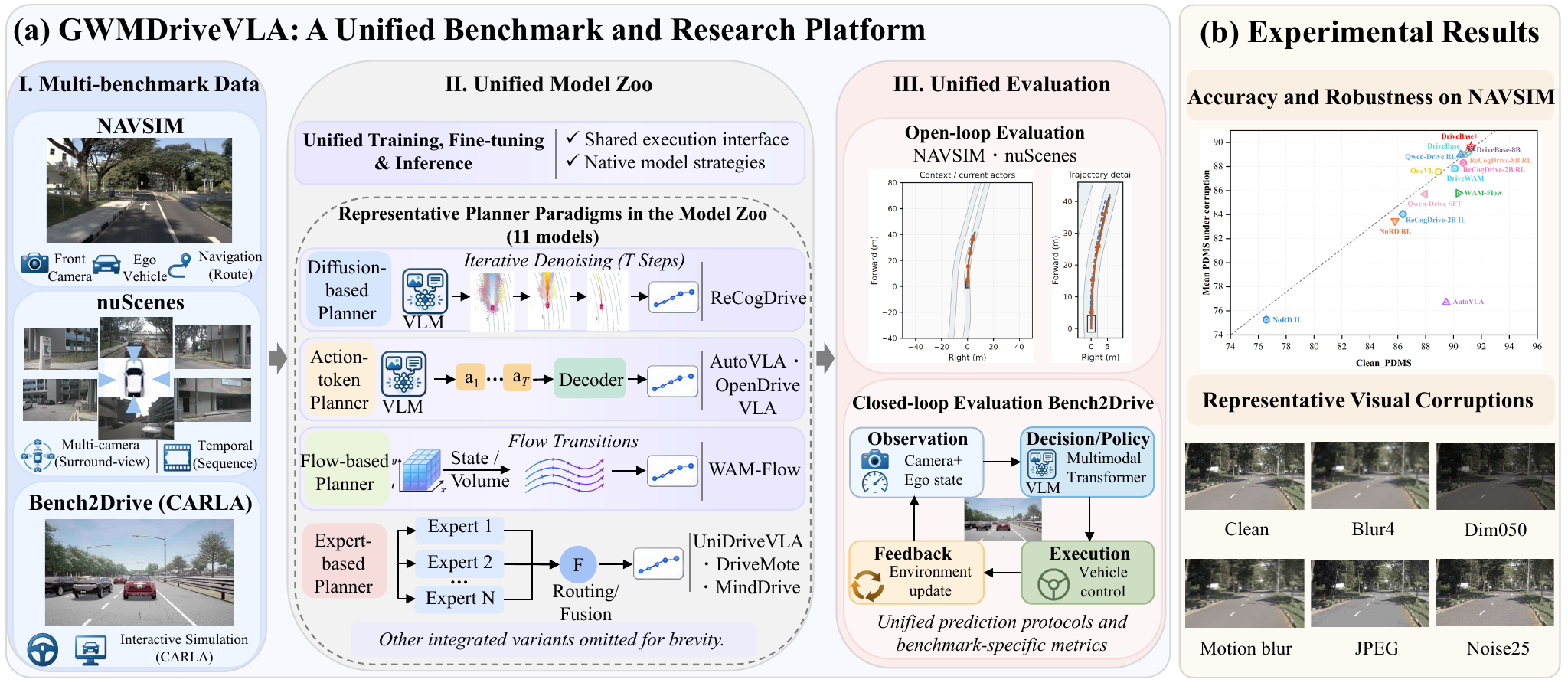}
  \caption{\textbf{\fw{}.} (a) Multi-benchmark data, unified model zoo, and open- and closed-loop evaluation. (b) Clean PDMS against mean PDMS over the seven corruptions not used for augmentation on NAVSIM navtest, and the clean input with the five primary corruptions.}
  \label{fig:framework}
  \vspace{-6pt}
\end{figure*}

To address these gaps, we introduce \fw{}, a unified codebase and benchmark for driving VLAs, shown in \cref{fig:framework}. It runs ten VLA families and a non-VLA world-action reference on NAVSIM, nuScenes and Bench2Drive with their native networks, inputs and preprocessing, connects them to fixed evaluators, a common image-corruption layer and paired statistics, and reproduces their reported results, so that all of them are trained, corrupted and scored under one protocol; adding a method takes one adapter and one configuration file.

Revisiting and deconstructing driving VLAs with \fw{}, we first find that selecting the best of six sampled trajectories with ground truth raises PDMS by 0.90 to 1.34 points, more than the gap between the five best prior methods, much of it because nearly identical plans can receive very different scores, and that clean accuracy does not predict robustness: rankings change under corruption, and released models lose up to 7.18 PDMS on average under blur and dimming. Varying one factor at a time on a shared diffusion planner, as \cref{fig:teaser}(b) outlines, we find that the action head matters most among the tested factors, that IL-stage augmentation cuts the drop under the augmented corruptions from 2.20 to 0.38 PDMS at the same clean score and transfers across planners and benchmarks, and that RL raises the clean score and lowers extended comfort while its effect on robustness depends on the method. Applied to an augmented model, standard GRPO even erodes part of its robustness, suggesting that robustness should be learned before RL.

These analyses lead to \dbplus{}, a strong and robust baseline that keeps the diffusion planner and the IL-stage augmentation, moves to an 8B VLM, and adapts this VLM jointly with the planner under the same augmentation before standard GRPO, so that robustness is learned in the representation. \cref{fig:teaser}(c) summarizes the result. On NAVSIM navtest, \dbplus{} ranks first among all compared methods on clean inputs and under every primary corruption; against ReCogDrive-8B RL~\cite{recogdrive}, the best prior method on clean inputs, it raises clean PDMS from 90.91 to 91.27 and cuts the mean drop under seen and unseen corruptions by 71.6\% and 32.6\%, at no additional inference cost. Matched controls show that adapting the VLM helps only together with augmentation, and that broader augmentation extends the benefit to more corruptions. Our contributions are threefold:
\begin{itemize}[leftmargin=*,itemsep=1pt,topsep=2pt]
  \item We introduce \fw{}, a unified codebase and benchmark that runs ten VLA families and a world-action reference natively on three benchmarks and evaluates them under one protocol with a common corruption layer.
  \item With \fw{}, we revisit and deconstruct driving VLAs and report key findings on evaluation inflation, clean-to-corrupted generalization, action heads, IL-stage augmentation and RL fine-tuning.
  \item We present \dbplus{}, a strong and robust baseline that ranks first among all compared methods on NAVSIM, on clean inputs and under every primary corruption, and halves the degradation of its 2B predecessor.
\end{itemize}

\section{Related Work}
\label{sec:related}

\subsection{Driving Vision--Language--Action Models}
End-to-end planners~\cite{hu2023uniad,jiang2023vad,chitta2023transfuser,liao2025diffusiondrive} map sensors to trajectories. Following robot VLAs~\cite{brohan2023rt2,kim2024openvla}, recent models bring VLMs into driving~\cite{sima2024drivelm,tian2024drivevlm,shao2024lmdrive,hwang2024emma,fu2025orion,renz2025simlingo}, and Appendix~\ref{sec:timeline} gives a timeline of recent driving VLAs.

\paragraph{VLAs Trained by Imitation.}
OpenDriveVLA~\cite{opendrivevla} decodes trajectories autoregressively from aligned 2D and 3D visual tokens, UniDriveVLA~\cite{unidrivevla} decouples understanding, perception and action into Mixture-of-Transformers experts, DriveMoE~\cite{drivemoe} adds scene- and skill-specialized experts to a $\pi_0$-style~\cite{black2024pi0} VLA, and OneVL~\cite{onevl} compresses reasoning into latent tokens. These methods change the action head, the inputs and the training stages at once; our controlled study separates them.

\paragraph{VLAs with RL Fine-tuning.}
ReCogDrive~\cite{recogdrive} fine-tunes a diffusion planner with GRPO~\cite{shao2024deepseekmath} over denoising chains~\cite{ren2025dppo}, AutoVLA~\cite{autovla} applies GRPO to action tokens and adaptive reasoning, WAM-Flow~\cite{wamflow} aligns a discrete flow-matching planner with simulator-guided GRPO, and MindDrive~\cite{minddrive} runs online RL over language decisions in CARLA. Later works shape the reward with a world model~\cite{sheng2026explorevla}, adapt GRPO to small policies~\cite{nord}, distill geometric priors into latent reasoning~\cite{lastvla} or release separate IL and RL checkpoints~\cite{qwendrive}. We compare how RL changes robustness in these families under one corruption protocol.

\subsection{Evaluation of Driving Models}
\paragraph{Evaluation Protocols.}
Open-loop evaluation on nuScenes has been questioned: an MLP on the ego status matches perception-based planners~\cite{zhai2023rethinking}, which barely degrade when their cameras are blanked~\cite{li2024egostatus}; NAVSIM~\cite{dauner2024navsim} and Bench2Drive~\cite{jia2024bench2drive} answer with simulation-based scores. Reporting practices and seed variance inflate results in deep RL~\cite{henderson2018deep}, metric learning~\cite{musgrave2020metric} and machine learning benchmarks~\cite{bouthillier2021accounting}, and recent driving VLAs also report the best of several samples chosen with ground truth~\cite{autovla,drivevlaw0}, an inflation we quantify.

\paragraph{Corruption Robustness.}
ImageNet-C~\cite{hendrycks2019benchmarking} standardized evaluation under common corruptions, later extended to detection~\cite{michaelis2019benchmarking}, 3D and BEV perception~\cite{dong2023benchmarking,kong2023robo3d,xie2025benchmarking}, VLM question answering~\cite{xie2025drivebench}, end-to-end planning~\cite{jiang2024robuste2e,que2026drivedegrade}, VLM trajectory prediction~\cite{liao2025robodrivevlm} and closed-loop driving~\cite{zhang2026bench2driverobust}. Closest to our work, DriveDegrade~\cite{que2026drivedegrade} measures the degradation tolerance of camera-only planners and Bench2Drive-Robust~\cite{zhang2026bench2driverobust} benchmarks closed-loop driving under deployment perturbations, both on released models. Training-time augmentation is the standard remedy in image classification~\cite{hendrycks2020augmix}, but its interplay with VLM representations and RL in driving VLAs has not been studied; \fw{} trains and evaluates driving VLAs under one protocol to study how the training recipe changes robustness.

\subsection{Codebase}
Unified codebases such as OpenStereo~\cite{guo2023openstereo}, OpenGait~\cite{fan2023opengait}, MMDetection~\cite{chen2019mmdetection} and Detectron2~\cite{wu2019detectron2} re-implement many networks under one protocol and revisit their design choices. For driving, NAVSIM~\cite{dauner2024navsim} and Bench2Drive~\cite{jia2024bench2drive} provide evaluators while each VLA ships its own pipeline; \fw{} connects these native implementations through shared interfaces.

\section{\fw{}}
\label{sec:drivescope}
\fw{} is a unified codebase and benchmark for driving VLAs. It trains and evaluates them in one standardized pipeline, so that methods can be compared fairly on clean and degraded inputs; \cref{fig:framework} gives an overview.

\subsection{Design Principles}
\label{sec:framework}
\paragraph{Modular Design.}
\fw{} unifies the interface rather than the models: a benchmark reader turns raw data into a canonical sample, every method reads it through an adapter and writes a standard prediction record, and each method keeps its native network, dependencies and, where needed, training backend. Adding a method requires an adapter and a configuration file, and every run writes a manifest with its configuration, source commits, checkpoint and file hashes, linking each reported number to its run.

\paragraph{Fair Comparison.}
Three rules keep comparisons fair. An evaluator chosen by benchmark and version reads the ground truth on its own, so no prediction can access it; corruptions act on the raw images before any method-specific preprocessing, so every method receives the same degraded input; and every method is scored with one sampled trajectory and compared with paired statistics on the same scenes (\cref{sec:data}).

\subsection{Supported Methods and Benchmarks}
\label{sec:supported}
\paragraph{Planner Paradigms and Benchmarks.}
\fw{} supports diffusion planners such as ReCogDrive, discrete action tokens such as AutoVLA and OpenDriveVLA, discrete flow matching such as WAM-Flow, and expert-based planners such as UniDriveVLA, DriveMoE and MindDrive, each reading the cameras, history frames, ego state and navigation command it was designed for, with 455 to 3{,}154 lines of adapter code, as \cref{tab:repro} lists. It covers open-loop evaluation on NAVSIM and nuScenes and closed-loop evaluation on Bench2Drive, and its corruption layer provides the nine corruption families of \cref{sec:data} at adjustable strengths.

\paragraph{Integrated Methods.}
\fw{} integrates both training and inference for ReCogDrive, AutoVLA, OpenDriveVLA, UniDriveVLA, DriveMoE and MindDrive, and we train and evaluate all six. WAM-Flow, NoRD~\cite{nord}, Qwen-Drive~\cite{qwendrive}, OneVL~\cite{onevl} and DriveWAM~\cite{drivewam} are integrated for inference and evaluated with their released weights through the same interface. The released NAVSIM branch of DriveWAM calls no VLM, so we report it as a non-VLA reference. Appendix~\ref{sec:zoo} reports model size, memory and latency measurements.

\section{Revisit Driving VLAs}
\label{sec:revisit}

\subsection{Datasets and Evaluation Metrics}
\label{sec:data}

\paragraph{NAVSIM}
~\cite{dauner2024navsim} is our main testbed. It scores 4-second trajectories with the non-reactive PDM Score~\cite{dauner2023parting}, or PDMS. We train on the 92{,}944 tokens of navtrain and evaluate on the full navtest split, 12{,}146 tokens from 136 logs, with the v1.1 evaluator. A separate development split of 4{,}000 tokens from 119 held-out logs serves recipe selection.

\paragraph{nuScenes and Bench2Drive.}
nuScenes~\cite{caesar2020nuscenes} evaluates open-loop planning by L2 error and collision rate on 6{,}019 validation samples under the ST-P3 protocol~\cite{hu2022stp3}, and Bench2Drive~\cite{jia2024bench2drive} evaluates closed-loop driving on 220 routes in CARLA 0.9.15~\cite{dosovitskiy2017carla} by driving score (DS), success rate (SR) and multi-ability success.

\paragraph{Corruptions.}
The five primary corruptions cover the main kinds of camera degradation, following ImageNet-C~\cite{hendrycks2019benchmarking} and the families that DriveDegrade~\cite{que2026drivedegrade} finds most damaging to planning; \cref{fig:framework}(b) shows them. We train on two frequent road conditions at moderate strength, the \emph{seen} corruptions Gaussian blur with radius 4, \tname{blur4}, and brightness scaled by 0.5, \tname{dim050}. The three \emph{training-unseen} corruptions, never used for gradient updates, test how far robustness transfers: horizontal motion blur with a 15-pixel kernel is a close relative of the seen blur, whereas JPEG compression at quality 10 and Gaussian noise with standard deviation 25/255, \tname{noise25}, differ in kind. Fog, rain, snow and overexposure form an extended unseen set, reported in Appendix~\ref{sec:weather}. Following \cref{sec:framework}, each corruption is applied to every camera image a model reads, including all 12 frames of AutoVLA.

\paragraph{Evaluation Metrics.}
PDMS is reported on a 0--100 scale. The drop of a model $f$ under corruption $k$ is the paired difference $D_k(f)=\mathrm{PDMS}_{\text{clean}}(f)-\mathrm{PDMS}_k(f)$, and lower is better; $\bar D^{\mathcal S}$ and $\bar D^{\mathcal T}$ average it over the seen set $\mathcal S=\{\tname{blur4},\tname{dim050}\}$ and the unseen set $\mathcal T=\{\text{motion},\text{JPEG},\tname{noise25}\}$, and the ten-condition mean averages PDMS over the clean input and all nine corruptions. Differences are paired per token, with 95\% intervals from a bootstrap~\cite{efron1979bootstrap} that resamples the 136 navtest logs 5{,}000 times; several IL seeds are first averaged per token. A difference is significant when its interval excludes zero, marked by an asterisk in tables.

\paragraph{Evaluation Protocol.}
Beyond the rules of \cref{sec:framework}, every method is evaluated at its final or released checkpoint with a fixed evaluation seed; main recipes use two or three IL seeds and the final ones a second GRPO seed. Recipes are chosen on the development split, navtest serves only for evaluation, all decision rules were fixed before the results, and an independent second pass recomputes each statistic.

\subsection{Evaluation of Prior Work}
\label{sec:prior}

\begin{table*}[t]
  \centering
  \caption{\textbf{Reproduction across three driving benchmarks.} Paper/Released/Ours: published scores, our evaluation of released weights, and training in \fw{}; bold: best per row; LOC: adapter code lines.}
  \label{tab:repro}
  \renewcommand{\arraystretch}{0.88}
  \setlength{\tabcolsep}{4.2pt}
  \scalebox{0.9}{
    \begin{tabular}{@{}lllll|ccc|r@{}}
      \toprule
      \multirow{2}{*}{Method} & \multirow{2}{*}{Benchmark} & \multirow{2}{*}{Planner} & \multirow{2}{*}{Visual input} & \multirow{2}{*}{Training} & \multicolumn{3}{c|}{Score} & \multirow{2}{*}{LOC} \\
      \cmidrule(lr){6-8}
      & & & & & Paper & Released & Ours & \\
      \midrule
      ReCogDrive-2B~\cite{recogdrive} & \multirow{4}{*}{\shortstack{NAVSIM\\PDMS$\uparrow$}} & diffusion & front, 1 frame & IL + GRPO & 90.80 & 90.71 & \best{91.03} & 567 \\
      ReCogDrive-8B~\cite{recogdrive} &  & diffusion & front, 1 frame & GRPO & 90.40 & -- & \best{90.91} & 567 \\
      AutoVLA~\cite{autovla} &  & discrete tokens & 3 cams $\times$ 4 frames & SFT only & 89.11 & \best{89.57} & 77.95 & 1{,}173 \\
      WAM-Flow~\cite{wamflow} &  & discrete flow & front, 1 frame & -- & 90.30 & \best{90.37} & -- & 455 \\
      \midrule
      OpenDriveVLA-0.5B~\cite{opendrivevla} & \multirow{2}{*}{\shortstack{nuScenes\\L2 (m)$\downarrow$}} & AR tokens & 6 cams + memory & fine-tune & 0.35 & 0.35 & \best{0.34} & 904 \\
      UniDriveVLA~\cite{unidrivevla} &  & MoT experts & 6 cams + memory & last 2 stages & 0.54 & -- & \best{0.53} & 3{,}154 \\
      \midrule
      UniDriveVLA~\cite{unidrivevla} & \multirow{3}{*}{\shortstack{Bench2Drive\\DS$\uparrow$}} & MoT experts & 6 cams + memory & -- & \best{78.37} & 76.99 & -- & 3{,}154 \\
      DriveMoE~\cite{drivemoe} &  & MoE & 6 cams + 1 history & 2 stages & \best{74.22} & 72.44 & 73.84 & 1{,}677 \\
      MindDrive~\cite{minddrive} &  & lang.\ + action expert & 6 cams + memory & IL, 3 stages & \best{78.04} & -- & 74.97 & 1{,}399 \\
      \bottomrule
    \end{tabular}
  }
\end{table*}

\paragraph{Reproduction.}
\label{sec:repro}
As \cref{tab:repro} shows, \fw{} reproduces the reported results on all three benchmarks. Trained in our framework, ReCogDrive-2B reaches 91.03 PDMS against the reported 90.80, the nuScenes models stay within 0.01\,m of their L2 errors and the Bench2Drive models within 0.88 DS, and the released weights of all NAVSIM methods reproduce within 0.46 PDMS. Only AutoVLA is not reproduced through training: its supervised checkpoint is not released and our supervised stage reaches 77.95 PDMS, so we do not run its RL stage. Table~\ref{tab:repro_full} and Appendix~\ref{sec:sota_app} give the sub-metrics and published NAVSIM scores.

\paragraph{Evaluation Inflation.}
\label{sec:inflation}
Recent NAVSIM papers also report the best of $N$ samples selected with ground truth~\cite{autovla,drivevlaw0}. For ReCogDrive, WAM-Flow and DriveBase, our 2B baseline of \cref{sec:deconstruct}, the best of six adds 0.90--1.34 PDMS against noise floors of 0.12--0.18 across evaluation seeds. For \dbplus{}, our final model of \cref{sec:drivebase}, it adds 1.16 $[1.04,1.29]$ over the mean of its six samples, while sampling seeds differ by at most 0.08 $[0.05,0.20]$. Much of this gain comes from discontinuities of PDMS: for IL seed 1, the 1.18\% of scenes whose best and worst plans stay within 1\,m yet differ by at least 50 PDMS provide 42\% of it, as Appendix~\ref{sec:inflationapp} details. The inflation exceeds the 0.83 PDMS between the best and the fifth-best prior method in \cref{tab:main}, reaches 3.00 PDMS as reported for AutoVLA, and uses test-set information~\cite{driveteach}, so comparisons use one sample.

\begin{table*}[t]
  \centering
  \caption{\textbf{Robustness on NAVSIM navtest (PDMS).} Base-IL/Aug-IL: clean/augmented IL initialization; 8B rows: three denoising steps; Scene- and Pair-GRPO: one IL seed, other rows of ours: two. Mean: average of the five corruption scores. Mean drop: clean minus corrupted PDMS, averaged over seen or unseen corruptions. Bold/underline: best/second.}
  \label{tab:main}
  \renewcommand{\arraystretch}{0.88}
  \setlength{\tabcolsep}{3.8pt}
  \begin{tabular}{@{}c|l|c|cc|ccc|c|cc@{}}
    \toprule
    \multirow{2}{*}{Group} & \multirow{2}{*}{Method} & \multirow{2}{*}{Clean$\uparrow$} & \multicolumn{2}{c|}{Seen$\uparrow$} & \multicolumn{3}{c|}{Unseen$\uparrow$} & \multirow{2}{*}{Mean$\uparrow$} & \multicolumn{2}{c}{Mean drop$\downarrow$} \\
    & & & Blur & Dim & Motion & JPEG & Noise & & Seen & Unseen \\
    \midrule
    \multirow{10}{*}{\shortstack{Prior\\VLAs}} & ReCogDrive-2B (RL)~\cite{recogdrive} & 90.71 & 87.34 & 90.18 & 88.02 & 88.81 & 89.44 & 88.76 & 1.94 & 1.95 \\
    & ReCogDrive-2B (IL) & 86.36 & 82.87 & 85.81 & 84.03 & 84.25 & 85.12 & 84.42 & 2.02 & 1.90 \\
    & ReCogDrive-8B (RL) & 90.91 & 89.24 & 90.40 & 89.79 & 89.77 & 90.47 & 89.93 & 1.09 & 0.89 \\
    & WAM-Flow~\cite{wamflow} & 90.35 & 88.22 & 87.98 & 88.60 & 88.29 & 89.35 & 88.49 & 2.25 & 1.60 \\
    & AutoVLA~\cite{autovla} & 89.48 & 84.20 & 80.41 & 85.15 & 82.87 & 86.99 & 83.92 & 7.18 & 4.48 \\
    & Qwen-Drive (SFT)~\cite{qwendrive} & 87.95 & 84.20 & 86.73 & 84.72 & 82.24 & 86.73 & 84.92 & 2.49 & 3.39 \\
    & Qwen-Drive (RL) & 90.50 & 88.21 & 89.91 & 88.74 & 86.54 & 89.68 & 88.62 & 1.44 & 2.18 \\
    & NoRD (IL)~\cite{nord} & 76.57 & 75.12 & 75.47 & 75.87 & 76.35 & 76.60 & 75.88 & 1.27 & \best{0.30} \\
    & NoRD (RL) & 85.80 & 83.77 & 84.10 & 84.49 & 85.18 & 85.44 & 84.59 & 1.87 & 0.77 \\
    & OneVL~\cite{onevl} & 88.93 & 86.98 & 88.18 & 88.49 & 88.05 & 88.09 & 87.96 & 1.35 & 0.72 \\
    \midrule
    Non-VLA & DriveWAM~\cite{drivewam} & 90.08 & 87.25 & 89.24 & 89.33 & 89.60 & 88.85 & 88.86 & 1.84 & 0.82 \\
    \midrule
    \multirow{4}{*}{\shortstack{Base-IL}} & GRPO & 91.02 & 86.60 & 90.60 & 88.20 & 89.69 & 89.88 & 88.99 & 2.42 & 1.77 \\
    & Scene-GRPO & 90.98 & 89.39 & 90.41 & 89.88 & 90.24 & 90.08 & 90.00 & 1.08 & 0.91 \\
    & Pair-GRPO & 90.88 & 89.33 & 90.39 & 89.82 & 90.16 & 89.89 & 89.92 & 1.02 & 0.93 \\
    & RA-GRPO & 90.51 & 90.20 & 90.44 & 89.73 & 89.40 & 89.63 & 89.88 & \best{0.19} & 0.92 \\
    \midrule
    \multirow{3}{*}{\shortstack{Aug-IL}} & IL only & 86.82 & 86.29 & 86.58 & 86.10 & 85.84 & 86.09 & 86.18 & 0.38 & 0.80 \\
    & DriveBase (GRPO) & 91.05 & 89.65 & 90.78 & 89.80 & 89.91 & 90.23 & 90.07 & 0.84 & 1.07 \\
    & RA-GRPO & 90.89 & 90.62 & 90.78 & 90.14 & 89.63 & 89.81 & 90.20 & \best{0.19} & 1.03 \\
    \midrule
    & DriveBase-8B & \second{91.24} & 90.62 & \second{90.82} & 90.74 & \second{90.35} & 90.59 & 90.62 & 0.52 & 0.68 \\
    & Equal-update control & 91.23 & \second{90.70} & 90.80 & \second{90.75} & 90.33 & \second{90.62} & \second{90.64} & 0.48 & 0.67 \\
    \hl \multirow{-3}{*}{\shortstack{Aug-IL\\8B}} & \textbf{DriveBase+ (Ours)} & \best{91.27} & \best{90.88} & \best{91.03} & \best{90.87} & \best{90.43} & \best{90.72} & \best{90.78} & \second{0.31} & \second{0.60} \\
    \bottomrule
  \end{tabular}
  \vspace{-2pt}
\end{table*}

\paragraph{Robustness of Prior Methods.}
In \cref{tab:main}, the released ReCogDrive, WAM-Flow and AutoVLA models lose 1.94 to 7.18 PDMS on average under seen corruptions and 1.60 to 4.48 under unseen ones, AutoVLA most. Corruptions also reorder the methods: the Kendall $\tau_b$ between the clean and the snow ranking is only 0.38, as Appendix~\ref{sec:validity} details, so a clean score does not predict the ranking under corruption.

\subsection{Deconstructing Driving VLAs}
\label{sec:deconstruct}
\label{sec:ablation}
\label{sec:necessity}
A reported gain mixes data, inputs, action head and post-training, so we change one factor at a time and pair all comparisons on the same scenes: augmentation in \cref{sec:aug}, the action head and inputs in \cref{sec:head}, and RL in \cref{sec:rl}.

\paragraph{Implementation Details.}
We build on the ReCogDrive-2B planner, the strongest NAVSIM method with released IL and RL weights, whose frozen VLM features can be cached for affordable controlled training. Two IL recipes train for 128 epochs, \textbf{Base-IL} on clean and \textbf{Aug-IL} on corrupted inputs; Base-IL has three IL seeds for the ablations, and RL runs start from two IL seeds unless noted. On top of either, we run standard GRPO and three robustness-aware variants with 14{,}530 updates and groups of 8, and Aug-IL with standard GRPO is our 2B baseline, \textbf{DriveBase}. As in the official code, GRPO includes a behavior-cloning term with coefficient 0.1; Appendix~\ref{sec:impl} gives all settings.

\subsubsection{Data Augmentation}
\label{sec:aug}

Aug-IL trains the IL planner on clean, \tname{blur4} and \tname{dim050} inputs in equal proportions. Tab.~\ref{tab:aug} in Appendix~\ref{sec:impl} shows that it cuts the seen drop from 2.20 to 0.38 PDMS at the same clean score. For IL seed 1, it also cuts the unseen drop by a significant 0.61 PDMS, almost entirely through motion blur, so augmentation generalizes within the blur family rather than to arbitrary degradations. Retraining with IL seed 1 shows that \tname{dim050} alone leaves the drops under \tname{blur4} and motion blur 1.96 and 1.08 PDMS higher, while \tname{blur4} alone keeps the seen robustness but costs 0.41 clean PDMS. Augmenting the IL stage thus buys robustness to the augmented families at no clean cost.

\subsubsection{Action Head and Inputs}
\label{sec:head}

\begin{table}[t]
  \centering
  \caption{\textbf{Ablation of the action head, inputs and inference.} Each row changes one factor of Base-IL; means over three IL seeds. Coloured: change from Base-IL before rounding, green better, red worse. Drop red.: reduction of the clean-to-\tname{blur4} drop. $^\ast$: 95\% interval excludes 0. Params in M; Lat. in ms; --: not measured.}
  \label{tab:ablation}
  \scriptsize
  \setlength{\tabcolsep}{1.2pt}
  \begin{tabular}{@{}llcccc@{}}
    \toprule
    Factor & Variant & Params & Lat.$\downarrow$ & Clean$\uparrow$ & Drop red.$\uparrow$ \\
    \midrule
    \hl Reference & Base-IL: diffusion, 5 steps & 34 & 80 & \best{86.827} & -- \\
    \midrule
    \multirow{3}{*}{Head} & flow matching & 34 & 78 & 86.524\dn{-0.303^\ast} & \textcolor{red!70!black}{$-0.134$} \\
    & MLP regression & 32 & $<$1 & 80.942\dn{-5.885^\ast} & \textcolor{red!70!black}{$-7.719^\ast$} \\
    & discrete vocabulary & 36 & $<$1 & 73.361\dn{-13.466^\ast} & \textcolor{red!70!black}{$-23.282^\ast$} \\
    \midrule
    \multirow{2}{*}{Inputs} & no navigation command & 34 & -- & 86.623\dn{-0.204} & \textcolor{green!45!black}{$+0.200$} \\
    & no ego state & 34 & -- & 86.198\dn{-0.629^\ast} & \textcolor{red!70!black}{$-0.609^\ast$} \\
    \midrule
    Inference & 3 denoising steps & 34 & \best{45} & \best{86.834}\up{+0.007} & \textcolor{green!45!black}{$+0.049$} \\
    \bottomrule
  \end{tabular}
  \vspace{-6pt}
\end{table}

All ablations start from Base-IL, which reaches 86.827 PDMS over three IL seeds against 86.36 for the released IL weights; each variant changes one factor, with 36 models trained and 111 evaluations. As \cref{tab:ablation} shows, no variant significantly improves on Base-IL, and the action head has the largest effect on both clean performance and robustness: a regression head and a discrete action vocabulary lose 5.885 and 13.466 PDMS, and their drop under blur grows by 7.719 and 23.282 PDMS. Flow matching~\cite{lipman2023flow} comes close to diffusion~\cite{ho2020ddpm} at 0.303 PDMS lower, and three denoising steps change clean PDMS by $+0.007$ and reduce the blur drop by $0.049$, neither significantly, while cutting planner latency from 80 to 45\,ms. Separately, the 8B model after GRPO changes clean PDMS by only $-0.17$ versus 2B but significantly lowers the blur drop by 1.89 PDMS, comparable to augmentation; this exploratory comparison changes both VLM and IL initialization.

\paragraph{Inputs.}
\label{sec:inputs}
In \cref{tab:ablation}, removing the ego state costs 0.629 PDMS and raises the drop under blur by 0.609, whereas removing the navigation command costs a non-significant 0.204, so the ego state is the only input whose removal hurts both the clean score and robustness.

\subsubsection{RL Fine-tuning}
\label{sec:rl}
\label{sec:robust}

\paragraph{Where the RL Gain Comes From.}
In the three 2B ReCogDrive pairs of Fig.~\ref{fig:rlgain} in the appendix, RL raises clean PDMS significantly, by 4.14 to 4.84. Writing $\mathrm{PDMS}=100\,G\,V$ with the gate $G=\mathrm{NC}\cdot\mathrm{DAC}$ and $V=(5\,\mathrm{EP}+5\,\mathrm{TTC}+2\,\mathrm{C})/12$ and splitting the per-token change additively, the gain comes mainly from drivable-area compliance and ego progress, and gate changes from zero to positive scores account for 52\% to 57\% of it. With the VLM fixed, the gain reflects planner adaptation on frozen features.

\begin{figure}[t]
  \centering
  \includegraphics[width=\linewidth]{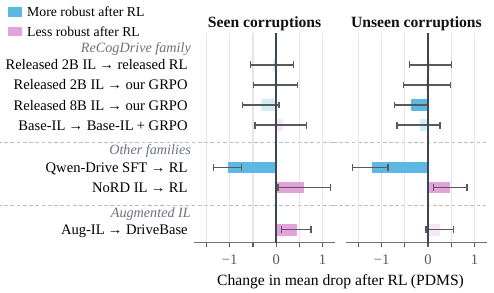}
  \caption{\textbf{Change in robustness from IL to RL.} Change in mean drop, RL minus IL, with 95\% CIs; light bars: CI includes zero.}
  \label{fig:ilrl}
  \vspace{-6pt}
\end{figure}

\paragraph{Robustness after RL.}
As \cref{fig:ilrl} shows, RL does not reduce the drop $D$ in general. Within the ReCogDrive family it leaves robustness almost unchanged: for the four pairs from non-augmented IL models, all intervals but one contain zero. Qwen-Drive and NoRD, with released weights, move in opposite directions: RL lowers the seen and unseen drops of Qwen-Drive by 1.04 and 1.21 PDMS and raises those of NoRD by 0.60 and 0.47, although both gain clean score, so a clean gain from RL does not predict a robustness gain, also under weather, as Appendix~\ref{sec:weather} shows. The visual reliance $R$, the PDMS lost when every camera image is replaced by uniform gray, rises after RL in all seven pairs in step with the drop, with Spearman correlations of 0.83 to 0.94; small drops can also come from ignoring the camera~\cite{li2024egostatus}, as the IL model of NoRD shows in Appendix~\ref{sec:addmetrics}.

\paragraph{Metrics Outside the Reward.}
NAVSIM v2~\cite{cao2025pseudo} scores the same scenes with EPDMS, which adds terms the PDMS reward ignores, such as extended comfort, the consistency between consecutive plans. In all four IL-to-RL pairs we score, RL raises drivable-area compliance but lowers extended comfort, from above 85 to 27.50 and 21.45 in the ReCogDrive family and from 63.69 to 19.67 for Qwen-Drive, so EPDMS falls by 3.79 to 6.08 while PDMS rises; NoRD, whose comfort drops by only 1.85, is the exception, as Appendices~\ref{sec:navsimv2} and~\ref{sec:addmetrics} detail.

\paragraph{RL on an Augmented Model.}
From the already robust Aug-IL, standard GRPO makes robustness worse: over both IL seeds, the seen drop grows significantly by 0.46 PDMS. Under \tname{blur4}, the erosion comes from losses of 0.58 and 0.28 PDMS in time-to-collision and at-fault collisions, partly offset by 0.29 PDMS in progress, so the policy keeps pursuing progress on degraded inputs at the cost of safety margin.

\paragraph{Robustness-aware GRPO.}
Letting corrupted inputs enter RL, by drawing one condition per scene with Scene-GRPO, pairing clean and corrupted rollouts with Pair-GRPO or distilling towards the best clean rollout with RA-GRPO, helps only from Base-IL, where Scene-GRPO alone also keeps the clean score, lowering the seen and unseen drops by 1.11 and 0.67 PDMS; from Aug-IL, it more than doubles the seen drop on the development split, as Appendices~\ref{sec:ragrpo} and~\ref{sec:posthoc} detail. RL thus improves what its reward measures, and robustness is best built before the RL stage.

\section{A Strong Baseline: DriveBase+}
\label{sec:drivebase}
\begin{figure*}[t]
  \centering
  \includegraphics[width=0.94\textwidth]{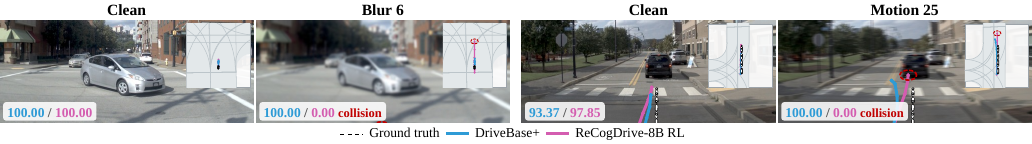}
  \caption{\textbf{\dbplus{} and ReCogDrive-8B RL on two navtest scenes.} Plans projected onto the front camera with a bird's-eye inset; scores give the PDMS of both methods, red ellipses mark plan ends that score zero, and Appendix~\ref{sec:qual} gives the selection rule.}
  \label{fig:m1}
  \vspace{-4pt}
\end{figure*}

The controlled study fixes the design of \dbplus{}: it keeps the diffusion head and the IL-stage augmentation, moves to the 8B VLM, which lowers the drop after RL, and, since RL on frozen features does not build robustness, learns robustness before RL in the representation itself by adapting the VLM under the same augmentation.

\subsection{Pipeline}
\label{sec:pipeline}
\textbf{DriveBase+} extends DriveBase to an 8B VLM with end-to-end adaptation, as in \cref{fig:teaser}(b). From an 8B IL checkpoint trained with the Aug-IL recipe, we jointly update the planner and rank-16 LoRA modules~\cite{hu2022lora} in the vision and language attention projections for three epochs under the same augmentation. We then freeze the adapted representation, extract features through the unmerged LoRA path, and train the planner with standard GRPO for 14{,}530 updates, with the released 8B IL model as behavior-cloning reference. At inference, one front image, ego state and navigation command give a 4-second plan in three steps; with 12.45\,M added LoRA parameters, this takes 583\,ms per frame on one H20 GPU against 598\,ms for ReCogDrive-8B, as Appendix~\ref{sec:zoo} lists. Final results average two IL seeds with evaluation seed 8101, and Appendix~\ref{sec:finaldetails} reports each seed.

Recipe selection on the development split keeps variants within 0.13 clean PDMS of DriveBase that raise the ten-condition mean by more than 0.13 and takes the highest single-sample mean, as Appendix~\ref{sec:posthoc} details. Two controls with the same IL seeds and GRPO stage separate the steps: DriveBase-8B without adaptation, and an equal-update control that continues IL for three epochs on frozen features.

\subsection{Final Performance}
\label{sec:sota}
\label{sec:main}
\begin{figure}[t]
  \centering
  \includegraphics[width=\linewidth]{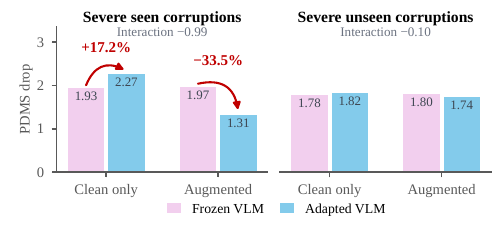}
  \caption{\textbf{Adaptation helps only together with augmentation.} PDMS drop on NAVSIM navtest for the 2$\times$2 of VLM state and inputs in three extra IL epochs. GRPO seed 1. Arrows: relative change from adapting the VLM; Appendix~\ref{sec:attribution} lists paired intervals.}
  \label{fig:factorial}
  \vspace{-4pt}
\end{figure}

\paragraph{Accuracy and Robustness.}
In \cref{tab:main}, \dbplus{} reaches 91.27 clean PDMS and 90.78 mean PDMS across the five primary corruptions, the best clean score and the best score under each of these corruptions among all compared methods. Relative to DriveBase, the five-corruption mean rises by 0.71 and the average drop falls from 0.98 to 0.48 PDMS, a 50.5\% reduction computed before rounding, and the ten-condition mean of 90.08 is higher by 0.48 $[0.18,0.79]$. Against ReCogDrive-8B RL, the clean and ten-condition gains are 0.36 $[0.13,0.60]$ and 0.67 $[0.44,0.90]$, as Appendix~\ref{sec:finaldetails} details, and in the two scenes of \cref{fig:m1}, ReCogDrive-8B RL plans a collision under blur and motion blur while \dbplus{} stays collision-free.

\begin{table}[t]
  \centering
  \caption{\textbf{Adaptation beyond extra training.} \dbplus{} minus the equal-update control, two IL seeds per GRPO seed. Severe drop: blur radius 6 and brightness 0.25. Brackets: paired 95\% log-bootstrap intervals.}
  \label{tab:adaptation-seeds}
  \setlength{\tabcolsep}{3pt}
  \begin{tabular}{@{}lcc@{}}
    \toprule
    GRPO seed & $\Delta$Ten$\uparrow$ & $\Delta$Severe drop$\downarrow$ \\
    \midrule
    1 & $+0.101$\ci{0.020}{0.179} & $-0.660$\ci{-0.831}{-0.479} \\
    2 & $+0.214$\ci{0.132}{0.298} & $-0.539$\ci{-0.729}{-0.352} \\
    \hl Pooled & $+0.157$\ci{0.088}{0.229} & $-0.600$\ci{-0.759}{-0.438} \\
    \bottomrule
  \end{tabular}
  \vspace{-4pt}
\end{table}

\paragraph{Where the Gain Comes From.}
The robustness gain holds for every training run: on the five-corruption mean, all four \dbplus{} runs, from two IL and two GRPO seeds, exceed all three DriveBase runs and both GRPO-seed-2 runs of the equal-update control, with 90.69 to 90.84 against at most 90.47. Moving the augmented recipe to the 8B family provides the largest share of the gain: DriveBase-8B raises the ten-condition mean over DriveBase by 0.37 $[0.06, 0.70]$ and lowers the seen drop of ReCogDrive-8B RL, which lacks only the augmented IL, by 0.56 $[0.30, 0.82]$. Adaptation adds a further 0.16 $[0.09, 0.23]$ to the ten-condition mean over the equal-update control, pooled over two IL and two GRPO seeds shown in \cref{tab:adaptation-seeds}, and over the same seeds \dbplus{} leads DriveBase-8B on this mean by 0.13 $[0.05, 0.20]$, with positive point differences in both GRPO seeds. The attribution and seed are detailed in Appendices~\ref{sec:attribution} and~\ref{sec:thirdseed}.

\paragraph{Adaptation Needs Augmentation.}
The effect of adaptation grows with severity: under blur radius 6 and brightness 0.25, \dbplus{} drops 0.60 $[0.44, 0.76]$ less than the equal-update control and 0.60 $[0.43, 0.76]$ less than DriveBase-8B over the 2$\times$2 runs, and against DriveBase-8B the gap widens from 0.02 PDMS at blur radius 2 to 1.35 $[1.02, 1.67]$ at radius 8. Two further paths repeat the epochs on clean inputs only. As \cref{fig:factorial} shows, for GRPO seed 1, adapting on clean inputs alone raises this drop by 0.33 $[0.13, 0.53]$ over its frozen counterpart, an interaction of $-0.99$ $[-1.24, -0.75]$ that is specific to the augmented families. The gain thus comes from letting the representation see the corruptions, not from extra training or adaptation as such, nor from ignoring the camera: the visual reliance $R$ of \dbplus{} exceeds that of the control by 0.57 $[0.18, 0.94]$. Adding motion blur, JPEG compression and noise to the augmentation extends this effect: adaptation then lowers the drop under their stronger versions by 0.47 $[0.20, 0.72]$ and raises the ten-condition mean by 0.28 $[0.18, 0.38]$, while frozen features gain no robustness from it, as Appendix~\ref{sec:attribution} details.

\paragraph{Behavior across Conditions.}
The largest gains over DriveBase occur under blur and motion blur, 1.23 and 1.06 PDMS, and over the seven training-unseen corruptions \dbplus{} averages 89.66 PDMS, the value in \cref{fig:framework}(b). Stronger corruptions widen its lead over WAM-Flow, OneVL and ReCogDrive-2B RL from 0.56--2.34 PDMS on clean inputs to 4.30--5.39 PDMS under blur radius 6 and brightness 0.25, while these three change order between conditions. Appendices~\ref{sec:navsimv2}, \ref{sec:severity} and~\ref{sec:qual} give NAVSIM v2 metrics, the severity sweep and more examples.

\subsection{Transfer of the Augmentation Recipe}
\label{sec:transfer}
Beyond diffusion planners and NAVSIM, augmentation on AutoVLA, which predicts action tokens from three cameras, and OpenDriveVLA on nuScenes cuts the seen drop over two seeds by 1.31 PDMS $[0.89, 1.74]$ and 0.53\,cm of L2 $[0.17, 0.85]$, without clean loss, as Appendix~\ref{sec:transferapp} details.

\section{Conclusion}
GWMDriveVLA brings ten driving VLA families and a world-action reference under a shared evaluation protocol. Our analysis shows that ground-truth sample selection can inflate scores more than the differences between methods, and that clean accuracy alone does not predict robustness. Among the factors tested, action-head design has the largest effect; IL-stage augmentation transfers across planners and benchmarks, while the effect of RL on robustness varies by method. These findings guide DriveBase+, which adapts the representation under corruption augmentation before RL. It ranks first among the evaluated methods on clean inputs and every primary corruption, while halving the mean degradation of its 2B predecessor. We will release the framework, corruption protocol, and models to support reproducible research on robust driving VLAs.

\paragraph{Scope and Outlook.}
Our controlled study focuses on diffusion planners on NAVSIM. GWMDriveVLA provides the infrastructure to extend this analysis to other planner families, closed-loop driving, and naturally degraded scenes. These extensions would test whether robustness gains persist when planners interact with other road users. Incorporating extended comfort into reward design offers a further direction for improving driving behavior beyond the objectives captured by the current PDMS reward.

\section*{LLM Usage Statement}
We used Claude and OpenAI Codex to assist with developing and debugging experimental code, organizing and checking experiment logs, and searching the literature. The authors designed the study, determined the experiments and analyses, and wrote the manuscript. Scientific decisions and verification of the reported results remain the authors' responsibility, and the authors take full responsibility for the manuscript.

{
  \small
  \bibliographystyle{ieeenat_fullname}
  \bibliography{main}

\begin{thebibliography}{67}
\providecommand{\natexlab}[1]{#1}
\providecommand{\url}[1]{\texttt{#1}}
\expandafter\ifx\csname urlstyle\endcsname\relax
  \providecommand{\doi}[1]{doi: #1}\else
  \providecommand{\doi}{doi: \begingroup \urlstyle{rm}\Url}\fi

\bibitem[Black et~al.(2024)Black, Brown, Driess, Esmail, Equi, Finn, Fusai,
  Groom, Hausman, Ichter, et~al.]{black2024pi0}
Kevin Black, Noah Brown, Danny Driess, Adnan Esmail, Michael Equi, Chelsea
  Finn, Niccolo Fusai, Lachy Groom, Karol Hausman, Brian Ichter, et~al.
\newblock $\pi_0$: A vision-language-action flow model for general robot
  control.
\newblock \emph{arXiv preprint arXiv:2410.24164}, 2024.

\bibitem[Bouthillier et~al.(2021)Bouthillier, Delaunay, Bronzi,
  et~al.]{bouthillier2021accounting}
Xavier Bouthillier, Pierre Delaunay, Mirko Bronzi, et~al.
\newblock Accounting for variance in machine learning benchmarks.
\newblock In \emph{Proceedings of Machine Learning and Systems}, 2021.

\bibitem[Brohan et~al.(2023)Brohan, Brown, Carbajal, Chebotar, Chen,
  Choromanski, Ding, Driess, Dubey, Finn, et~al.]{brohan2023rt2}
Anthony Brohan, Noah Brown, Justice Carbajal, Yevgen Chebotar, Xi Chen,
  Krzysztof Choromanski, Tianli Ding, Danny Driess, Avinava Dubey, Chelsea
  Finn, et~al.
\newblock {RT-2}: Vision-language-action models transfer web knowledge to
  robotic control.
\newblock \emph{arXiv preprint arXiv:2307.15818}, 2023.

\bibitem[Caesar et~al.(2020)Caesar, Bankiti, Lang, Vora, Liong, Xu, Krishnan,
  Pan, Baldan, and Beijbom]{caesar2020nuscenes}
Holger Caesar, Varun Bankiti, Alex~H. Lang, Sourabh Vora, Venice~Erin Liong,
  Qiang Xu, Anush Krishnan, Yu Pan, Giancarlo Baldan, and Oscar Beijbom.
\newblock {nuScenes}: A multimodal dataset for autonomous driving.
\newblock In \emph{Proceedings of the IEEE/CVF Conference on Computer Vision
  and Pattern Recognition}, 2020.

\bibitem[Cao et~al.(2025)Cao, Hallgarten, Li, Dauner, Gu, Wang, Miron, Aiello,
  Li, Gilitschenski, et~al.]{cao2025pseudo}
Wei Cao, Marcel Hallgarten, Tianyu Li, Daniel Dauner, Xunjiang Gu, Caojun Wang,
  Yakov Miron, Marco Aiello, Hongyang Li, Igor Gilitschenski, et~al.
\newblock Pseudo-simulation for autonomous driving.
\newblock In \emph{Conference on Robot Learning}, 2025.

\bibitem[Chen et~al.(2019)Chen, Wang, Pang, Cao, Xiong, Li, Sun, Feng, Liu, Xu,
  et~al.]{chen2019mmdetection}
Kai Chen, Jiaqi Wang, Jiangmiao Pang, Yuhang Cao, Yu Xiong, Xiaoxiao Li,
  Shuyang Sun, Wansen Feng, Ziwei Liu, Jiarui Xu, et~al.
\newblock {MMDetection}: Open {MMLab} detection toolbox and benchmark.
\newblock \emph{arXiv preprint arXiv:1906.07155}, 2019.

\bibitem[Chitta et~al.(2023)Chitta, Prakash, Jaeger, Yu, Renz, and
  Geiger]{chitta2023transfuser}
Kashyap Chitta, Aditya Prakash, Bernhard Jaeger, Zehao Yu, Katrin Renz, and
  Andreas Geiger.
\newblock {TransFuser}: Imitation with transformer-based sensor fusion for
  autonomous driving.
\newblock \emph{IEEE Transactions on Pattern Analysis and Machine
  Intelligence}, 45\penalty0 (11):\penalty0 12878--12895, 2023.

\bibitem[Dauner et~al.(2023)Dauner, Hallgarten, Geiger, and
  Chitta]{dauner2023parting}
Daniel Dauner, Marcel Hallgarten, Andreas Geiger, and Kashyap Chitta.
\newblock Parting with misconceptions about learning-based vehicle motion
  planning.
\newblock In \emph{Conference on Robot Learning}, 2023.

\bibitem[Dauner et~al.(2024)Dauner, Hallgarten, Li, Weng, Huang, Yang, Li,
  Gilitschenski, Ivanovic, Pavone, et~al.]{dauner2024navsim}
Daniel Dauner, Marcel Hallgarten, Tianyu Li, Xinshuo Weng, Zhiyu Huang, Zetong
  Yang, Hongyang Li, Igor Gilitschenski, Boris Ivanovic, Marco Pavone, et~al.
\newblock {NAVSIM}: Data-driven non-reactive autonomous vehicle simulation and
  benchmarking.
\newblock In \emph{Advances in Neural Information Processing Systems}, 2024.

\bibitem[Dong et~al.(2023)Dong, Kang, Zhang, Zhu, Wang, Yang, Su, Wei, and
  Zhu]{dong2023benchmarking}
Yinpeng Dong, Caixin Kang, Jinlai Zhang, Zijian Zhu, Yikai Wang, Xiao Yang,
  Hang Su, Xingxing Wei, and Jun Zhu.
\newblock Benchmarking robustness of {3D} object detection to common
  corruptions in autonomous driving.
\newblock In \emph{Proceedings of the IEEE/CVF Conference on Computer Vision
  and Pattern Recognition}, 2023.

\bibitem[Dosovitskiy et~al.(2017)Dosovitskiy, Ros, Codevilla, Lopez, and
  Koltun]{dosovitskiy2017carla}
Alexey Dosovitskiy, German Ros, Felipe Codevilla, Antonio Lopez, and Vladlen
  Koltun.
\newblock {CARLA}: An open urban driving simulator.
\newblock In \emph{Conference on Robot Learning}, 2017.

\bibitem[Efron(1979)]{efron1979bootstrap}
Bradley Efron.
\newblock Bootstrap methods: Another look at the jackknife.
\newblock \emph{The Annals of Statistics}, 7\penalty0 (1):\penalty0 1--26,
  1979.

\bibitem[Fan et~al.(2023)Fan, Liang, Shen, Hou, Huang, and Yu]{fan2023opengait}
Chao Fan, Junhao Liang, Chuanfu Shen, Saihui Hou, Yongzhen Huang, and Shiqi Yu.
\newblock {OpenGait}: Revisiting gait recognition towards better practicality.
\newblock In \emph{Proceedings of the IEEE/CVF Conference on Computer Vision
  and Pattern Recognition}, 2023.

\bibitem[Fu et~al.(2025{\natexlab{a}})Fu, Zhang, Zhao, Cui, Liang, Zhang,
  Zhang, Xie, Wang, and Bai]{fu2025orion}
Haoyu Fu, Diankun Zhang, Zongchuang Zhao, Jianfeng Cui, Dingkang Liang, Chong
  Zhang, Dingyuan Zhang, Hongwei Xie, Bing Wang, and Xiang Bai.
\newblock {ORION}: A holistic end-to-end autonomous driving framework by
  vision-language instructed action generation.
\newblock In \emph{Proceedings of the IEEE/CVF International Conference on
  Computer Vision}, 2025{\natexlab{a}}.

\bibitem[Fu et~al.(2025{\natexlab{b}})Fu, Zhang, Zhao, Cui, Xie, Wang, Chen,
  Ye, Liang, and Bai]{minddrive}
Haoyu Fu, Diankun Zhang, Zongchuang Zhao, Jianfeng Cui, Hongwei Xie, Bing Wang,
  Guang Chen, Hangjun Ye, Dingkang Liang, and Xiang Bai.
\newblock {MindDrive}: A vision-language-action model for autonomous driving
  via online reinforcement learning.
\newblock \emph{arXiv preprint arXiv:2512.13636}, 2025{\natexlab{b}}.

\bibitem[Guo et~al.(2023)Guo, Zhang, Lu, Duan, Wang, Yang, Zhu, and
  Chen]{guo2023openstereo}
Xianda Guo, Chenming Zhang, Juntao Lu, Yiqun Duan, Yiqi Wang, Tian Yang, Zheng
  Zhu, and Long Chen.
\newblock {OpenStereo}: A comprehensive benchmark for stereo matching and
  strong baseline.
\newblock \emph{arXiv preprint arXiv:2312.00343}, 2023.

\bibitem[Henderson et~al.(2018)Henderson, Islam, Bachman, Pineau, Precup, and
  Meger]{henderson2018deep}
Peter Henderson, Riashat Islam, Philip Bachman, Joelle Pineau, Doina Precup,
  and David Meger.
\newblock Deep reinforcement learning that matters.
\newblock In \emph{Proceedings of the AAAI Conference on Artificial
  Intelligence}, 2018.

\bibitem[Hendrycks and Dietterich(2019)]{hendrycks2019benchmarking}
Dan Hendrycks and Thomas Dietterich.
\newblock Benchmarking neural network robustness to common corruptions and
  perturbations.
\newblock In \emph{International Conference on Learning Representations}, 2019.

\bibitem[Hendrycks et~al.(2020)Hendrycks, Mu, Cubuk, Zoph, Gilmer, and
  Lakshminarayanan]{hendrycks2020augmix}
Dan Hendrycks, Norman Mu, Ekin~D. Cubuk, Barret Zoph, Justin Gilmer, and Balaji
  Lakshminarayanan.
\newblock {AugMix}: A simple data processing method to improve robustness and
  uncertainty.
\newblock In \emph{International Conference on Learning Representations}, 2020.

\bibitem[Ho et~al.(2020)Ho, Jain, and Abbeel]{ho2020ddpm}
Jonathan Ho, Ajay Jain, and Pieter Abbeel.
\newblock Denoising diffusion probabilistic models.
\newblock In \emph{Advances in Neural Information Processing Systems}, 2020.

\bibitem[Hu et~al.(2022{\natexlab{a}})Hu, Shen, Wallis, Allen-Zhu, Li, Wang,
  Wang, and Chen]{hu2022lora}
Edward~J. Hu, Yelong Shen, Phillip Wallis, Zeyuan Allen-Zhu, Yuanzhi Li, Shean
  Wang, Lu Wang, and Weizhu Chen.
\newblock {LoRA}: Low-rank adaptation of large language models.
\newblock In \emph{International Conference on Learning Representations},
  2022{\natexlab{a}}.

\bibitem[Hu et~al.(2022{\natexlab{b}})Hu, Chen, Wu, Li, Yan, and
  Tao]{hu2022stp3}
Shengchao Hu, Li Chen, Penghao Wu, Hongyang Li, Junchi Yan, and Dacheng Tao.
\newblock {ST-P3}: End-to-end vision-based autonomous driving via
  spatial-temporal feature learning.
\newblock In \emph{Proceedings of the European Conference on Computer Vision},
  2022{\natexlab{b}}.

\bibitem[Hu et~al.(2023)Hu, Yang, Chen, Li, Sima, Zhu, Chai, Du, Lin, Wang,
  et~al.]{hu2023uniad}
Yihan Hu, Jiazhi Yang, Li Chen, Keyu Li, Chonghao Sima, Xizhou Zhu, Siqi Chai,
  Senyao Du, Tianwei Lin, Wenhai Wang, et~al.
\newblock Planning-oriented autonomous driving.
\newblock In \emph{Proceedings of the IEEE/CVF Conference on Computer Vision
  and Pattern Recognition}, 2023.

\bibitem[Hwang et~al.(2025)Hwang, Xu, Lin, Hung, Ji, Choi, Huang, He,
  Covington, Sapp, et~al.]{hwang2024emma}
Jyh-Jing Hwang, Runsheng Xu, Hubert Lin, Wei-Chih Hung, Jingwei Ji, Kristy
  Choi, Di Huang, Tong He, Paul Covington, Benjamin Sapp, et~al.
\newblock {EMMA}: End-to-end multimodal model for autonomous driving.
\newblock \emph{Transactions on Machine Learning Research}, 2025.

\bibitem[Jia et~al.(2024)Jia, Yang, Li, Zhang, and Yan]{jia2024bench2drive}
Xiaosong Jia, Zhenjie Yang, Qifeng Li, Zhiyuan Zhang, and Junchi Yan.
\newblock {Bench2Drive}: Towards multi-ability benchmarking of closed-loop
  end-to-end autonomous driving.
\newblock In \emph{Advances in Neural Information Processing Systems, Datasets
  and Benchmarks Track}, 2024.

\bibitem[Jiang et~al.(2023)Jiang, Chen, Xu, Liao, Chen, Zhou, Zhang, Liu,
  Huang, and Wang]{jiang2023vad}
Bo Jiang, Shaoyu Chen, Qing Xu, Bencheng Liao, Jiajie Chen, Helong Zhou, Qian
  Zhang, Wenyu Liu, Chang Huang, and Xinggang Wang.
\newblock {VAD}: Vectorized scene representation for efficient autonomous
  driving.
\newblock In \emph{Proceedings of the IEEE/CVF International Conference on
  Computer Vision}, 2023.

\bibitem[Jiang et~al.(2024)Jiang, Wang, Zhang, Chen, Dong, Bao, Zhang, and
  Fu]{jiang2024robuste2e}
Wei Jiang, Lu Wang, Tianyuan Zhang, Yuwei Chen, Jian Dong, Wei Bao, Zichao
  Zhang, and Qiang Fu.
\newblock {RobustE2E}: Exploring the robustness of end-to-end autonomous
  driving.
\newblock \emph{Electronics}, 13\penalty0 (16):\penalty0 3299, 2024.

\bibitem[Kendall(1938)]{kendall1938}
Maurice~G. Kendall.
\newblock A new measure of rank correlation.
\newblock \emph{Biometrika}, 30\penalty0 (1/2):\penalty0 81--93, 1938.

\bibitem[Kim et~al.(2024)Kim, Pertsch, Karamcheti, Xiao, Balakrishna, Nair,
  Rafailov, Foster, Lam, Sanketi, et~al.]{kim2024openvla}
Moo~Jin Kim, Karl Pertsch, Siddharth Karamcheti, Ted Xiao, Ashwin Balakrishna,
  Suraj Nair, Rafael Rafailov, Ethan Foster, Grace Lam, Pannag Sanketi, et~al.
\newblock {OpenVLA}: An open-source vision-language-action model.
\newblock In \emph{Conference on Robot Learning}, 2024.

\bibitem[Kong et~al.(2023)Kong, Liu, Li, Chen, Zhang, Ren, Pan, Chen, and
  Liu]{kong2023robo3d}
Lingdong Kong, Youquan Liu, Xin Li, Runnan Chen, Wenwei Zhang, Jiawei Ren,
  Liang Pan, Kai Chen, and Ziwei Liu.
\newblock {Robo3D}: Towards robust and reliable {3D} perception against
  corruptions.
\newblock In \emph{Proceedings of the IEEE/CVF International Conference on
  Computer Vision}, 2023.

\bibitem[Li et~al.(2026{\natexlab{a}})Li, Shang, Liu, Zhan, Wang, Wang, Chen,
  Wang, An, Tang, et~al.]{drivevlaw0}
Yingyan Li, Shuyao Shang, Weisong Liu, Bing Zhan, Haochen Wang, Yuqi Wang,
  Yuntao Chen, Xiaoman Wang, Yasong An, Chufeng Tang, et~al.
\newblock {DriveVLA-W0}: World models amplify data scaling law in autonomous
  driving.
\newblock In \emph{International Conference on Learning Representations},
  2026{\natexlab{a}}.

\bibitem[Li et~al.(2026{\natexlab{b}})Li, Xiong, Guo, Li, Yan, Xu, Zhou, Chen,
  Sun, Wang, et~al.]{recogdrive}
Yongkang Li, Kaixin Xiong, Xiangyu Guo, Fang Li, Sixu Yan, Gangwei Xu, Lijun
  Zhou, Long Chen, Haiyang Sun, Bing Wang, et~al.
\newblock {ReCogDrive}: A reinforced cognitive framework for end-to-end
  autonomous driving.
\newblock In \emph{International Conference on Learning Representations},
  2026{\natexlab{b}}.

\bibitem[Li et~al.(2026{\natexlab{c}})Li, Zhou, Yan, Liao, Yan, Xiong, Chen,
  Xie, Wang, Chen, et~al.]{unidrivevla}
Yongkang Li, Lijun Zhou, Sixu Yan, Bencheng Liao, Tianyi Yan, Kaixin Xiong,
  Long Chen, Hongwei Xie, Bing Wang, Guang Chen, et~al.
\newblock {UniDriveVLA}: Unifying understanding, perception, and action
  planning for autonomous driving.
\newblock \emph{arXiv preprint arXiv:2604.02190}, 2026{\natexlab{c}}.

\bibitem[Li et~al.(2024{\natexlab{a}})Li, Li, Wang, Lan, Yu, Ji, Li, Zhu,
  Kautz, Wu, et~al.]{hydramdp}
Zhenxin Li, Kailin Li, Shihao Wang, Shiyi Lan, Zhiding Yu, Yishen Ji, Zhiqi Li,
  Ziyue Zhu, Jan Kautz, Zuxuan Wu, et~al.
\newblock {Hydra-MDP}: End-to-end multimodal planning with multi-target
  hydra-distillation.
\newblock \emph{arXiv preprint arXiv:2406.06978}, 2024{\natexlab{a}}.

\bibitem[Li et~al.(2024{\natexlab{b}})Li, Yu, Lan, Li, Kautz, Lu, and
  Alvarez]{li2024egostatus}
Zhiqi Li, Zhiding Yu, Shiyi Lan, Jiahan Li, Jan Kautz, Tong Lu, and Jose~M.
  Alvarez.
\newblock Is ego status all you need for open-loop end-to-end autonomous
  driving?
\newblock In \emph{Proceedings of the IEEE/CVF Conference on Computer Vision
  and Pattern Recognition}, 2024{\natexlab{b}}.

\bibitem[Li et~al.(2025)Li, Yao, Wang, Sun, Chen, Chang, Shen, Wu, Lan, and
  Alvarez]{gtrs}
Zhenxin Li, Wenhao Yao, Zi Wang, Xinglong Sun, Joshua Chen, Nadine Chang,
  Maying Shen, Zuxuan Wu, Shiyi Lan, and Jose~M. Alvarez.
\newblock Generalized trajectory scoring for end-to-end multimodal planning.
\newblock \emph{arXiv preprint arXiv:2506.06664}, 2025.

\bibitem[Liao et~al.(2025{\natexlab{a}})Liao, Chen, Yin, Jiang, Wang, Yan,
  Zhang, Li, Zhang, Zhang, et~al.]{liao2025diffusiondrive}
Bencheng Liao, Shaoyu Chen, Haoran Yin, Bo Jiang, Cheng Wang, Sixu Yan, Xinbang
  Zhang, Xiangyu Li, Ying Zhang, Qian Zhang, et~al.
\newblock {DiffusionDrive}: Truncated diffusion model for end-to-end autonomous
  driving.
\newblock In \emph{Proceedings of the IEEE/CVF Conference on Computer Vision
  and Pattern Recognition}, 2025{\natexlab{a}}.

\bibitem[Liao et~al.(2025{\natexlab{b}})Liao, Qi, Shu, Zhang, Lin, Liu, and
  Ma]{liao2025robodrivevlm}
Dacheng Liao, Mengshi Qi, Peng Shu, Zhining Zhang, Yuxin Lin, Liang Liu, and
  Huadong Ma.
\newblock {RoboDriveVLM}: A novel benchmark and baseline towards robust
  vision-language models for autonomous driving.
\newblock \emph{arXiv preprint arXiv:2512.01300}, 2025{\natexlab{b}}.

\bibitem[Lipman et~al.(2023)Lipman, Chen, Ben-Hamu, Nickel, and
  Le]{lipman2023flow}
Yaron Lipman, Ricky T.~Q. Chen, Heli Ben-Hamu, Maximilian Nickel, and Matt Le.
\newblock Flow matching for generative modeling.
\newblock In \emph{International Conference on Learning Representations}, 2023.

\bibitem[Loshchilov and Hutter(2019)]{loshchilov2019adamw}
Ilya Loshchilov and Frank Hutter.
\newblock Decoupled weight decay regularization.
\newblock In \emph{International Conference on Learning Representations}, 2019.

\bibitem[Lu et~al.(2026)Lu, Guan, Huang, Li, Li, Kong, Li, Wang, Liu, Luo,
  et~al.]{onevl}
Jinghui Lu, Jiayi Guan, Zhijian Huang, Jinlong Li, Guang Li, Lingdong Kong,
  Yingyan Li, Han Wang, Zheng Liu, Yuechen Luo, et~al.
\newblock {OneVL}: One-step latent reasoning and planning with vision-language
  explanation.
\newblock In \emph{Advances in Neural Information Processing Systems}, 2026.
\newblock arXiv:2604.18486.

\bibitem[Luo et~al.(2025)Luo, Li, Xu, Lai, Yang, Chen, Luo, Xie, Jiang, Liu,
  et~al.]{adathinkdrive}
Yuechen Luo, Fang Li, Shaoqing Xu, Zhiyi Lai, Lei Yang, Qimao Chen, Ziang Luo,
  Zixun Xie, Shengyin Jiang, Jiaxin Liu, et~al.
\newblock {AdaThinkDrive}: Adaptive thinking via reinforcement learning for
  autonomous driving.
\newblock \emph{arXiv preprint arXiv:2509.13769}, 2025.

\bibitem[Luo et~al.(2026)Luo, Li, Xu, Ji, Zhang, Wang, Shen, Cui, Chen, Chen,
  et~al.]{lastvla}
Yuechen Luo, Fang Li, Shaoqing Xu, Yang Ji, Zehan Zhang, Bing Wang, Yuannan
  Shen, Jianwei Cui, Long Chen, Guang Chen, et~al.
\newblock {LaST-VLA}: Thinking in latent spatio-temporal space for
  vision-language-action in autonomous driving.
\newblock In \emph{Advances in Neural Information Processing Systems}, 2026.

\bibitem[Michaelis et~al.(2019)Michaelis, Mitzkus, Geirhos, Rusak, Bringmann,
  Ecker, Bethge, and Brendel]{michaelis2019benchmarking}
Claudio Michaelis, Benjamin Mitzkus, Robert Geirhos, Evgenia Rusak, Oliver
  Bringmann, Alexander~S. Ecker, Matthias Bethge, and Wieland Brendel.
\newblock Benchmarking robustness in object detection: Autonomous driving when
  winter is coming.
\newblock \emph{arXiv preprint arXiv:1907.07484}, 2019.

\bibitem[Musgrave et~al.(2020)Musgrave, Belongie, and Lim]{musgrave2020metric}
Kevin Musgrave, Serge Belongie, and Ser-Nam Lim.
\newblock A metric learning reality check.
\newblock In \emph{Proceedings of the European Conference on Computer Vision},
  2020.

\bibitem[Que and Yao(2027)]{que2026drivedegrade}
Haohua Que and Handong Yao.
\newblock When does corruption break the plan? {A} degradation-tolerance
  benchmark for camera-only end-to-end driving.
\newblock In \emph{Proceedings of the IEEE/CVF Winter Conference on
  Applications of Computer Vision}, 2027.

\bibitem[Rawal et~al.(2026)Rawal, Gupta, Hu, and Zhan]{nord}
Ishaan Rawal, Shubh Gupta, Yihan Hu, and Wei Zhan.
\newblock {NoRD}: A data-efficient vision-language-action model that drives
  without reasoning.
\newblock In \emph{Proceedings of the IEEE/CVF Conference on Computer Vision
  and Pattern Recognition}, 2026.

\bibitem[Ren et~al.(2025)Ren, Lidard, Ankile, Simeonov, Agrawal, Majumdar,
  Burchfiel, Dai, and Simchowitz]{ren2025dppo}
Allen~Z. Ren, Justin Lidard, Lars~L. Ankile, Anthony Simeonov, Pulkit Agrawal,
  Anirudha Majumdar, Benjamin Burchfiel, Hongkai Dai, and Max Simchowitz.
\newblock Diffusion policy policy optimization.
\newblock In \emph{International Conference on Learning Representations}, 2025.

\bibitem[Renz et~al.(2025)Renz, Chen, Arani, and Sinavski]{renz2025simlingo}
Katrin Renz, Long Chen, Elahe Arani, and Oleg Sinavski.
\newblock {SimLingo}: Vision-only closed-loop autonomous driving with
  language-action alignment.
\newblock In \emph{Proceedings of the IEEE/CVF Conference on Computer Vision
  and Pattern Recognition}, 2025.

\bibitem[Shao et~al.(2024{\natexlab{a}})Shao, Hu, Wang, Song, Waslander, Liu,
  and Li]{shao2024lmdrive}
Hao Shao, Yuxuan Hu, Letian Wang, Guanglu Song, Steven~L. Waslander, Yu Liu,
  and Hongsheng Li.
\newblock {LMDrive}: Closed-loop end-to-end driving with large language models.
\newblock In \emph{Proceedings of the IEEE/CVF Conference on Computer Vision
  and Pattern Recognition}, 2024{\natexlab{a}}.

\bibitem[Shao et~al.(2024{\natexlab{b}})Shao, Wang, Zhu, Xu, Song, Bi, Zhang,
  Zhang, Li, Wu, et~al.]{shao2024deepseekmath}
Zhihong Shao, Peiyi Wang, Qihao Zhu, Runxin Xu, Junxiao Song, Xiao Bi, Haowei
  Zhang, Mingchuan Zhang, Y.~K. Li, Y. Wu, et~al.
\newblock {DeepSeekMath}: Pushing the limits of mathematical reasoning in open
  language models.
\newblock \emph{arXiv preprint arXiv:2402.03300}, 2024{\natexlab{b}}.

\bibitem[Sheng et~al.(2026)Sheng, Ye, Luo, Chen, and Ren]{sheng2026explorevla}
Zihao Sheng, Xin Ye, Jingru Luo, Sikai Chen, and Liu Ren.
\newblock {ExploreVLA}: Dense world modeling and exploration for end-to-end
  autonomous driving.
\newblock \emph{arXiv preprint arXiv:2604.02714}, 2026.

\bibitem[Shi et~al.(2026)Shi, Xu, Shi, Sheng, Zhang, and Jiang]{drivewam}
Chen Shi, Jinrui Xu, Shaoshuai Shi, Kehua Sheng, Bo Zhang, and Li Jiang.
\newblock {DriveWAM}: Video generative priors enable scalable world-action
  modeling for autonomous driving.
\newblock In \emph{Advances in Neural Information Processing Systems}, 2026.

\bibitem[Sima et~al.(2024)Sima, Renz, Chitta, Chen, Zhang, Xie, Bei{\ss}wenger,
  Luo, Geiger, and Li]{sima2024drivelm}
Chonghao Sima, Katrin Renz, Kashyap Chitta, Li Chen, Hanxue Zhang, Chengen Xie,
  Jens Bei{\ss}wenger, Ping Luo, Andreas Geiger, and Hongyang Li.
\newblock {DriveLM}: Driving with graph visual question answering.
\newblock In \emph{Proceedings of the European Conference on Computer Vision},
  2024.

\bibitem[Tian et~al.(2024)Tian, Gu, Li, Liu, Wang, Zhao, Zhan, Jia, Lang, and
  Zhao]{tian2024drivevlm}
Xiaoyu Tian, Junru Gu, Bailin Li, Yicheng Liu, Yang Wang, Zhiyong Zhao, Kun
  Zhan, Peng Jia, Xianpeng Lang, and Hang Zhao.
\newblock {DriveVLM}: The convergence of autonomous driving and large
  vision-language models.
\newblock In \emph{Conference on Robot Learning}, 2024.

\bibitem[Wu et~al.(2019)Wu, Kirillov, Massa, Lo, and
  Girshick]{wu2019detectron2}
Yuxin Wu, Alexander Kirillov, Francisco Massa, Wan-Yen Lo, and Ross Girshick.
\newblock Detectron2.
\newblock \url{https://github.com/facebookresearch/detectron2}, 2019.

\bibitem[Xie et~al.(2025{\natexlab{a}})Xie, Kong, Dong, Sima, Zhang, Chen, Liu,
  and Pan]{xie2025drivebench}
Shaoyuan Xie, Lingdong Kong, Yuhao Dong, Chonghao Sima, Wenwei Zhang, Qi~Alfred
  Chen, Ziwei Liu, and Liang Pan.
\newblock Are {VLMs} ready for autonomous driving? {A}n empirical study from
  the reliability, data and metric perspectives.
\newblock In \emph{Proceedings of the IEEE/CVF International Conference on
  Computer Vision}, 2025{\natexlab{a}}.

\bibitem[Xie et~al.(2025{\natexlab{b}})Xie, Kong, Zhang, Ren, Pan, Chen, and
  Liu]{xie2025benchmarking}
Shaoyuan Xie, Lingdong Kong, Wenwei Zhang, Jiawei Ren, Liang Pan, Kai Chen, and
  Ziwei Liu.
\newblock Benchmarking and improving bird's eye view perception robustness in
  autonomous driving.
\newblock \emph{IEEE Transactions on Pattern Analysis and Machine
  Intelligence}, 2025{\natexlab{b}}.

\bibitem[Xu et~al.(2026)Xu, Cui, Zhu, Shang, Luan, Xu, Cai, Zhang, Li, Cai,
  et~al.]{wamflow}
Yifang Xu, Jiahao Cui, Zhihao Zhu, Hanlin Shang, Shan Luan, Mingwang Xu,
  Feipeng Cai, Neng Zhang, Yaoyi Li, Jia Cai, et~al.
\newblock {WAM-Flow}: Parallel coarse-to-fine motion planning via discrete flow
  matching for autonomous driving.
\newblock In \emph{Proceedings of the IEEE/CVF Conference on Computer Vision
  and Pattern Recognition}, 2026.

\bibitem[Yang et~al.(2026{\natexlab{a}})Yang, Chen, Tan, Wang, Feng, Liu,
  Sheng, Zhang, Yang, Zhang, et~al.]{driveteach}
Yuguang Yang, Canyu Chen, Zhewen Tan, Yizhi Wang, Zichao Feng, Chunyang Liu,
  Kehua Sheng, Juan Zhang, Linlin Yang, Baochang Zhang, et~al.
\newblock Teaching vision-language-action models what to see and where to look.
\newblock In \emph{Proceedings of the European Conference on Computer Vision},
  2026{\natexlab{a}}.

\bibitem[Yang et~al.(2026{\natexlab{b}})Yang, Chai, Jia, Li, Shao, Zhu, Su, and
  Yan]{drivemoe}
Zhenjie Yang, Yilin Chai, Xiaosong Jia, Qifeng Li, Yuqian Shao, Xuekai Zhu,
  Haisheng Su, and Junchi Yan.
\newblock {DriveMoE}: Mixture-of-experts for vision-language-action model in
  end-to-end autonomous driving.
\newblock In \emph{Proceedings of the IEEE/CVF Conference on Computer Vision
  and Pattern Recognition}, 2026{\natexlab{b}}.

\bibitem[Yao et~al.(2026)Yao, Li, Lan, Wang, Sun, Alvarez, and Wu]{drivesuprim}
Wenhao Yao, Zhenxin Li, Shiyi Lan, Zi Wang, Xinglong Sun, Jose~M. Alvarez, and
  Zuxuan Wu.
\newblock {DriveSuprim}: Towards precise trajectory selection for end-to-end
  planning.
\newblock In \emph{Proceedings of the AAAI Conference on Artificial
  Intelligence}, 2026.

\bibitem[Zhai et~al.(2023)Zhai, Feng, Du, Mao, Liu, Tan, Zhang, Ye, and
  Wang]{zhai2023rethinking}
Jiang-Tian Zhai, Ze Feng, Jinhao Du, Yongqiang Mao, Jiang-Jiang Liu, Zichang
  Tan, Yifu Zhang, Xiaoqing Ye, and Jingdong Wang.
\newblock Rethinking the open-loop evaluation of end-to-end autonomous driving
  in {nuScenes}.
\newblock \emph{arXiv preprint arXiv:2305.10430}, 2023.

\bibitem[Zhang et~al.(2026)Zhang, Jin, Peng, Guo, Liu, Zhang, Ma, Wu, Yan, Jia,
  et~al.]{zhang2026bench2driverobust}
Zhiyuan Zhang, Zhenghao Jin, Yanlun Peng, Xianda Guo, Haoran Liu, Shaofeng
  Zhang, Xingjun Ma, Zuxuan Wu, Junchi Yan, Xiaosong Jia, et~al.
\newblock {Bench2Drive-Robust}: Benchmarking closed-loop autonomous driving
  under deployment perturbations.
\newblock \emph{arXiv preprint arXiv:2605.18059}, 2026.

\bibitem[Zhou et~al.(2026{\natexlab{a}})Zhou, Han, Yang, Ma, Tresp, and
  Knoll]{opendrivevla}
Xingcheng Zhou, Xuyuan Han, Feng Yang, Yunpu Ma, Volker Tresp, and Alois Knoll.
\newblock {OpenDriveVLA}: Towards end-to-end autonomous driving with large
  vision language action model.
\newblock In \emph{Proceedings of the AAAI Conference on Artificial
  Intelligence}, 2026{\natexlab{a}}.

\bibitem[Zhou et~al.(2026{\natexlab{b}})Zhou, Zhao, Yang, Li, Zhong, Bai, Chu,
  Chen, Li, Tang, et~al.]{qwendrive}
Xin Zhou, Zongchuang Zhao, Zhibo Yang, Mingsheng Li, Humen Zhong, Shuai Bai, Du
  Chu, Ruizhe Chen, Zhaohai Li, Jun Tang, et~al.
\newblock {Qwen-Drive-1.0}: An initial step towards a vision-language
  foundation model for autonomous driving.
\newblock \emph{arXiv preprint arXiv:2609.00111}, 2026{\natexlab{b}}.

\bibitem[Zhou et~al.(2025)Zhou, Cai, Zhao, Zhang, Huang, Zhou, and Ma]{autovla}
Zewei Zhou, Tianhui Cai, Seth~Z. Zhao, Yun Zhang, Zhiyu Huang, Bolei Zhou, and
  Jiaqi Ma.
\newblock {AutoVLA}: A vision-language-action model for end-to-end autonomous
  driving with adaptive reasoning and reinforcement fine-tuning.
\newblock In \emph{Advances in Neural Information Processing Systems}, 2025.

\end{thebibliography}
}

\onecolumn
\appendix
\crefname{section}{Appendix}{Appendices}
\Crefname{section}{Appendix}{Appendices}
\captionsetup[figure]{name=Fig.}
\captionsetup[table]{name=Tab.}
\crefname{figure}{Fig.}{Figs.}
\Crefname{figure}{Fig.}{Figs.}
\crefname{table}{Tab.}{Tabs.}
\Crefname{table}{Tab.}{Tabs.}
\newcolumntype{N}[1]{>{\centering\arraybackslash}p{#1}}
\renewcommand{\yes}{\ensuremath{\checkmark}}
\renewcommand{\no}{\ensuremath{\times}}
\renewcommand{\dn}[1]{\makebox[0pt][l]{\ensuremath{{}_{\scriptscriptstyle\color{red!70!black}#1}}}}
\renewcommand{\up}[1]{\makebox[0pt][l]{\ensuremath{{}_{\scriptscriptstyle\color{green!45!black}#1}}}}
\renewcommand{\vci}[3]{\begin{tabular}[c]{@{}c@{}}$#1$\\[-1pt]{\fontsize{6}{7}\selectfont[$#2$,\,$#3$]}\end{tabular}}
\makeatletter
\setlength{\@fptop}{0pt}
\setlength{\@fpsep}{12pt}
\setlength{\@fpbot}{0pt plus 1fil}
\makeatother
\setlength{\textfloatsep}{12pt plus 2pt minus 2pt}
\setlength{\intextsep}{8pt plus 2pt minus 2pt}
\renewcommand{\floatpagefraction}{0.8}
\renewcommand{\topfraction}{0.9}
\renewcommand{\bottomfraction}{0.9}
\renewcommand{\textfraction}{0.05}
\newcommand{\supppart}[1]{\par\addvspace{10pt}\noindent{\large\bfseries #1}\par\nopagebreak\addvspace{2pt}}

\begin{center}
  {\large\bfseries Supplementary Material\par}
  \vspace{1.2em}
\end{center}

This supplement extends each part of the main paper and keeps its order. It provides the implementation details and the integrated model zoo of \fw{}, the full benchmark and evaluation results behind Sec.~\ref{sec:prior}, the detailed controlled study of Sec.~\ref{sec:deconstruct}, and the complete evaluation, development and attribution of \dbplus{} from Sec.~\ref{sec:drivebase}, as \cref{tab:supp_overview} lists.
\begin{table}[htbp]
  \setlength{\tabcolsep}{4pt}
  \renewcommand{\arraystretch}{1.06}
  \centering
  \caption{\textbf{Organization of the supplementary material.}}
  \label{tab:supp_overview}
  \small
  \begin{tabular}{@{}lll@{}}
    \toprule
    Part & Main text & Sections \\
    \midrule
    I. Implementation and framework & Secs.~\ref{sec:drivescope} and \ref{sec:deconstruct} & \ref{sec:impl}~Implementation details; \ref{sec:timeline}~Driving VLA timeline; \ref{sec:zoo}~Model zoo and efficiency \\
    II. Benchmark and evaluation & Secs.~\ref{sec:data} and \ref{sec:prior} & \ref{sec:weather}~Weather and lighting; \ref{sec:repro_full}~Reproduction; \ref{sec:sota_app}~Published results; \\
    & & \ref{sec:inflationapp}~Evaluation inflation; \ref{sec:validity}~Validity of the corruption suite \\
    III. Controlled study & Sec.~\ref{sec:deconstruct} & \ref{sec:rl_detail}~RL results; \ref{sec:navsimv2}~NAVSIM v2; \ref{sec:addmetrics}~Visual reliance; \ref{sec:ragrpo}~Robustness-aware GRPO \\
    IV. \dbplus{} & Sec.~\ref{sec:drivebase} & \ref{sec:finaldetails}~Final evaluation; \ref{sec:posthoc}~Development; \ref{sec:attribution}~Attribution; \ref{sec:thirdseed}~Seeds; \\
    & & \ref{sec:severity}~Severity sweep; \ref{sec:qual}~Qualitative results; \ref{sec:transferapp}~Transfer \\
    V. Action-head analyses & Sec.~\ref{sec:head} & \ref{sec:headsens}~Sensitivity of the comparison; \ref{sec:table3fp}~Full-precision values of Tab.~\ref{tab:ablation} \\
    \bottomrule
  \end{tabular}
\end{table}

\FloatBarrier
\supppart{Part I. Implementation and Framework}
\section{Implementation Details}
\label{sec:impl}
This section gives the settings behind every experiment of the main paper.

\paragraph{Seed Indices.}
All corruptions except the noise are deterministic, and the noise is seeded per token, camera and frame. We index the IL seeds as 1, 2 and 3 and the GRPO seeds as 1 and 2. For the 2B study, these indices correspond to random seeds 17, 23 and 29 for IL and 2027 for GRPO. The 8B adaptation uses IL seeds 1 and 2, the second GRPO seed is 2028, and the evaluation seed is 8101. The bootstrap resampling uses seed 20271003 for the main analyses and seed 1 for the comparisons across scales and of the mean PDMS under corruption.

\paragraph{Statistics.}
Unless a table states otherwise, intervals are paired 95\% percentile intervals from 5{,}000 log-level bootstrap resamples over the 136 navtest logs with seed 20271003, and results of several seeds are averaged per token before pairing. Invalid outputs count as zero PDMS for every method. The intervals are not adjusted for multiple comparisons and describe scene sampling for fixed trained models; \cref{sec:thirdseed} discusses the variation across training seeds.

\paragraph{Data Augmentation.}
\Cref{tab:aug} lists the IL-stage augmentation results of Sec.~\ref{sec:aug}.

\begin{table}[htbp]
  \setlength{\tabcolsep}{4pt}
  \renewcommand{\arraystretch}{1.06}
  \centering
  \caption{\textbf{Data augmentation at the IL stage.} Blur: \tname{blur4}; Dim: \tname{dim050}. $\ast$: significant clean-score change from Both for IL seed 1.}
  \label{tab:aug}
  \small
  \setlength{\tabcolsep}{5pt}
  \begin{tabular}{@{}llccN{64pt}cc@{}}
    \toprule
    & & & & & \multicolumn{2}{c}{Mean drop$\downarrow$} \\
    \cmidrule(lr){6-7}
    Training seeds & Variant & Blur & Dim & Clean$\uparrow$ & Seen & Unseen \\
    \midrule
    \multirow{2}{*}{2 IL seeds} & \cellcolor{gray!12} Aug-IL & \cellcolor{gray!12}\yes & \cellcolor{gray!12}\yes & \cellcolor{gray!12}86.82 & \cellcolor{gray!12}\best{0.38} & \cellcolor{gray!12}0.80 \\
    & Base-IL & \no & \no & 86.70 & 2.20 & -- \\
    \midrule
    \multirow{3}{*}{\shortstack[l]{IL seed 1\\(all retrained)}} & \cellcolor{gray!12} Both & \cellcolor{gray!12}\yes & \cellcolor{gray!12}\yes & \cellcolor{gray!12}87.09 & \cellcolor{gray!12}\best{0.38} & \cellcolor{gray!12}1.14 \\
    & Blur only & \yes & \no & 86.68\dn{-0.41^\ast} & \best{0.38} & 1.04 \\
    & Dim only & \no & \yes & 86.83 & 1.36 & 1.36 \\
    \bottomrule
  \end{tabular}
\end{table}

\paragraph{GRPO Objective.}
For a scene $s$ with clean features $c_0$, GRPO samples a group of $G=8$ denoising chains $\tau_{1:G}$, scores each with the PDMS-based reward $r_i$, and maximizes $\sum_i \hat A_i\,\tilde\ell_\theta(\tau_i\mid c)$, where $\hat A_i=(r_i-\mu)/\sigma$ with $\mu$ and $\sigma$ the mean and standard deviation of the group rewards, and $\tilde\ell_\theta$ is the discounted sum over denoising steps of per-coordinate log-likelihoods clipped to $[-5,2]$. A behavior-cloning term $\lambda_{\text{BC}}\mathcal L_{\text{BC}}$ with $\lambda_{\text{BC}}=0.1$ is added as in ReCogDrive. It maximizes, on clean features, the clipped log-likelihood of the actor on denoising chains sampled from a frozen reference policy. For all runs on the 2B VLM, including DriveBase and the robustness-aware variants, this reference is the released ReCogDrive-2B IL planner, which reads its own front-camera features. The 8B family uses the released 8B IL as its BC reference; the actor is initialized from the checkpoint specified by each recipe. The actor and the reference are loaded independently, so initializing the actor from Aug-IL or Base-IL does not change the reference.

The decision rules below define the degradation with the opposite sign, $\mathrm{PDMS}_k-\mathrm{PDMS}_{\text{clean}}$, so a positive $\Delta\bar D$ corresponds to a reduction of the mean drop in the main text. Our controlled experiments start from \textbf{Base-IL}, the ReCogDrive-2B planner retrained by imitation inside our framework on cached VLM features with three IL seeds, and from \textbf{Aug-IL}, the same planner trained on corrupted inputs, as detailed in Sec.~\ref{sec:aug}. On top of either, we run standard GRPO and three robustness-aware variants, as detailed in Sec.~\ref{sec:rl}. From Aug-IL, standard GRPO and RA-GRPO are denoted Aug-IL + GRPO and Aug-IL + RA-GRPO; Aug-IL + GRPO is the DriveBase recipe. We compare with the released IL and RL weights of ReCogDrive-2B and the released weights of WAM-Flow and AutoVLA on NAVSIM, and with the remaining methods of Tab.~\ref{tab:repro} on their native benchmarks. The controlled GRPO runs start from an IL checkpoint and use 14{,}530 updates with GRPO seed 1, a group size $G=8$, a discount $\gamma=0.6$ and five denoising steps. We use AdamW~\cite{loshchilov2019adamw} with a learning rate of $10^{-4}$, betas of 0.9 and 0.95, a weight decay of $10^{-4}$ and a cosine schedule, and a global batch of 64, formed by a micro-batch of 8 on four GPUs with two accumulation steps. Following the official ReCogDrive code, log-likelihoods are clipped per coordinate to $[-5,2]$, gradients are clipped to norm 1.0, and the behavior-cloning coefficient is 0.1, whereas the original paper reports 0.01 and 0.001. The training reward weights progress by 10, while the evaluation score uses 5. IL planners are trained for 128 epochs with a global batch of 128 in FP32 on frozen VLM features. Aug-IL assigns the three conditions in turn every epoch, at the evaluation strengths. A GRPO run takes 90 to 198 GPU-hours on four NVIDIA H20 GPUs, and a Base-IL run 25 to 28 GPU-hours. Every model is evaluated at its final checkpoint with a single sampled trajectory. Methods whose papers use another NAVSIM version are re-scored with v1.1. Unless marked otherwise, our results average two IL seeds that share one GRPO seed.

\paragraph{Decision Rules.}
Three decisions use fixed margins. \textbf{RL criterion:} a robustness-aware RL variant works if, relative to standard GRPO from the same initialization, the lower 95\% bounds of $\Delta\bar D^{\mathcal S}$ and $\Delta\bar D^{\mathcal T}$ are both above 0 and the lower bound of the clean difference is above $-0.5$ PDMS. \textbf{Overall criterion:} a model is better overall than the reproduced NAVSIM methods if its clean PDMS is at least 90.53, the score of ReCogDrive-2B trained in our framework minus a margin of 0.5 PDMS, if it degrades less than the least-degraded reproduced method under \tname{blur4}, $-2.13$ for WAM-Flow, and under \tname{dim050}, $-0.53$ for ReCogDrive-2B RL, and if it degrades less than every reproduced method on at least two of the three unseen corruptions. \textbf{Final recipe:} if Aug-IL + RA-GRPO satisfies the RL criterion against Aug-IL + GRPO, it becomes our recipe; otherwise Aug-IL + GRPO does, provided that it satisfies the overall criterion.

\section{Driving VLA Timeline}
\label{sec:timeline}
\fw{} covers the main planner paradigms of recent driving VLAs, which the timeline below places in order.

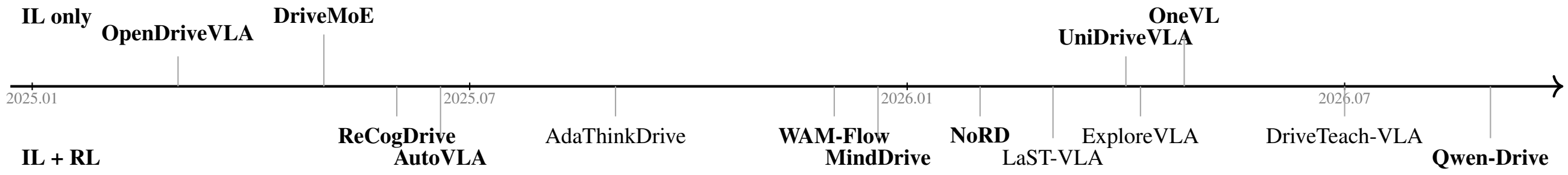
\begin{figure}[htbp]
  \centering
  \begin{tikzpicture}[x=0.8cm, y=0.6cm, font=\scriptsize]
    \draw[->, thick] (-0.3,0) -- (21.0,0);
    \foreach \m/\lab in {0/2025.01, 6/2025.07, 12/2026.01, 18/2026.07} {
      \draw (\m,0.07) -- (\m,-0.07);
      \node[font=\tiny, text=gray] at (\m,-0.22) {\lab};
    }
    \node[anchor=west, font=\scriptsize\bfseries] at (-0.3,1.25) {IL only};
    \node[anchor=west, font=\scriptsize\bfseries] at (-0.3,-1.30) {IL + RL};
    \foreach \x/\y/\t in {2/0.55/\textbf{OpenDriveVLA}, 4/0.95/\textbf{DriveMoE}, 15/0.55/\textbf{UniDriveVLA}, 15.8/0.95/\textbf{OneVL}} {
      \draw[gray!70] (\x,0) -- (\x,\y);
      \node[anchor=south] at (\x,\y) {\t};
    }
    \foreach \x/\y/\t in {5/-0.55/\textbf{ReCogDrive}, 5.6/-0.95/\textbf{AutoVLA}, 8/-0.55/AdaThinkDrive, 11/-0.55/\textbf{WAM-Flow}, 11.6/-0.95/\textbf{MindDrive}, 13/-0.55/\textbf{NoRD}, 14/-0.95/LaST-VLA, 15.2/-0.55/ExploreVLA, 18/-0.55/DriveTeach-VLA, 20/-0.95/\textbf{Qwen-Drive}} {
      \draw[gray!70] (\x,0) -- (\x,\y);
      \node[anchor=north] at (\x,\y) {\t};
    }
  \end{tikzpicture}
  \caption{\textbf{Timeline of driving VLAs.} Top: imitation only. Bottom: with RL fine-tuning. Bold: integrated in \fw{}.}
  \label{fig:timeline}
\end{figure}

\section{Model Zoo and Efficiency}
\label{sec:zoo}
Every integrated method runs with its native network, and \dbplus{} adds only 12.45\,M LoRA parameters while running faster than ReCogDrive-8B, in 583\,ms against 598\,ms per frame.

\Cref{tab:zoo} lists the size, memory and latency of the integrated methods, all measured on one H20 GPU with a batch size of one, 20 warm-up and 200 measured samples. Each method runs its native pipeline, so the end-to-end times also differ in input loading and preprocessing, and only the planner times of methods with a separable planner are directly comparable. DriveBase shares the architecture of ReCogDrive-2B but runs through a dedicated inference backend, so we do not report its end-to-end time. DriveBase+ is timed through its released end-to-end entry, which runs the 8B VLM with the 12.45M unmerged LoRA parameters ($^\ast$) and three denoising steps. It reproduces the evaluated trajectories exactly on 32 clean navtest tokens per IL seed. End to end, it takes 583 ms and 17.6 GB, on par with ReCogDrive-8B. WAM-Flow runs on unmodified official code, and the other methods on official code with documented changes. For DriveMoE and MindDrive, part of the parameters cannot be assigned to the VLM or the planner, and the planner of UniDriveVLA contains VLM submodules.

\begin{table}[htbp]
  \setlength{\tabcolsep}{4pt}
  \renewcommand{\arraystretch}{1.06}
  \centering
  \caption{\textbf{Model zoo on one H20 GPU.} Memory: peak PyTorch allocation. Planner: the native planning call, which includes token generation for autoregressive models. --: not separable or not reported.}
  \label{tab:zoo}
  \small
  \setlength{\tabcolsep}{3pt}
  \begin{tabular}{@{}lcccccc@{}}
    \toprule
    Method & Bench. & VLM (B) & Planner (M) & Mem. (GB) & Planner (ms) & Total (ms) \\
    \midrule
    ReCogDrive-2B & NAVSIM & 2.1 & 34 & 5.5 & 85 & 598 \\
    DriveBase & NAVSIM & 2.1 & 34 & 5.5 & 79 & -- \\
    ReCogDrive-8B & NAVSIM & 7.9 & 35 & 17.5 & 82 & 598 \\
    DriveBase+ & NAVSIM & 7.9$^\ast$ & 35 & 17.6 & -- & 583 \\
    AutoVLA & NAVSIM & 3.8 & -- & 8.1 & 1224 & 1499 \\
    WAM-Flow & NAVSIM & -- & -- & 12.1 & 626 & 642 \\
    Qwen-Drive & NAVSIM & 4.5 & 1040 & 12.6 & 260 & 2056 \\
    NoRD & NAVSIM & 4.1 & -- & -- & 1130 & 1159 \\
    OneVL & NAVSIM & 4.4 & -- & 9.6 & -- & 4067 \\
    OpenDriveVLA-0.5B & nuScenes & 0.6 & -- & 2.7 & -- & 2036 \\
    UniDriveVLA & nuScenes, B2D & -- & 3007 & 7.6 & 937 & 1163 \\
    DriveMoE & B2D & 2.4 & 667 & 16.0 & 393 & 510 \\
    MindDrive & B2D & -- & 15 & 5.0 & 19 & 519 \\
    \bottomrule
  \end{tabular}
\end{table}

\FloatBarrier
\supppart{Part II. Benchmark and Evaluation}
\section{Weather and Lighting Corruptions}
\label{sec:weather}
Four weather and lighting corruptions, never used for augmentation, extend the evaluation of Sec.~\ref{sec:prior}. \dbplus{} keeps the best clean score and the best score under overexposure, and its mean drop of 2.36 PDMS is lower than the 2.53 of ReCogDrive-8B RL.

\Cref{tab:weather} reports the four extended corruptions, which are excluded from our training augmentation, on the same 12{,}146 navtest tokens. Rankings differ from those of the main protocol: AutoVLA loses 19.02 PDMS on average, mostly under fog and snow, and Qwen-Drive RL degrades least. Relative to the IL models, RL changes the mean drop by $-0.50$ for ReCogDrive-8B, $-0.39$ for Qwen-Drive and $+1.43$ for NoRD, all with 95\% intervals excluding zero, and DriveBase degrades 0.54 PDMS more than Aug-IL, again a significant erosion. For the released ReCogDrive-2B, the change of $+0.10$ is not significant.

\begin{table}[htbp]
  \setlength{\tabcolsep}{4pt}
  \renewcommand{\arraystretch}{1.06}
  \centering
  \caption{\textbf{Weather and lighting corruptions} on NAVSIM navtest: PDMS per corruption, their mean, and the mean drop from the clean score. $^\ddagger$GRPO in \fw{} from the released IL weights. Bold: best; underline: second.}
  \label{tab:weather}
  \small
  \setlength{\tabcolsep}{2.6pt}
  \begin{tabular}{@{}lccccccc@{}}
    \toprule
    Model & Clean$\uparrow$ & Fog & Rain & Snow & Overexp. & Mean$\uparrow$ & Drop$\downarrow$ \\
    \midrule
    ReCogDrive-2B (RL) & 90.71 & 88.21 & 88.63 & 84.94 & 90.12 & 87.97 & 2.73 \\
    ReCogDrive-2B (IL) & 86.36 & 84.20 & 84.01 & 80.82 & 85.88 & 83.73 & 2.63 \\
    ReCogDrive-8B (RL)$^\ddagger$ & 90.91 & 88.70 & 89.06 & 85.09 & 90.64 & 88.37 & 2.53 \\
    WAM-Flow & 90.35 & 82.57 & 84.83 & 78.28 & 88.60 & 83.57 & 6.78 \\
    AutoVLA & 89.48 & 59.22 & 79.75 & 56.87 & 86.00 & 70.46 & 19.02 \\
    Qwen-Drive (SFT) & 87.95 & 85.37 & 87.45 & 85.65 & 87.81 & 86.57 & \second{1.38} \\
    Qwen-Drive (RL) & 90.50 & 89.01 & \best{90.00} & \best{88.72} & 90.29 & \best{89.51} & \best{1.00} \\
    NoRD (IL) & 76.57 & 71.94 & 76.84 & 73.21 & 76.01 & 74.50 & 2.07 \\
    NoRD (RL) & 85.80 & 79.87 & 84.66 & 79.99 & 84.70 & 82.31 & 3.50 \\
    OneVL & 88.93 & 86.19 & 87.58 & 86.13 & 88.43 & 87.08 & 1.84 \\
    \midrule
    Base-IL + GRPO & 91.02 & 88.39 & 88.87 & 85.72 & 90.67 & 88.41 & 2.61 \\
    RA-GRPO & 90.51 & 89.23 & 88.86 & 86.49 & 90.13 & 88.68 & 1.83 \\
    Aug-IL & 86.82 & 85.44 & 85.11 & 82.76 & 86.49 & 84.95 & 1.87 \\
    DriveBase & 91.05 & 89.06 & 88.89 & 86.02 & 90.60 & 88.64 & 2.41 \\
    Aug-IL + RA-GRPO & 90.89 & \best{89.50} & 89.03 & \second{86.76} & 90.45 & \second{88.93} & 1.95 \\
    DriveBase-8B & \second{91.24} & 89.08 & 89.19 & 86.17 & \second{90.93} & 88.85 & 2.39 \\
    \hl DriveBase+ & \best{91.27} & \second{89.33} & \second{89.21} & 86.02 & \best{91.06} & 88.91 & 2.36 \\
    \bottomrule
  \end{tabular}
\end{table}

\section{Detailed Reproduction Results}
\label{sec:repro_full}
Every stage trained in \fw{} matches or comes close to its reported result.

\Cref{tab:repro_full} lists the sub-metrics behind Tab.~\ref{tab:repro}. For AutoVLA, whose supervised checkpoint is not released, we train the supervised stage in our framework, which reaches 77.95 PDMS with mixed chain-of-thought supervision on the 11{,}929 navtest tokens with prepared inputs, and evaluate the released RL weights; all other evaluations cover the full 12{,}146 tokens. For MindDrive, we run one of its two PPO rounds. Tab.~\ref{tab:main} uses paired re-runs of the released models, which agree with the runs in Tab.~\ref{tab:repro} within 0.09 PDMS.

\begin{table}[htbp]
  \setlength{\tabcolsep}{4pt}
  \renewcommand{\arraystretch}{1.06}
  \centering
  \caption{\textbf{Reproduction results.} DS/SR: driving score/success rate; Eff./Comf.: efficiency/comfortness; Abil.: mean multi-ability success.}
  \label{tab:repro_full}
  \small
  \setlength{\tabcolsep}{4pt}
  \begin{minipage}[t]{\linewidth}
    \centering
    \textbf{(a) NAVSIM}\\[2pt]
    \begin{tabular}{@{}llcccccc@{}}
      \toprule
      Method & Source & NC & DAC & TTC & EP & C & PDMS \\
      \midrule
      \multirow{3}{*}{ReCogDrive-2B} & Paper & 97.90 & 97.30 & 94.90 & 87.30 & 100.00 & 90.80 \\
      & Released & 98.19 & 97.77 & 95.30 & 86.03 & 100.00 & 90.71 \\
      & Framework & 98.20 & 98.16 & 95.17 & 86.47 & 100.00 & 91.03 \\
      \midrule
      \multirow{2}{*}{ReCogDrive-8B} & Release report & -- & -- & -- & -- & -- & 90.40 \\
      & Framework & 98.11 & 98.09 & 95.35 & 86.13 & 100.00 & 90.91 \\
      \midrule
      \multirow{2}{*}{WAM-Flow} & Paper & 99.20 & 98.30 & 97.00 & 82.30 & 99.70 & 90.30 \\
      & Released & 99.45 & 98.45 & 97.14 & 82.26 & 99.57 & 90.37 \\
      \midrule
      \multirow{4}{*}{AutoVLA} & Paper$^\ast$ & 98.41 & 95.64 & 98.04 & 81.87 & 99.94 & 89.11 \\
      & Released & 99.40 & 95.97 & 97.62 & 82.73 & 99.89 & 89.57 \\
      & SFT, no CoT$^\dagger$ & 95.85 & 85.79 & 90.19 & 70.57 & 100.00 & 75.75 \\
      & SFT, mixed CoT$^\dagger$ & 96.22 & 87.76 & 90.44 & 73.21 & 99.98 & 77.95 \\
      \bottomrule
    \end{tabular}
    \\[1pt]{\footnotesize $^\ast$Single sample after RL fine-tuning; the paper also reports a best-of-$N$ result. $^\dagger$Framework-trained; 11{,}929 tokens.}
  \end{minipage}
  \par
  \vspace{8pt}
  \footnotesize
  \setlength{\tabcolsep}{2.6pt}
  \begin{minipage}[t]{0.48\linewidth}
    \vspace{0pt}
    \centering
    \textbf{(b) nuScenes}\\[2pt]
    \begin{tabular}{@{}llccccc@{}}
      \toprule
      & & \multicolumn{4}{c}{L2 (m) $\downarrow$} & Col. \\
      \cmidrule(lr){3-6}
      Method & Source & 1s & 2s & 3s & Avg & Avg \\
      \midrule
      \multirow{3}{*}{OpenDriveVLA-0.50B} & Paper & 0.15 & 0.32 & 0.57 & 0.35 & 0.09 \\
      & Released & 0.15 & 0.32 & 0.57 & 0.35 & 0.12$^\ddagger$ \\
      & Fine-tuned & 0.15 & 0.31 & 0.55 & 0.34 & 0.12$^\ddagger$ \\
      \midrule
      \multirow{2}{*}{UniDriveVLA} & Paper$^\S$ & 0.28 & 0.51 & 0.82 & 0.54 & 0.17 \\
      & Framework & 0.27 & 0.51 & 0.82 & 0.53 & 0.13$^\ddagger$ \\
      \bottomrule
    \end{tabular}
    \\[1pt]{\footnotesize $^\ddagger$Collisions with vehicles and pedestrians for OpenDriveVLA and with vehicles only for UniDriveVLA; the papers do not specify the classes. $^\S$Base model, without ego status.}
  \end{minipage}
  \hfill
  \begin{minipage}[t]{0.48\linewidth}
    \vspace{0pt}
    \centering
    \textbf{(c) Bench2Drive}\\[2pt]
    \begin{tabular}{@{}llccccc@{}}
      \toprule
      Method & Source & DS & SR & Eff. & Comf. & Abil. \\
      \midrule
      \multirow{3}{*}{DriveMoE} & Paper & 74.22 & 48.64 & 175.96 & 15.31 & 47.91 \\
      & Released & 72.44 & 45.00 & 181.66 & 18.75 & 46.34 \\
      & Framework & 73.84 & 49.09 & 180.57 & 14.72 & 51.07 \\
      \midrule
      \multirow{2}{*}{UniDriveVLA} & Paper & 78.37 & 51.82 & 198.86 & 11.78 & 51.53 \\
      & Released & 76.99 & 49.55 & 195.92 & 12.63 & -- \\
      \midrule
      \multirow{3}{*}{MindDrive} & Paper RL & 78.04 & 55.09 & -- & -- & 56.94 \\
      & Paper IL & 75.85 & -- & -- & -- & -- \\
      & Framework IL & 74.97 & 49.55 & -- & -- & 51.20 \\
      \bottomrule
    \end{tabular}
  \end{minipage}
\end{table}

\section{Comparison with Published Results}
\label{sec:sota_app}
\dbplus{} reaches the level of the strongest published single-sample results on NAVSIM.

\Cref{tab:sota} lists published single-sample results of driving VLAs on NAVSIM navtest, with values taken from the original papers. Planners that score a set of trajectory candidates with dedicated modules, such as Hydra-MDP~\cite{hydramdp}, GTRS~\cite{gtrs} and DriveSuprim~\cite{drivesuprim}, form a different family and report higher scores, up to 93.50 PDMS for DriveSuprim. We do not list them, because they select among many candidates rather than output a single VLA trajectory.

\begin{table}[htbp]
  \setlength{\tabcolsep}{4pt}
  \renewcommand{\arraystretch}{1.06}
  \centering
  \caption{\textbf{Published single-sample NAVSIM navtest results.} Rows above the rule are reported scores from the cited papers.}
  \label{tab:sota}
  \small
  \begin{tabular}{@{}lc@{}}
    \toprule
    Method & PDMS$\uparrow$ \\
    \midrule
    AutoVLA~\cite{autovla} & 89.10 \\
    WAM-Flow~\cite{wamflow} & 90.30 \\
    AdaThinkDrive~\cite{adathinkdrive} & 90.30 \\
    DriveTeach-VLA~\cite{driveteach} & 90.40 \\
    ReCogDrive~\cite{recogdrive} & 90.80 \\
    LaST-VLA-2B~\cite{lastvla} & 91.10 \\
    LaST-VLA-8B~\cite{lastvla} & 91.30 \\
    \midrule
    DriveBase & 91.05 \\
    \hl DriveBase+ & 91.27 \\
    \bottomrule
  \end{tabular}
\end{table}

\section{Evaluation Inflation across Methods}
\label{sec:inflationapp}
For the diffusion and flow-matching planners, selecting the best of six samples with ground truth raises PDMS by 0.90 to 1.34, far above noise floors of 0.12 to 0.18, which is why all comparisons in the main paper use one sample.

We evaluate five released or trained models with six evaluation seeds each on all navtest tokens and keep their default decoding. The noise floor is the largest difference in mean PDMS among the six seeds, and the inflation is the gain from choosing, for every token, the best of the six trajectories with the ground-truth score. Both use 5{,}000 log-level bootstrap resamples, and the maximum in the noise floor is recomputed in every resample. As listed in \cref{tab:inflation}, the inflation is 6.6 to 10.1 times the noise floor for all models except AutoVLA. Its default low-temperature decoding barely varies across seeds, and its inflation is 2.3 times its noise floor. Selecting the best of ten intermediate checkpoints adds only 0.15 PDMS, close to the noise floor of 0.13.

\begin{table}[htbp]
  \setlength{\tabcolsep}{4pt}
  \renewcommand{\arraystretch}{1.06}
  \centering
  \caption{\textbf{Noise floor and ground-truth selection} on NAVSIM navtest, with 95\% intervals.}
  \label{tab:inflation}
  \small
  \setlength{\tabcolsep}{3pt}
  \begin{tabular}{@{}lcc@{}}
    \toprule
    Model & Noise floor & Best of six with ground truth \\
    \midrule
    ReCogDrive-2B (RL) & \vci{0.18}{0.11}{0.33} & \vci{1.20}{1.07}{1.34} \\
    WAM-Flow & \vci{0.12}{0.08}{0.27} & \vci{0.90}{0.74}{1.09} \\
    AutoVLA & \vci{0.06}{0.04}{0.14} & \vci{0.15}{0.11}{0.19} \\
    DriveWAM, non-VLA & \vci{0.11}{0.07}{0.26} & \vci{1.00}{0.86}{1.14} \\
    \hl DriveBase & \vci{0.13}{0.08}{0.30} & \vci{1.34}{1.16}{1.54} \\
    \bottomrule
  \end{tabular}
\end{table}

\begin{figure}[htbp]
  \centering
  \includegraphics[width=0.9\textwidth]{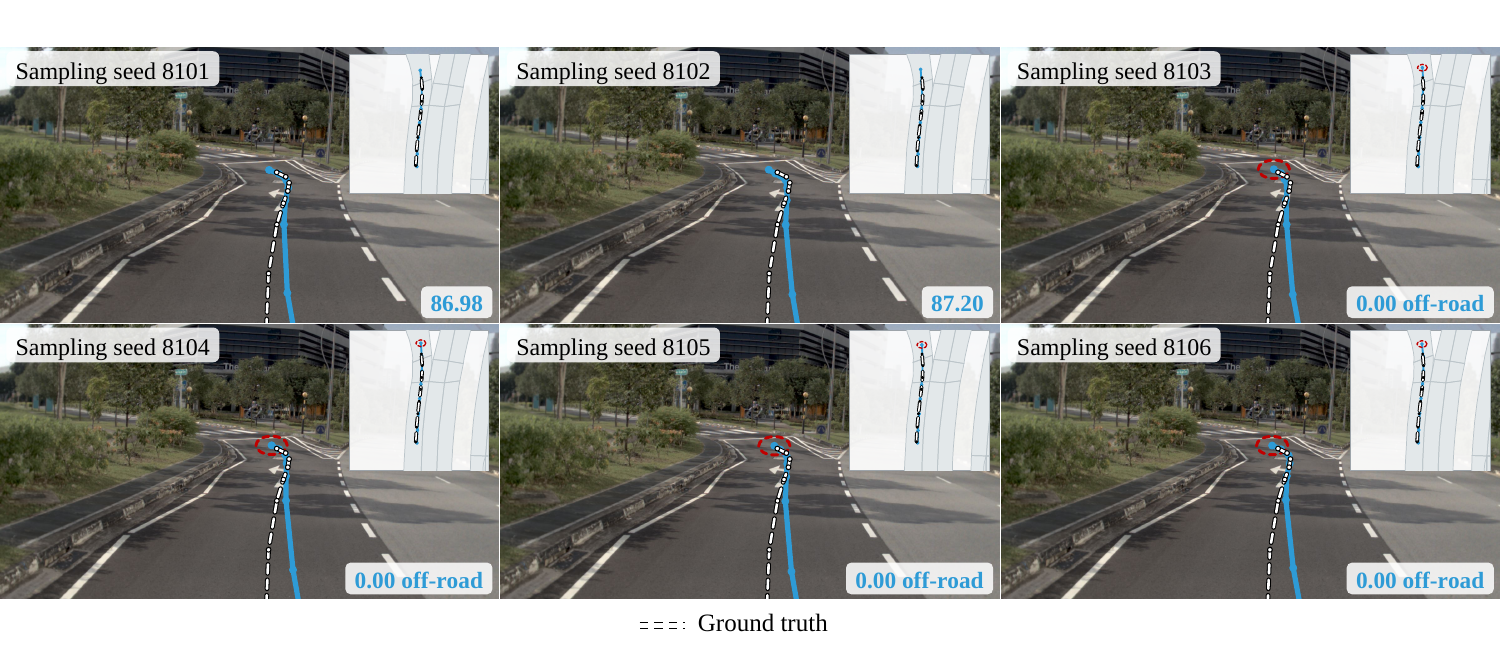}
  \caption{\textbf{Six samples of \dbplus{} on one navtest scene.} The best and the worst plan differ by at most 0.46\,m; two samples score 86.98 and 87.20, and the other four score zero because their plans leave the drivable area. Selecting the best sample with the ground truth turns such scenes from failures into successes.}
  \label{fig:q4}
\end{figure}

\paragraph{Discontinuities of PDMS.}
PDMS multiplies the no-collision and drivable-area terms, so a small change of a plan can move its score from high to zero. Over the 12,146 navtest scenes, the six samples of \dbplus{} with IL seed 1 differ by at least 50 PDMS in 153 scenes, 1.26\% of navtest. In 143 of them, the best and the worst plan stay within 1\,m of each other at all eight points, and the median of this distance over the 153 scenes is 0.43\,m. The 143 scenes provide 0.48 of the 1.16 PDMS that best-of-six selection adds for this IL seed, or 42\%. Of the 72,876 samples, 2,400 score zero: 1,241 through a collision and 1,189 by leaving the drivable area, 30 of them through both. \Cref{fig:q4} shows one of the 143 scenes.

\section{Validity of the Corruption Suite}
\label{sec:validity}
Corruptions reorder the methods, so the suite measures robustness that clean scores do not reveal.

\begin{table}[htbp]
  \setlength{\tabcolsep}{4pt}
  \renewcommand{\arraystretch}{1.06}
  \centering
  \caption{\textbf{Ranking stability.} Kendall $\tau_b$ between the clean ranking of 11 models and their ranking under each corruption, and the number of the 55 model pairs whose order flips.}
  \label{tab:ranking}
  \small
  \begin{tabular}{@{}lcc@{}}
    \toprule
    Condition & $\tau_b$ & Flips \\
    \midrule
    \tname{blur4} & 0.78 & 6 \\
    \tname{dim050} & 0.82 & 5 \\
    Motion blur & 0.82 & 5 \\
    JPEG & 0.71 & 8 \\
    \tname{noise25} & \best{0.85} & \best{4} \\
    Fog & 0.64 & 10 \\
    Rain & 0.60 & 11 \\
    Snow & 0.38 & 17 \\
    Overexposure & \best{0.85} & \best{4} \\
    \midrule
    Mean of the nine & 0.78 & 6 \\
    \bottomrule
  \end{tabular}
\end{table}

\paragraph{Corruptions Reorder Methods.}
We rank the 11 models in the main comparison that have results under all ten conditions by their PDMS and compare each corrupted ranking with the clean one. As listed in \cref{tab:ranking}, the Kendall $\tau_b$~\cite{kendall1938} ranges from 0.85 under noise and overexposure to 0.38 under snow, where 17 of the 55 model pairs change order. AutoVLA falls from sixth to last under fog and snow, while OneVL rises from seventh to second under snow. DriveBase, the clean leader in this cohort, also leads the mean over the nine corruptions in 77\% of the bootstrap replicates. A clean score does not determine the ranking under corruption, least of all under weather.

\FloatBarrier
\supppart{Part III. Controlled Study}
\section{Detailed RL Results}
\label{sec:rl_detail}
These results detail the RL analysis of Sec.~\ref{sec:rl} and support its conclusion that robustness is best built before RL.

\begin{figure}[htbp]
  \centering
  \captionsetup{skip=4pt}
  \includegraphics[width=0.83\textwidth]{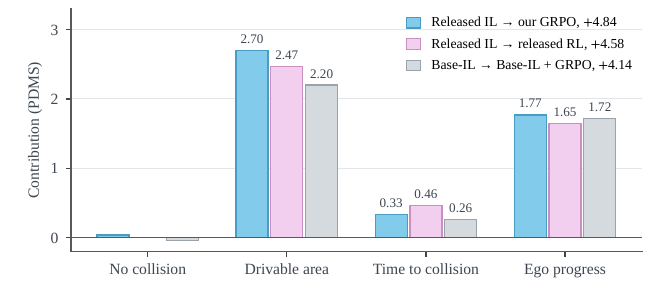}
  \caption{\textbf{Where the RL gain comes from.} Additive decomposition of clean PDMS gains; totals are shown at the right. NC: no collision; DAC: drivable-area compliance; TTC: time to collision; EP: ego progress. Exact values are listed in \cref{tab:rlgain}.}
  \label{fig:rlgain}
\end{figure}

\Cref{tab:rlgain,tab:erosion} list the exact values behind Figs.~3 and~\ref{fig:rlgain}.

\paragraph{Stability across RL Seeds.}
For IL seed 1, a second GRPO seed changes Aug-IL + GRPO's clean score and both mean drops by under 0.1 PDMS, none significantly. This is an order of magnitude below the RL and augmentation effects above.

\begin{table}[htbp]
  \setlength{\tabcolsep}{4pt}
  \renewcommand{\arraystretch}{1.06}
  \centering
  \caption{\textbf{Where the RL gain comes from.} Additive accounting into PDMS sub-scores. Rel.: released. \sig: 95\% interval excludes $0$.}
  \label{tab:rlgain}
  \small
  \setlength{\tabcolsep}{1.6pt}
  \begin{tabular}{@{}lcccccc@{}}
    \toprule
    IL $\rightarrow$ RL & $\Delta$PDMS & NC & DAC & TTC & EP & Gate \\
    \midrule
    Rel.\ IL $\rightarrow$ our GRPO & $+4.84$\sig & $+0.04$ & \best{$+2.70$}\sig & $+0.33$\sig & $+1.77$\sig & 57\% \\
    Rel.\ IL $\rightarrow$ rel.\ RL & $+4.58$\sig & $-0.00$ & \best{$+2.47$}\sig & $+0.46$\sig & $+1.65$\sig & 54\% \\
    Base-IL $\rightarrow$ Base-IL + GRPO & $+4.14$\sig & $-0.04$ & \best{$+2.20$}\sig & $+0.26$\sig & $+1.72$\sig & 52\% \\
    \bottomrule
  \end{tabular}
\end{table}

\begin{table}[htbp]
  \setlength{\tabcolsep}{4pt}
  \renewcommand{\arraystretch}{1.06}
  \centering
  \caption{\textbf{Change in the mean drop from IL to RL}; a negative change means that RL makes the model more robust. Green and red mark significant decreases and increases of the drop, and \sig{} marks a 95\% interval that excludes $0$. Rel.\ stands for released.}
  \label{tab:erosion}
  \small
  \setlength{\tabcolsep}{3.2pt}
  \begin{tabular}{@{}llcc@{}}
    \toprule
    IL $\rightarrow$ RL & Seeds & $\Delta$Seen drop$\downarrow$ & $\Delta$Unseen drop$\downarrow$ \\
    \midrule
    \multicolumn{4}{@{}l}{\emph{ReCogDrive family, non-augmented IL models}} \\
    Rel.\ 2B IL $\rightarrow$ rel.\ RL & -- & $-0.08$ & $+0.05$ \\
    Rel.\ 2B IL $\rightarrow$ our GRPO & -- & $-0.02$ & $-0.05$ \\
    Rel.\ 8B IL $\rightarrow$ our GRPO & -- & $-0.34$ & {\color{green!45!black}$-0.37$}\sig \\
    Base-IL $\rightarrow$ Base-IL + GRPO & IL 1 & $+0.14$ & $-0.18$ \\
    \midrule
    \multicolumn{4}{@{}l}{\emph{Other families, released weights}} \\
    Qwen-Drive SFT $\rightarrow$ RL & -- & {\color{green!45!black}$-1.04$}\sig & {\color{green!45!black}$-1.21$}\sig \\
    NoRD IL $\rightarrow$ RL & -- & {\color{red!70!black}$+0.60$}\sig & {\color{red!70!black}$+0.47$}\sig \\
    \midrule
    \multicolumn{4}{@{}l}{\emph{Augmented IL model}} \\
    \hl Aug-IL $\rightarrow$ DriveBase & 2 IL & {\color{red!70!black}$+0.46$}\sig & $+0.27$ \\
    \bottomrule
  \end{tabular}
\end{table}

\section{NAVSIM v2 Results}
\label{sec:navsimv2}
NAVSIM v2 scores the same predictions with EPDMS and supports the comfort analysis of Sec.~\ref{sec:rl}.

We score the clean predictions of the NAVSIM models and controlled variants with the NAVSIM v2 metrics on navtest and on the two-stage navhard split, using the official aggregation. Inputs with a navigation command that the frozen processors do not accept are excluded for all models, 13 auxiliary inputs on navtest and 14 on navhard. All 12{,}146 navtest anchors are scored, extended comfort uses the 12{,}104 anchors with a valid previous frame, and 224 of the 225 navhard groups have a valid combined score. As shown in \cref{tab:navsimv2}, DriveBase matches Base-IL + GRPO in EPDMS on both splits and reaches a higher extended comfort on navhard. RL changes the released ReCogDrive-2B by $-5.11$ EPDMS on navtest, with a 95\% interval of $[-6.13,-4.07]$, mostly through extended comfort and lane keeping, and by $+1.79$ on navhard, with an interval of $[-1.52, 5.55]$. For our 8B GRPO, the changes are $-6.08$, with an interval of $[-7.03,-5.10]$, and $-0.86$, with $[-4.57, 3.08]$. For NoRD, they are $+8.21$, with $[6.92, 9.53]$, and $+9.27$, with $[5.77, 12.96]$. For Qwen-Drive, the change on navtest is $-3.79$, with $[-4.29, -3.28]$. We do not score Qwen-Drive on navhard, because its synthetic inputs provide a 2\,Hz ego history while the model reads 10\,Hz, and we do not interpolate inputs. OneVL, which has no RL stage, keeps an extended comfort of 85.91 and reaches 88.85 navtest EPDMS, the highest among the VLA endpoints of the main comparison. The non-VLA DriveWAM reference reaches 90.33 EPDMS and 88.33 extended comfort, as \cref{tab:addmetrics} lists. Re-running the diffusion planners with a shared initial noise for all tokens, or averaging four sampled trajectories without any ground-truth selection, raises the extended comfort of DriveBase by 0.19 $[0.11, 0.28]$ and 0.50 $[0.21, 0.81]$ points at unchanged PDMS, so comfort depends on the learned policy rather than on sampling, a natural target for reward design. The main results keep the original inference.

\begin{table}[htbp]
  \setlength{\tabcolsep}{4pt}
  \renewcommand{\arraystretch}{1.06}
  \centering
  \caption{\textbf{NAVSIM v2 results} on clean inputs. EC: extended comfort; LK: lane keeping; both on navtest. Aug-IL, DriveBase, DriveBase+ and Base-IL + GRPO average two IL seeds. A dash denotes an unreported metric. $^\ddagger$GRPO in \fw{} from the released IL weights.}
  \label{tab:navsimv2}
  \small
  \setlength{\tabcolsep}{4pt}
  \begin{tabular}{@{}lcccc@{}}
    \toprule
    Model & EPDMS navtest$\uparrow$ & EC$\uparrow$ & LK$\uparrow$ & EPDMS navhard$\uparrow$ \\
    \midrule
    ReCogDrive-2B (IL) & 86.39 & 86.18 & 97.13 & 25.55 \\
    ReCogDrive-2B (RL) & 81.28 & 27.50 & 90.90 & 27.34 \\
    ReCogDrive-8B (IL) & 86.74 & 85.40 & 96.80 & 26.11 \\
    ReCogDrive-8B (RL)$^\ddagger$ & 80.65 & 21.45 & 89.46 & 25.25 \\
    NoRD (IL) & 76.26 & 81.88 & 93.27 & 18.13 \\
    NoRD (RL) & 84.47 & 80.03 & 93.23 & 27.40 \\
    Qwen-Drive (SFT) & 85.75 & 63.69 & 97.54 & -- \\
    Qwen-Drive (RL) & 81.96 & 19.67 & 95.97 & -- \\
    WAM-Flow & 83.55 & 32.45 & 96.91 & \best{31.85} \\
    AutoVLA & 88.80 & 83.22 & 98.08 & 20.43 \\
    OneVL & \best{88.85} & 85.91 & 98.00 & 28.44 \\
    Base-IL + GRPO & 81.03 & 24.29 & 90.28 & 26.28 \\
    Aug-IL & 86.81 & 86.64 & -- & -- \\
    DriveBase & 81.11 & 24.64 & 90.20 & 26.74 \\
    DriveBase-8B & 81.29 & 23.12 & -- & -- \\
    DriveBase+ & 81.24 & 22.75 & -- & -- \\
    \bottomrule
  \end{tabular}
\end{table}

\paragraph{Augmented IL Baseline.}
\Cref{tab:aug_il_v2} adds the NAVSIM v2 scores of Aug-IL, computed from its clean predictions. Before the standard GRPO stage that produces DriveBase, Aug-IL reaches 86.81 EPDMS and 86.64 extended comfort over two IL seeds.
\begin{table}[htbp]
  \setlength{\tabcolsep}{4pt}
  \renewcommand{\arraystretch}{1.06}
  \centering
  \caption{\textbf{Aug-IL on NAVSIM v2 navtest.} EC: extended comfort. Each row uses 12,146 anchors, with 12,104 valid EC scores. IL seed indices 1 and 2 correspond to training seeds 17 and 23.}
  \label{tab:aug_il_v2}
  \small
  \begin{tabular}{@{}lcccc@{}}
    \toprule
    Endpoint & EPDMS & 95\% interval & EC & 95\% interval \\
    \midrule
    Aug-IL, IL seed 1 & 86.90 & [85.64, 88.13] & 86.54 & [84.90, 88.20] \\
    Aug-IL, IL seed 2 & 86.73 & [85.48, 87.96] & 86.74 & [85.06, 88.43] \\
    Aug-IL, two-seed mean & 86.81 & [85.59, 88.03] & 86.64 & [85.01, 88.29] \\
    \bottomrule
  \end{tabular}
\end{table}

\section{Additional Visual-Dependence and NAVSIM v2 Results}
\label{sec:addmetrics}
Visual reliance supports the RL analysis of Sec.~\ref{sec:rl}: it rises after RL in every pair, in step with the drop.

After RL, $R$ increases by 1.4 PDMS for Qwen-Drive, by 2.1 to 8.0 PDMS in the ReCogDrive family and by 17.7 PDMS for NoRD. On gray images, the IL model of NoRD loses 18.6 PDMS, against 43.3 to 49.6 for DriveBase and AutoVLA. \Cref{tab:addmetrics} extends the main comparison with the controlled RL variants, the non-VLA DriveWAM reference and DriveBase+. The paired-token aggregation and original log-bootstrap draws are unchanged. Scene-GRPO and Pair-GRPO each use one IL seed; RA-GRPO and Aug-IL + RA-GRPO average two.
\begin{table}[htbp]
  \setlength{\tabcolsep}{4pt}
  \renewcommand{\arraystretch}{1.06}
  \centering
  \caption{\textbf{Additional navtest endpoints.} $R$: clean minus gray PDMS. EC: extended comfort. Intervals are 95\% log-bootstrap intervals.}
  \label{tab:addmetrics}
  \small
  \begin{tabular}{@{}lccc@{}}
    \toprule
    Endpoint & $R$ [95\% interval] & EPDMS [95\% interval] & EC [95\% interval] \\
    \midrule
    Scene-GRPO & \vci{42.67}{40.80}{44.51} & \vci{83.34}{82.57}{84.15} & \vci{36.56}{32.78}{40.60} \\
    Pair-GRPO & \vci{45.63}{43.66}{47.52} & \vci{83.33}{82.44}{84.23} & \vci{38.01}{34.15}{42.20} \\
    RA-GRPO & \vci{44.22}{42.35}{46.17} & \vci{79.19}{78.09}{80.28} & \vci{12.17}{10.77}{13.65} \\
    Aug-IL + RA-GRPO & \vci{45.08}{43.26}{46.85} & \vci{79.85}{78.77}{80.93} & \vci{15.21}{13.55}{17.10} \\
    DriveWAM (non-VLA) & \vci{35.56}{33.46}{37.75} & \vci{90.33}{89.49}{91.18} & \vci{88.33}{86.72}{89.88} \\
    DriveBase-8B & \vci{38.36}{36.45}{40.28} & \vci{81.29}{80.37}{82.19} & \vci{23.12}{20.46}{25.91} \\
    DriveBase+ & \vci{38.30}{36.41}{40.24} & \vci{81.24}{80.31}{82.16} & \vci{22.75}{20.30}{25.39} \\
    \bottomrule
  \end{tabular}
\end{table}

\section{Robustness-aware GRPO: Components and Pairwise Results}
\label{sec:ragrpo}
The robustness-aware variants of Sec.~\ref{sec:rl} improve robustness from Base-IL, and their behavior from Aug-IL supports learning robustness before RL.

\paragraph{Robustness-aware GRPO.}
Let $c_k$ denote the features of the same scene under corruption $k$. We compare four arms that differ only in how corrupted inputs enter RL, with the same number of rollouts per scene. Standard \textbf{GRPO} uses clean inputs only. \textbf{Scene-GRPO} draws one condition per scene, choosing clean, \tname{blur4} or \tname{dim050} with probabilities 0.5, 0.25 and 0.25, and uses it for all rollouts of the group. \textbf{Pair-GRPO} uses paired groups: each scene draws $k$ from $\{\tname{blur4},\tname{dim050}\}$, the first $G/2$ rollouts use $c_0$ and the others use $c_k$, and advantages are standardized over all $G$ rollouts. \textbf{RA-GRPO} adds to Pair-GRPO the distillation term $\mathcal L_{\text{distill}}=-\lambda_{\text{d}}\,\tilde\ell_\theta(x^\star\mid c_k)$, where $x^\star$ is the highest-reward clean chain of the group. The term is used only if this reward is positive and exceeds the mean reward of the corrupted rollouts. We treat $x^\star$ as a constant, set $\lambda_{\text{d}}=\lambda_{\text{BC}}=0.1$ and apply the same clipping. Clipping affects only 0.12\% of the coordinates of $\mathcal L_{\text{distill}}$.

\paragraph{Robustness-Aware Variants from Base-IL.}
As listed in \cref{tab:ragrpo}, RA-GRPO from Base-IL reduces the seen and unseen drops by 2.23 and 0.85 PDMS. It also lowers the clean PDMS significantly by 0.51, so it trades clean performance for robustness under the RL criterion. From Aug-IL, its main effect is to undo the erosion described above. Over both IL seeds, it improves the seen degradation by 0.65 PDMS over Aug-IL + GRPO but leaves the unseen degradation and the clean score unchanged. Its robustness is statistically indistinguishable from that of Aug-IL itself. In this augmented-IL comparison, RA-GRPO mainly preserves the robustness learned during imitation.

\label{sec:components}
\Cref{tab:components} separates the arms. Exposing RL to corrupted inputs in Scene-GRPO improves robustness by $+1.61$ on seen and $+0.94$ on unseen corruptions, with a clean change of $-0.12$ whose interval contains zero. Paired groups in Pair-GRPO are indistinguishable from Scene-GRPO. Distillation in RA-GRPO adds the largest seen-corruption robustness, $+2.41$, at a clean change of $-0.48$. Over both IL seeds, Scene-GRPO meets all three conditions of the RL criterion, with seen $+1.11$\ci{0.43}{1.75}, unseen $+0.67$\ci{0.14}{1.16}, clean $-0.14$\ci{-0.42}{0.13}. RA-GRPO trades part of the clean score for this robustness, as listed in \cref{tab:ragrpo}. Exposing RL to corrupted inputs improves robustness without a significant clean cost, and distillation adds seen-corruption robustness at a clean cost.

\begin{table}[htbp]
  \setlength{\tabcolsep}{4pt}
  \renewcommand{\arraystretch}{1.06}
  \centering
  \caption{\textbf{Ablation of the components} of robustness-aware GRPO from Base-IL with IL seed 1, as differences from GRPO. Red.: reduction of the mean drop. Green/red: improvement/degradation. \sig: 95\% interval excludes $0$.}
  \label{tab:components}
  \small
  \setlength{\tabcolsep}{4pt}
  \begin{tabular}{@{}lcccccc@{}}
    \toprule
    Arm & Corrupt. & Paired & Distill. & $\Delta$Clean$\uparrow$ & Seen red.$\uparrow$ & Unseen red.$\uparrow$ \\
    \midrule
    \hl RA-GRPO & \yes & \yes & \yes & \textcolor{red!70!black}{$-0.48$\sig} & \textcolor{green!45!black}{\best{$+2.41$}\sig} & \textcolor{green!45!black}{$+0.91$\sig} \\
    Pair-GRPO & \yes & \yes & \no & \textcolor{red!70!black}{$-0.22$} & \textcolor{green!45!black}{$+1.67$\sig} & \textcolor{green!45!black}{$+0.93$\sig} \\
    Scene-GRPO & \yes & \no & \no & \textcolor{red!70!black}{\best{$-0.12$}} & \textcolor{green!45!black}{$+1.61$\sig} & \textcolor{green!45!black}{\best{$+0.94$}\sig} \\
    GRPO & \no & \no & \no & -- & -- & -- \\
    \bottomrule
  \end{tabular}
\end{table}

\begin{table}[htbp]
  \setlength{\tabcolsep}{4pt}
  \renewcommand{\arraystretch}{1.06}
  \centering
  \caption{\textbf{Robustness-aware vs. standard GRPO}, 2 initializations, 2 IL seeds; red.:\ mean-drop reduction. \sig: 95\% interval excludes $0$.}
  \label{tab:ragrpo}
  \small
  \setlength{\tabcolsep}{5pt}
  \begin{tabular}{@{}lccc@{}}
    \toprule
    Comparison & Seen red.$\uparrow$ & Unseen red.$\uparrow$ & $\Delta$Clean \\
    \midrule
    Base-IL: Scene-GRPO vs.\ GRPO & $+1.11$\sig & $+0.67$\sig & $-0.14$ \\
    Base-IL: RA-GRPO vs.\ GRPO & $+2.23$\sig & $+0.85$\sig & $-0.51$\sig \\
    Aug-IL: RA-GRPO vs.\ GRPO & $+0.65$\sig & $+0.04$ & $-0.17$ \\
    Aug-IL + RA-GRPO vs.\ Aug-IL & $+0.19$ & $-0.22$ & -- \\
    \bottomrule
  \end{tabular}
\end{table}

\FloatBarrier
\supppart{Part IV. DriveBase+}
\section{DriveBase+ Final Evaluation}
\label{sec:finaldetails}
Both IL seeds of \dbplus{} exceed ReCogDrive-8B RL on clean inputs, with 91.30 and 91.24 against 90.91, and their mean leads every compared method under each primary corruption, as in Tab.~\ref{tab:main}.

We evaluate two IL seeds with three denoising steps and sampling seed 8101 on all 12{,}146 tokens from 136 logs. Scores are averaged across IL seeds per token, with 5{,}000 log-bootstrap replicates. Extended comfort uses the 12{,}104 tokens with valid previous-frame predictions for both seeds.
\begin{figure}[htbp]
  \centering
  \captionsetup{skip=4pt}
  \includegraphics[width=0.98\textwidth]{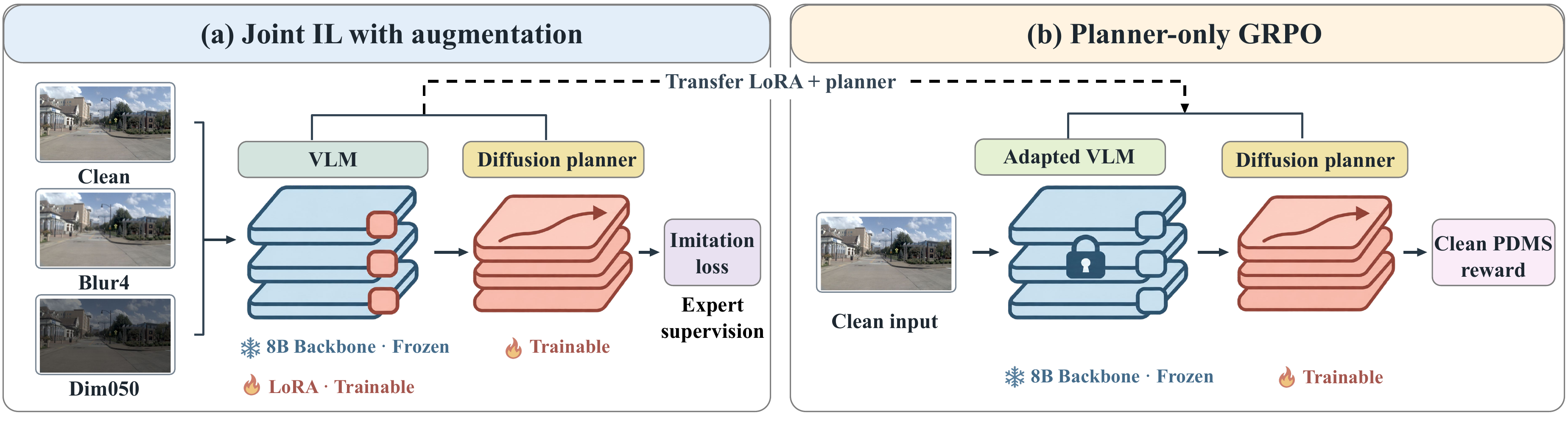}
  \caption{\textbf{DriveBase+ adapts the representation during IL and optimizes the planner with GRPO.} Ego state and navigation command condition the planner in both stages.}
  \label{fig:drivebase}
\end{figure}
\begin{table}[htbp]
  \setlength{\tabcolsep}{4pt}
  \renewcommand{\arraystretch}{1.06}
  \centering
  \caption{\textbf{DriveBase+ on NAVSIM navtest.} Means and intervals use the paired two-IL-seed aggregation.}
  \label{tab:final_e36}
  \small
  \begin{tabular}{@{}lcccc@{}}
    \toprule
    Condition / metric & IL seed 1 & IL seed 2 & Mean & 95\% interval \\
    \midrule
    Clean & 91.30 & 91.24 & 91.27 & [90.51, 92.02] \\
    Blur & 91.03 & 90.74 & 90.88 & [90.08, 91.71] \\
    Dim & 91.12 & 90.93 & 91.03 & [90.23, 91.81] \\
    Motion blur & 90.90 & 90.83 & 90.87 & [90.06, 91.69] \\
    JPEG & 90.46 & 90.40 & 90.43 & [89.56, 91.32] \\
    Noise & 90.72 & 90.73 & 90.72 & [89.92, 91.52] \\
    Fog & 89.67 & 88.99 & 89.33 & [88.39, 90.32] \\
    Rain & 89.32 & 89.10 & 89.21 & [88.28, 90.19] \\
    Snow & 86.74 & 85.31 & 86.02 & [84.84, 87.23] \\
    Overexposure & 91.08 & 91.03 & 91.06 & [90.26, 91.86] \\
    Gray & 51.40 & 54.54 & 52.97 & [50.67, 55.24] \\
    Ten-condition mean & 90.23 & 89.93 & 90.08 & [89.23, 90.94] \\
    Visual reliance $R$ & 39.90 & 36.70 & 38.30 & [36.41, 40.24] \\
    EPDMS & 81.22 & 81.26 & 81.24 & [80.31, 82.16] \\
    Extended comfort & 22.47 & 23.03 & 22.75 & [20.30, 25.39] \\
    \bottomrule
  \end{tabular}
\end{table}
\begin{table}[htbp]
  \setlength{\tabcolsep}{4pt}
  \renewcommand{\arraystretch}{1.06}
  \centering
  \caption{\textbf{Paired final comparisons.} Difference: DriveBase+ minus the reference endpoint. DriveBase-8B and the equal-update control in \cref{sec:attribution} share DriveBase+'s 8B family, both IL seeds and three-step inference; the other references are the endpoints of Tab.~\ref{tab:main}.}
  \label{tab:final_paired}
  \small
  \begin{tabular}{@{}lcccc@{}}
    \toprule
    Reference & $\Delta$Clean & 95\% interval & $\Delta$Ten-condition mean & 95\% interval \\
    \midrule
    DriveBase-8B & +0.03 & [-0.09, 0.16] & +0.11 & [0.02, 0.19] \\
    Equal-update control & +0.04 & [-0.08, 0.15] & +0.10 & [0.02, 0.18] \\
    DriveBase & +0.22 & [-0.10, 0.53] & +0.48 & [0.18, 0.79] \\
    ReCogDrive-8B RL & +0.36 & [0.13, 0.60] & +0.67 & [0.44, 0.90] \\
    WAM-Flow & +0.92 & [0.39, 1.42] & +3.37 & [2.80, 3.96] \\
    AutoVLA & +1.79 & [0.91, 2.68] & +10.99 & [10.17, 11.81] \\
    \bottomrule
  \end{tabular}
\end{table}
\begin{figure}[htbp]
  \centering
  \captionsetup{skip=4pt}
  \includegraphics[width=0.98\textwidth]{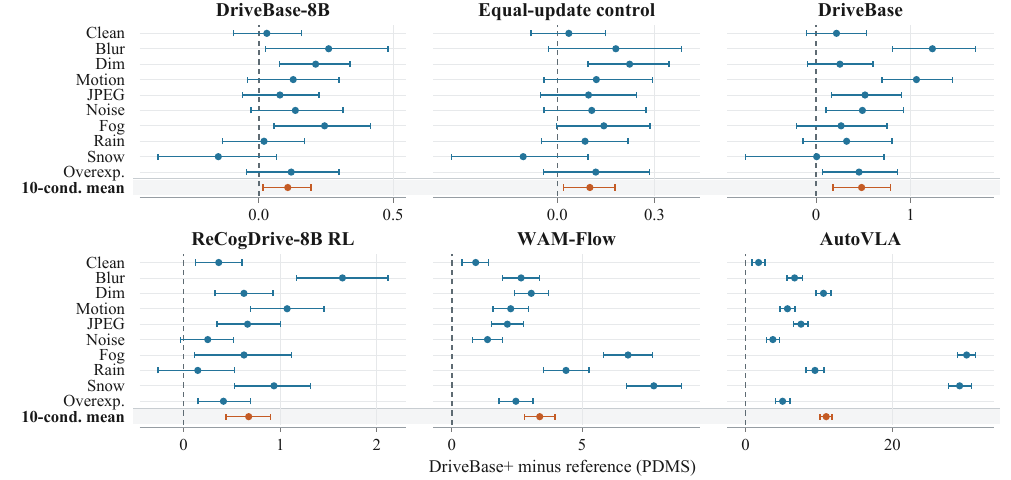}
  \caption{\textbf{Paired gains of DriveBase+ on NAVSIM navtest}, relative to references in \cref{tab:final_paired}. Points/bars: estimates/paired 95\% log-bootstrap intervals; orange: ten-condition mean. Intervals are positive for all conditions against WAM-Flow and AutoVLA, and eight of ten against ReCogDrive-8B RL. DriveBase-8B spans the full additional stage; equal-update control compares adaptation at matched updates.}
  \label{fig:final_paired}
\end{figure}

\paragraph{Adaptation Details.}
\Cref{fig:drivebase} shows the two stages. Joint IL uses rank-16 LoRA with scaling 32 and zero dropout in the vision and language attention projections, with the planner initialized from the augmented 8B IL checkpoint. The three-epoch stage uses 2{,}181 optimizer updates, global batch 128, AdamW with learning rate $10^{-4}$, betas $(0.9,0.95)$, weight decay $10^{-4}$ and gradient clipping at norm 1. The VLM and LoRA computations use BF16 and the planner uses FP32. The LoRA modules remain unmerged when extracting the actor features for GRPO and evaluation. The BC reference retains the released 8B IL model and its native features.

\begin{figure}[htbp]
  \centering
  \captionsetup{skip=4pt}
  \includegraphics[width=0.70\textwidth]{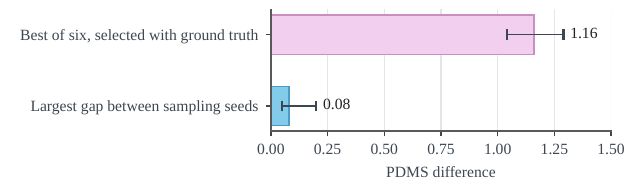}
  \caption{\textbf{Ground-truth selection inflates DriveBase+ scores.} Best-of-six minus the per-token six-sample mean, versus the maximum mean difference among 15 seed pairs. Error bars: 95\% intervals.}
  \label{fig:inflation}
\end{figure}

\paragraph{Sampling Diagnostic.}
For each token, the inflation from ground-truth selection is the maximum score across evaluation seeds 8101--8106 minus their mean. After averaging the two IL seeds, it is 1.16 PDMS with a 95\% interval of $[1.04,1.29]$. The largest absolute mean difference among the 15 pairs of evaluation seeds is 0.08 with an interval of $[0.05,0.20]$; the maximum is recomputed within every bootstrap replicate. \Cref{fig:inflation} shows both quantities.

\section{Development of DriveBase+}
\label{sec:posthoc}
A selection rule fixed on the held-out development split leads from DriveBase to \dbplus{}.

We first tested five variants of DriveBase on the held-out development split, four of the RL stage and one of the IL augmentation. A variant replaces DriveBase only if its clean PDMS is at most 0.13 lower and its mean PDMS over the clean input and the nine corruptions is more than 0.13 higher. \textbf{Corrupted-input RL} feeds corrupted inputs to RL as Scene-GRPO does, and \textbf{Stronger BC} raises $\lambda_{\text{BC}}$ from 0.1 to 0.5. \textbf{Both} combines the two. \textbf{Self-anchored} replaces the frozen BC reference, the released IL planner, by Aug-IL itself. \textbf{Broader aug.} adds the photometric operations of AugMix~\cite{hendrycks2020augmix}, which exclude our test corruptions, to the IL augmentation of Aug-IL. All variants start from IL seed 1 of Aug-IL and use GRPO seed 1. A control run of DriveBase uses the same GRPO seed as the variants.

As \cref{tab:posthoc} lists, none of these early variants improves robustness and the clean score together on the frozen 2B representation, which motivates the later candidates that change the representation itself.

\paragraph{Room on a Frozen Representation.}
The ten-condition mean equals the clean score minus 0.9 times the mean drop over the nine corruptions. With a mean drop of only 1.28 PDMS for DriveBase, even perfect robustness adds at most 1.15 PDMS, so a variant must gain robustness without a clean cost, which adapting the representation achieves. Subsequent development evaluates an 8B family recipe, LoRA adaptation and a change to the ego-progress reward weight. The 8B recipe changes both the representation and the published IL reference used for behavior cloning. Its comparison with the 2B recipe measures a change of model family.

\begin{table}[htbp]
  \setlength{\tabcolsep}{4pt}
  \renewcommand{\arraystretch}{1.06}
  \centering
  \caption{\textbf{Candidate variants of DriveBase} on the development split with three denoising steps. Mean: over the clean input and the nine corruptions. Subscripts: change from DriveBase.}
  \label{tab:posthoc}
  \small
  \setlength{\tabcolsep}{3pt}
  \begin{tabular}{@{}lN{64pt}N{64pt}ccc@{}}
    \toprule
    Variant & Clean$\uparrow$ & Mean$\uparrow$ & Seen drop$\downarrow$ & Unseen drop$\downarrow$ & Weather drop$\downarrow$ \\
    \midrule
    \hl DriveBase & 92.97 & 91.82 & 0.61 & 0.97 & 1.84 \\
    DriveBase, same seed & \best{93.02}\up{+0.05} & \best{91.87}\up{+0.04} & 0.70 & 0.95 & 1.83 \\
    RL on corrupted inputs & 93.00\up{+0.03} & 91.51\dn{-0.31} & 1.52 & 1.36 & 1.95 \\
    $\lambda_{\text{BC}}=0.5$ & 92.19\dn{-0.78} & 91.03\dn{-0.79} & 0.56 & 0.86 & 1.97 \\
    Both & 92.00\dn{-0.97} & 90.73\dn{-1.09} & 1.09 & 0.79 & 2.06 \\
    Self-anchored BC & 92.70\dn{-0.27} & 91.69\dn{-0.13} & 0.44 & 0.81 & 1.68 \\
    Broader aug. & 92.71\dn{-0.26} & 91.80\dn{-0.02} & \best{0.31} & \best{0.67} & \best{1.62} \\
    \bottomrule
  \end{tabular}
\end{table}

\paragraph{Later Candidate Recipes.}
The same point-estimate eligibility rule is applied to the later candidates. DriveBase scores 92.97 clean PDMS and 91.82 on the ten-condition mean of the development split. Eligible candidates must reach at least 92.84 clean PDMS and exceed 91.95 on that mean. The ordinary 8B recipe, 2B LoRA recipe, reduced progress-weight variant and 8B LoRA recipe all pass. We select the highest mean, the 8B adapted recipe named DriveBase+.
\begin{table}[htbp]
  \setlength{\tabcolsep}{4pt}
  \renewcommand{\arraystretch}{1.06}
  \centering
  \caption{\textbf{Recipe selection on the development split.} Three denoising steps and one sampled trajectory; values on the development split. $\Delta$: difference from DriveBase.}
  \label{tab:selection_plus}
  \small
  \begin{tabular}{@{}llcc@{}}
    \toprule
    Recipe & Main change from DriveBase & Ten-condition mean & $\Delta$ \\
    \midrule
    DriveBase & Original 2B augmentation + GRPO & 91.82 & -- \\
    2B adapted recipe & Joint vision/language LoRA IL & 92.07 & $+0.24$ \\
    Progress-weight variant & Reward weight 10 to 5 & 92.09 & $+0.27$ \\
    DriveBase-8B & 8B representation and 8B BC reference & 92.29 & $+0.47$ \\
    \hl DriveBase+ & 8B recipe with joint LoRA adaptation & \best{92.35} & $+0.52$ \\
    \bottomrule
  \end{tabular}
\end{table}
The selected recipe has clean PDMS 92.98. Against DriveBase, its clean difference is $+0.01$ with interval $[-0.40,0.40]$, and its ten-condition difference is $+0.52$ with $[0.24,0.83]$. Candidates are ranked by point estimate; the intervals give the uncertainty of the development-set contrasts. The original 8B and 2B recipes also differ in their BC references. \Cref{sec:attribution} compares both 8B recipes and an equal-update control at both training stages and on navtest.

\section{Attribution of DriveBase+}
\label{sec:attribution}
These analyses support Sec.~\ref{sec:sota}: the gain of \dbplus{} comes from adapting the representation under augmentation.

\dbplus{} differs from DriveBase in its VLM family, in three further IL epochs and in adapting the representation during those epochs. Three paths separate these steps. All start from the augmented 8B IL checkpoints of IL seeds 1 and 2 and share the standard GRPO stage of 14{,}530 updates, whose BC reference is the released 8B IL model with its original features. \textbf{DriveBase-8B} applies GRPO directly. The \textbf{equal-update control} first continues IL for the three epochs and 2{,}181 updates of \dbplus{}, with the same data, augmentation, optimizer and schedule, but keeps the VLM frozen and the LoRA increments at zero, so its GRPO actor reads the original 8B features. \textbf{\dbplus{}} adapts the vision and language attention projections during these epochs. DriveBase-8B was a recipe-selection candidate, as listed in \cref{tab:selection_plus}; the control was not included.

On navtest, both controls follow the final protocol of \dbplus{} without the six-sample diagnostic, with the same features, tokens, conditions and bootstrap draws as DriveBase-8B, as listed in \cref{tab:attr_navtest,tab:attr_navtest_paired}. Against DriveBase-8B, \dbplus{} raises the ten-condition mean by 0.11 $[0.02, 0.19]$ at an unchanged clean score, $+0.03$ $[-0.09, 0.16]$. The gain is positive for both IL seeds, $+0.16$ $[0.05, 0.27]$ and $+0.06$ $[-0.06, 0.17]$, and is largest under \tname{blur4}, \tname{dim050} and fog, as shown in \cref{fig:final_paired}, so the seen drop falls by 0.21 $[0.04, 0.36]$ while the training-unseen and weather drops change by less than 0.10. Across the 2$\times$2 IL/GRPO-seed runs, the gain is 0.13 $[0.05, 0.20]$ and its interval excludes zero for each IL seed, as detailed in \cref{sec:thirdseed}.

The equal-update control separates the two parts of this stage. Three further IL epochs on frozen features leave the 8B recipe nearly unchanged: against DriveBase-8B, the control moves the ten-condition mean by $+0.01$ $[-0.04, 0.05]$, the clean score by $-0.01$ $[-0.09, 0.08]$ and every drop by less than 0.04 PDMS, with $+0.01$ in each IL seed. \dbplus{} exceeds the control by 0.10 $[0.02, 0.18]$ on the ten-condition mean and by 0.15 $[0.04, 0.25]$ on the five-corruption mean, and its seen drop is 0.17 $[0.00, 0.32]$ lower. The per-seed gains are $+0.15$ $[0.05, 0.26]$ and $+0.05$ $[-0.05, 0.15]$, both positive, at a clean difference of $+0.04$ $[-0.08, 0.15]$. Of the gain of 0.11 PDMS that the additional stage brings on the ten-condition mean, 0.10 comes from adapting the representation. A second GRPO seed shifts \dbplus{} and DriveBase-8B alike and leaves this gain in place, as \cref{sec:thirdseed} shows. The contrasts above use GRPO seed 1; \cref{tab:equal_update_grposeed} gives the comparison over two IL and two GRPO seeds, and the severity extrapolation below separates the two parts of the stage. The control's camera reliance $R$ is lower by 0.63 $[0.24, 1.01]$ than DriveBase-8B and by 0.57 $[0.18, 0.94]$ than \dbplus{}, without corresponding changes in corruption drops.

On the development split, we evaluate five paths at their IL and GRPO endpoints. In addition to the three paths above, \textbf{clean-input adaptation} and \textbf{clean frozen} repeat the three additional IL epochs on clean inputs with the VLM adapted and frozen, respectively; these are the clean-input paths of the factorial below. All GRPO endpoints use GRPO seed 1. The protocol follows recipe selection: 4{,}000 tokens from 119 logs, clean inputs and nine corruptions, three denoising steps and evaluation seed 8101. Scores average both IL seeds per token, with 5{,}000 log-bootstrap draws and seed 1. Because the GRPO endpoints for IL seed 1 took part in recipe selection, the development split describes the training stages, and the conclusions rest on navtest. \Cref{tab:attr_stage,tab:attr_contrast} list development endpoints and paired contrasts.

\begin{table}[htbp]
  \setlength{\tabcolsep}{4pt}
  \renewcommand{\arraystretch}{1.06}
  \centering
  \caption{\textbf{Attribution endpoints on NAVSIM navtest}, two IL seeds, GRPO seed 1, three denoising steps. Mean: five primary corruptions. $R$: clean minus gray PDMS.}
  \label{tab:attr_navtest}
  \small
  \begin{tabular}{@{}lccccccc@{}}
    \toprule
    Endpoint & Clean$\uparrow$ & Mean$\uparrow$ & Ten-condition$\uparrow$ & Seen drop$\downarrow$ & Unseen drop$\downarrow$ & $R$ & EPDMS$\uparrow$ \\
    \midrule
    DriveBase-8B & 91.24 & 90.62 & 89.97 & 0.52 & 0.68 & 38.36 & 81.29 \\
    Equal-update control & 91.23 & 90.64 & 89.98 & 0.48 & 0.67 & 37.73 & 81.22 \\
    \hl \dbplus{} & 91.27 & 90.78 & 90.08 & 0.31 & 0.60 & 38.30 & 81.24 \\
    \bottomrule
  \end{tabular}
\end{table}

\begin{table}[htbp]
  \setlength{\tabcolsep}{2.5pt}
  \renewcommand{\arraystretch}{1.06}
  \centering
  \caption{\textbf{Paired attribution contrasts on NAVSIM navtest.} Left minus right, two IL seeds and GRPO seed 1, with paired 95\% log-bootstrap intervals; drop differences are negative when the left path degrades less. The last two columns give the per-seed differences of the ten-condition mean. All three contrasts use navtest, 12{,}146 tokens from 136 logs.}
  \label{tab:attr_navtest_paired}
  \small
  \setlength{\tabcolsep}{2.5pt}
  \begin{tabular}{@{}lccccccc@{}}
    \toprule
    Contrast & $\Delta$Clean & $\Delta$Ten-condition & $\Delta$Mean & $\Delta$Seen drop & $\Delta$Unseen drop & IL seed 1 & IL seed 2 \\
    \midrule
    \dbplus{} $-$ DriveBase-8B & \vci{+0.03}{-0.09}{0.16} & \vci{+0.11}{0.02}{0.19} & \vci{+0.16}{0.05}{0.27} & \vci{-0.21}{-0.36}{-0.04} & \vci{-0.08}{-0.22}{0.05} & \vci{+0.16}{0.05}{0.27} & \vci{+0.06}{-0.06}{0.17} \\
    \dbplus{} $-$ control & \vci{+0.04}{-0.08}{0.15} & \vci{+0.10}{0.02}{0.18} & \vci{+0.15}{0.04}{0.25} & \vci{-0.17}{-0.32}{0.00} & \vci{-0.07}{-0.21}{0.07} & \vci{+0.15}{0.05}{0.26} & \vci{+0.05}{-0.05}{0.15} \\
    Control $-$ DriveBase-8B & \vci{-0.01}{-0.09}{0.08} & \vci{+0.01}{-0.04}{0.05} & \vci{+0.02}{-0.04}{0.07} & \vci{-0.04}{-0.14}{0.06} & \vci{-0.01}{-0.11}{0.08} & \vci{+0.01}{-0.06}{0.08} & \vci{+0.01}{-0.06}{0.07} \\
    \bottomrule
  \end{tabular}
\end{table}

\begin{table}[htbp]
  \setlength{\tabcolsep}{4pt}
  \renewcommand{\arraystretch}{1.06}
  \centering
  \caption{\textbf{Five 8B paths across training stages} on the development split. Two IL seeds, GRPO seed 1. Mean: clean and nine corruptions. Drops: clean minus PDMS over two seen, three training-unseen and four weather corruptions. $^\ddagger$Includes the recipe-selection run of IL seed 1.}
  \label{tab:attr_stage}
  \small
  \begin{tabular}{@{}llccccc@{}}
    \toprule
    Path & Endpoint & Clean$\uparrow$ & Mean$\uparrow$ & Seen drop$\downarrow$ & Unseen drop$\downarrow$ & Weather drop$\downarrow$ \\
    \midrule
    \multirow{2}{*}{DriveBase-8B} & IL & 90.36 & 89.05 & 0.53 & 0.67 & 2.50 \\
    & GRPO$^\ddagger$ & 93.03 & 92.19 & 0.27 & 0.38 & 1.69 \\
    \midrule
    \multirow{2}{*}{Equal-update control} & IL & 87.61 & 86.70 & 0.01 & 0.42 & 1.96 \\
    & GRPO & 93.03 & 92.19 & 0.18 & 0.38 & 1.71 \\
    \midrule
    \multirow{2}{*}{\dbplus{}} & IL & 87.25 & 86.28 & 0.17 & 0.72 & 1.80 \\
    & GRPO$^\ddagger$ & 92.97 & 92.22 & 0.11 & 0.30 & 1.61 \\
    \midrule
    \multirow{2}{*}{Clean-input adaptation} & IL & 87.67 & 86.45 & 0.52 & 0.86 & 2.15 \\
    & GRPO & 93.05 & 92.14 & 0.38 & 0.40 & 1.78 \\
    \midrule
    \multirow{2}{*}{Clean frozen} & IL & 87.72 & 86.61 & 0.32 & 0.63 & 2.15 \\
    & GRPO & 93.04 & 92.19 & 0.23 & 0.39 & 1.71 \\
    \bottomrule
  \end{tabular}
\end{table}

\begin{table}[!htb]
  \setlength{\tabcolsep}{2.5pt}
  \renewcommand{\arraystretch}{1.06}
  \centering
  \caption{\textbf{Additional attribution contrasts} on the development split. Two-IL-seed differences with 95\% log-bootstrap intervals; the last two columns give the per-seed differences of the ten-condition mean. $^\ast$(\dbplus{} GRPO $-$ \dbplus{} IL) $-$ (control GRPO $-$ control IL). $^\dagger$The analogous stage interaction for clean-input adaptation minus clean frozen.}
  \label{tab:attr_contrast}
  \small
  \begin{tabular}{@{}lccccc@{}}
    \toprule
    Contrast & $\Delta$Clean & $\Delta$Mean & $\Delta$Seen drop & IL seed 1 & IL seed 2 \\
    \midrule
    Control $-$ DriveBase-8B, IL endpoint & \vci{-2.75}{-3.42}{-2.14} & \vci{-2.35}{-2.88}{-1.88} & \vci{-0.53}{-0.86}{-0.21} & $-2.51$ & $-2.19$ \\
    Control $-$ DriveBase-8B, GRPO endpoint & \vci{0.00}{-0.13}{0.12} & \vci{+0.01}{-0.07}{0.08} & \vci{-0.09}{-0.24}{0.06} & $-0.04$ & $+0.05$ \\
    \dbplus{} $-$ control, IL endpoint & \vci{-0.36}{-0.87}{0.10} & \vci{-0.42}{-0.90}{0.03} & \vci{+0.16}{-0.14}{0.45} & $-0.95$ & $+0.11$ \\
    \dbplus{} $-$ control, GRPO endpoint & \vci{-0.06}{-0.27}{0.13} & \vci{+0.02}{-0.06}{0.10} & \vci{-0.07}{-0.34}{0.17} & $+0.10$ & $-0.05$ \\
    \midrule
    Clean-input adaptation $-$ clean frozen, IL endpoint & \vci{-0.05}{-0.36}{0.25} & \vci{-0.16}{-0.34}{0.01} & \vci{+0.20}{-0.12}{0.57} & $+0.40$ & $-0.72$ \\
    Clean-input adaptation $-$ clean frozen, GRPO endpoint & \vci{+0.01}{-0.17}{0.16} & \vci{-0.06}{-0.14}{0.02} & \vci{+0.16}{-0.05}{0.37} & $-0.01$ & $-0.10$ \\
    \dbplus{} $-$ clean-input adaptation, GRPO endpoint & \vci{-0.08}{-0.28}{0.11} & \vci{+0.08}{-0.02}{0.17} & \vci{-0.272}{-0.534}{-0.001} & $+0.13$ & $+0.03$ \\
    \dbplus{} $-$ clean frozen, GRPO endpoint & \vci{-0.07}{-0.29}{0.13} & \vci{+0.02}{-0.07}{0.11} & \vci{-0.12}{-0.36}{0.10} & $+0.11$ & $-0.07$ \\
    \midrule
    GRPO $-$ IL, DriveBase-8B & \vci{+2.67}{2.00}{3.35} & \vci{+3.13}{2.61}{3.71} & \vci{-0.27}{-0.59}{0.07} & $+3.10$ & $+3.17$ \\
    GRPO $-$ IL, control & \vci{+5.41}{4.64}{6.34} & \vci{+5.49}{4.74}{6.35} & \vci{+0.17}{-0.19}{0.54} & $+5.58$ & $+5.41$ \\
    GRPO $-$ IL, \dbplus{} & \vci{+5.72}{4.66}{6.90} & \vci{+5.93}{4.91}{7.15} & \vci{-0.06}{-0.46}{0.36} & $+6.62$ & $+5.25$ \\
    GRPO $-$ IL, clean-input adaptation & \vci{+5.38}{4.49}{6.45} & \vci{+5.69}{4.80}{6.76} & \vci{-0.14}{-0.54}{0.25} & $+5.65$ & $+5.73$ \\
    GRPO $-$ IL, clean frozen & \vci{+5.32}{4.44}{6.38} & \vci{+5.59}{4.74}{6.63} & \vci{-0.10}{-0.50}{0.33} & $+6.07$ & $+5.11$ \\
    Interaction$^\ast$ & \vci{+0.30}{-0.22}{0.85} & \vci{+0.44}{0.00}{0.93} & \vci{-0.24}{-0.63}{0.16} & $+1.04$ & $-0.16$ \\
    Clean-input interaction$^\dagger$ & \vci{+0.06}{-0.28}{0.38} & \vci{+0.10}{-0.08}{0.30} & \vci{-0.05}{-0.46}{0.34} & $-0.41$ & $+0.62$ \\
    \bottomrule
  \end{tabular}
\end{table}

On the development split, GRPO raises the ten-condition mean of all five paths by 3.13 to 5.93 PDMS. For the original three paths, the seen drop changes by $-0.27$ $[-0.59, 0.07]$, $+0.17$ $[-0.19, 0.54]$ and $-0.06$ $[-0.46, 0.36]$ for DriveBase-8B, the control and \dbplus{}, whereas GRPO from Aug-IL at 2B increases it significantly on navtest, as detailed in Sec.~\ref{sec:rl}. The interaction for the seen drop is $-0.24$ $[-0.63, 0.16]$. The three original GRPO endpoints span at most 0.03 PDMS on the ten-condition mean; \dbplus{} exceeds the control by $+0.02$ $[-0.06, 0.10]$ ($+0.10$ and $-0.05$ per IL seed).

The two clean-input paths show a similar stage pattern: GRPO raises the ten-condition mean from 86.45 to 92.14 for clean-input adaptation and from 86.61 to 92.19 for clean frozen, gains of $+5.69$ $[4.80, 6.76]$ and $+5.59$ $[4.74, 6.63]$. At the GRPO endpoint, \dbplus{} exceeds clean-input adaptation by $+0.077$ $[-0.016, 0.170]$ and clean frozen by $+0.022$ $[-0.073, 0.109]$ on the ten-condition mean of the development split.

\paragraph{Severity Extrapolation.}
We evaluate the 8B paths, ReCogDrive-8B RL and DriveBase under seven further corruptions on navtest, as listed in \cref{tab:attr_severity}: Gaussian blur with radius 2 and 6, brightness scaled by 0.75 and 0.25, motion blur with a 25-pixel kernel, JPEG quality 5 and noise with $\sigma=50/255$. The primary contrast holds. Against the equal-update control, \dbplus{} lowers the mean drop under the two severe seen corruptions, radius 6 and factor 0.25, by 0.66 $[0.48, 0.83]$ PDMS, with 0.72 $[0.45, 0.99]$ and 0.60 $[0.35, 0.86]$ for the two IL seeds. This is four times its reduction of 0.17 at the training severities, and under radius 6 alone its drop is 1.34 against 2.32 PDMS. The additional training alone has no such effect: under every new condition, the drops of the control and DriveBase-8B differ by less than 0.08 PDMS, with intervals that contain zero. Under the three severe unseen corruptions, the difference is $-0.06$ $[-0.21, 0.09]$. Adapting the representation extends the robustness learned from augmentation beyond the training strengths, within the trained corruption families. Relative to DriveBase, \dbplus{} lowers the severe seen and unseen drops from 3.43 and 2.78 to 1.31 and 1.74 PDMS.

\begin{table}[!htb]
  \setlength{\tabcolsep}{4pt}
  \renewcommand{\arraystretch}{1.06}
  \centering
  \caption{\textbf{Severity extrapolation on NAVSIM navtest.} PDMS drop relative to clean inputs; lower is better. Seen: mean over blur radius 6 and brightness 0.25. Unseen: mean over the three severe unseen corruptions. Training strengths are blur radius 4 and brightness 0.5. $^\dagger$Original setting with five denoising steps. ReCogDrive-8B RL: released weights, one endpoint; other rows average two IL seeds. \dbplus{} $-$ control on Seen: $-0.66$ $[-0.83, -0.48]$; on Unseen: $-0.06$ $[-0.21, 0.09]$.}
  \label{tab:attr_severity}
  \small
  \setlength{\tabcolsep}{3.4pt}
  \begin{tabular}{@{}lccccccccc@{}}
    \toprule
    Endpoint & blur 2 & blur 6 & dim 0.75 & dim 0.25 & motion 25 & JPEG 5 & noise 50 & Seen & Unseen \\
    \midrule
    DriveBase$^\dagger$ & \best{0.06} & 5.32 & 0.03 & 1.54 & 2.90 & 3.43 & 2.01 & 3.43 & 2.78 \\
    ReCogDrive-8B RL & 0.25 & 4.41 & 0.12 & 2.09 & 2.67 & 3.39 & 1.53 & 3.25 & 2.53 \\
    DriveBase-8B & 0.11 & 2.29 & 0.07 & 1.58 & 1.43 & 2.66 & 1.35 & 1.93 & 1.82 \\
    Equal-update control & 0.15 & 2.32 & 0.08 & 1.62 & 1.43 & \best{2.59} & 1.37 & 1.97 & 1.80 \\
    \hl \dbplus{} & 0.09 & \best{1.34} & \best{-0.06} & \best{1.29} & \best{1.28} & 2.61 & \best{1.32} & \best{1.31} & \best{1.74} \\
    \bottomrule
  \end{tabular}
\end{table}

\begin{figure}[htbp]
  \centering
  \includegraphics[width=0.75\textwidth]{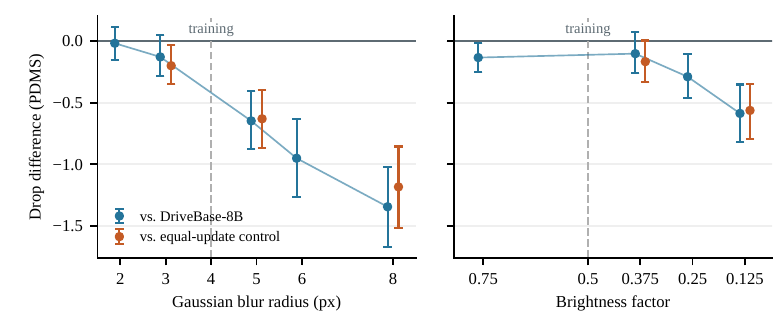}
  \caption{\textbf{The advantage of \dbplus{} grows with corruption strength.} Difference in PDMS drop, \dbplus{} minus reference, under Gaussian blur and dimming on NAVSIM navtest; negative values favour \dbplus{}. Bars: paired 95\% log-bootstrap intervals; two IL seeds, GRPO seed 1. Dashed lines: training strengths, summarized in \cref{tab:attr_navtest_paired}. The equal-update control is shown at the five further strengths.}
  \label{fig:attr_curve}
\end{figure}

\paragraph{Finer Strengths and a Second GRPO Seed.}
Five further strengths of the seen corruptions, blur radius 3, 5 and 8 and brightness factors 0.375 and 0.125, trace the same pattern, as shown in \cref{fig:attr_curve,tab:attr_curve}. Against DriveBase-8B, the drop difference of \dbplus{} is $-0.02$ at radius 2 and $-0.13$ $[-0.29, 0.05]$ at radius 3, and widens to $-0.65$ $[-0.88, -0.41]$ at radius 5, $-0.95$ at radius 6 and $-1.35$ $[-1.67, -1.02]$ at radius 8. For brightness, it is $-0.13$ and $-0.10$ at factors 0.75 and 0.375 and grows to $-0.29$ at 0.25 and $-0.59$ $[-0.82, -0.35]$ at 0.125. Against the equal-update control, the five new differences are close to those against DriveBase-8B, from $-0.17$ $[-0.33, 0.00]$ at factor 0.375 to $-1.18$ $[-1.51, -0.86]$ at radius 8. The advantage under the severe seen corruptions also holds for a second GRPO seed. That seed changes the severe seen drop of \dbplus{} by $+0.06$ $[-0.10, 0.22]$, and with both recipes averaged over two IL and two GRPO seeds, \dbplus{} drops 0.60 $[0.43, 0.76]$ less than DriveBase-8B, with 0.62 $[0.41, 0.82]$ and 0.58 $[0.34, 0.81]$ for the two IL seeds.

\begin{table}[htbp]
  \setlength{\tabcolsep}{4pt}
  \renewcommand{\arraystretch}{1.06}
  \centering
  \caption{\textbf{Further strengths of the seen corruptions on NAVSIM navtest.} Difference in PDMS drop, \dbplus{} minus reference, with paired 95\% log-bootstrap intervals; negative values favour \dbplus{}. Two IL seeds, GRPO seed 1.}
  \label{tab:attr_curve}
  \small
  \begin{tabular}{@{}lcc@{}}
    \toprule
    Strength & vs.\ DriveBase-8B & vs.\ equal-update control \\
    \midrule
    Blur radius 3 & \vci{-0.13}{-0.29}{0.05} & \vci{-0.20}{-0.35}{-0.03} \\
    Blur radius 5 & \vci{-0.65}{-0.88}{-0.41} & \vci{-0.63}{-0.87}{-0.39} \\
    Blur radius 8 & \vci{-1.35}{-1.67}{-1.02} & \vci{-1.18}{-1.51}{-0.86} \\
    Brightness 0.375 & \vci{-0.10}{-0.26}{0.07} & \vci{-0.17}{-0.33}{0.00} \\
    Brightness 0.125 & \vci{-0.59}{-0.82}{-0.35} & \vci{-0.56}{-0.79}{-0.35} \\
    \bottomrule
  \end{tabular}
\end{table}

\paragraph{Adaptation with and without Augmentation.}
A 2$\times$2 design separates adaptation from the augmentation used during it. Two further paths repeat the three additional epochs on clean inputs only. One adapts the VLM as \dbplus{} does, and the other keeps it frozen as the equal-update control does. Both share the IL seeds, the GRPO stage and the evaluation protocol of the other paths. The interaction is the effect of adaptation under augmentation minus its effect on clean inputs, as listed in \cref{tab:f_endpoints,tab:f_contrasts}. At the training strengths, the seen drop shows the predicted interaction of $-0.30$ $[-0.48, -0.11]$. On the ten-condition mean, the interaction of $+0.07$ $[0.00, 0.14]$ has the predicted sign, with an interval that reaches zero. Adaptation lowers the seen drop by 0.17 $[0.00, 0.32]$ under augmentation and raises it by 0.13 $[-0.01, 0.28]$ on clean inputs. For frozen features, the inputs of the additional epochs make no difference: the two frozen paths differ by less than 0.04 PDMS on the ten-condition mean and on every mean drop.

Under the severe seen corruptions, the pattern sharpens. Adaptation on clean inputs raises the drop by 0.33 $[0.13, 0.53]$ over the clean frozen path, adaptation under augmentation lowers it by 0.66 $[0.48, 0.83]$, and the interaction is $-0.99$ $[-1.24, -0.75]$, with $-1.04$ $[-1.37, -0.69]$ and $-0.95$ $[-1.28, -0.64]$ for the two IL seeds. Against clean-input adaptation, \dbplus{} lowers the severe seen drop by 0.96 $[0.71, 1.20]$. Under the severe unseen corruptions, the interaction is $-0.10$ $[-0.29, 0.09]$, with $-0.29$ $[-0.55, -0.01]$ and $+0.09$ $[-0.18, 0.34]$ for the two IL seeds, so the benefit is specific to the trained families. The training-unseen and weather drops change by less than 0.09 PDMS in every contrast. Adaptation on clean inputs also lowers the visual reliance $R$ by 1.15 $[0.76, 1.55]$ relative to the clean frozen path, while \dbplus{} relies on the camera 1.96 $[1.53, 2.39]$ more than the clean-adapted path. All four paths keep the same clean PDMS and EPDMS.

\begin{table}[htbp]
  \setlength{\tabcolsep}{4pt}
  \renewcommand{\arraystretch}{1.06}
  \centering
  \caption{\textbf{Four paths of the adaptation-stage factorial on NAVSIM navtest}, with two IL seeds, GRPO seed 1 and three denoising steps. Rows specify VLM state and inputs for the three additional IL epochs. Ten: clean plus nine corruptions. Severe seen: blur radius 6 and brightness 0.25. Severe unseen: motion blur 25, JPEG 5 and noise 50. $R$: clean minus gray PDMS. $^{\dagger}$Equal-update control. $^{\ddagger}$\dbplus{}.}
  \label{tab:f_endpoints}
  \small
  \setlength{\tabcolsep}{4pt}
  \begin{tabular}{@{}lcccccccc@{}}
    \toprule
    VLM, inputs & Clean$\uparrow$ & Ten$\uparrow$ & Seen drop$\downarrow$ & Unseen drop$\downarrow$ & Severe seen$\downarrow$ & Severe unseen$\downarrow$ & $R$ & EC$\uparrow$ \\
    \midrule
    Frozen, clean & 91.24 & 89.99 & 0.52 & 0.68 & 1.93 & 1.78 & 37.49 & 23.32 \\
    Frozen, augmented$^{\dagger}$ & 91.23 & 89.98 & 0.48 & 0.67 & 1.97 & 1.80 & 37.73 & 23.02 \\
    Adapted, clean & 91.29 & 90.02 & 0.65 & 0.63 & 2.27 & 1.82 & 36.34 & 22.98 \\
    Adapted, augmented$^{\ddagger}$ & 91.27 & 90.08 & 0.31 & 0.60 & 1.31 & 1.74 & 38.30 & 22.75 \\
    \bottomrule
  \end{tabular}
\end{table}

\begin{table}[htbp]
  \setlength{\tabcolsep}{4pt}
  \renewcommand{\arraystretch}{1.06}
  \centering
  \caption{\textbf{Paired contrasts of the adaptation-stage factorial on NAVSIM navtest.} Adaptation contrasts are adapted minus frozen, augmentation contrasts are augmented minus clean, and the interaction is the first row minus the second. Paired 95\% log-bootstrap intervals over two IL seeds with GRPO seed 1; drop differences are negative when the first path degrades less. $^{\ast}$Difference of the family means; no paired interval was computed for this contrast.}
  \label{tab:f_contrasts}
  \small
  \setlength{\tabcolsep}{4pt}
  \begin{tabular}{@{}lccccc@{}}
    \toprule
    Contrast & $\Delta$Ten & $\Delta$Seen drop & $\Delta$Severe seen & $\Delta$Severe unseen & $\Delta R$ \\
    \midrule
    Adaptation under augmentation & \vci{+0.10}{0.02}{0.18} & \vci{-0.17}{-0.32}{0.00} & \vci{-0.66}{-0.83}{-0.48} & \vci{-0.06}{-0.21}{0.09} & \vci{+0.57}{0.18}{0.94} \\
    Adaptation on clean inputs & \vci{+0.03}{-0.04}{0.10} & \vci{+0.13}{-0.01}{0.28} & \vci{+0.33}{0.13}{0.53} & \vci{+0.04}{-0.14}{0.21} & \vci{-1.15}{-1.55}{-0.76} \\
    Augmentation with adaptation & \vci{+0.06}{-0.01}{0.14} & \vci{-0.33}{-0.51}{-0.16} & \vci{-0.96}{-1.20}{-0.71} & \vci{-0.09}{-0.26}{0.09} & \vci{+1.96}{1.53}{2.39} \\
    Augmentation without adaptation & \vci{0.00}{-0.05}{0.04} & \vci{-0.04}{-0.13}{0.06} & $+0.04^{\ast}$ & $+0.01^{\ast}$ & \vci{+0.24}{-0.14}{0.62} \\
    \midrule
    Interaction & \vci{+0.07}{0.00}{0.14} & \vci{-0.30}{-0.48}{-0.11} & \vci{-0.99}{-1.24}{-0.75} & \vci{-0.10}{-0.29}{0.09} & \vci{+1.72}{1.10}{2.34} \\
    \bottomrule
  \end{tabular}
\end{table}

\paragraph{Factorial Replication with GRPO Seed 2.}
With IL seed 1 and GRPO seed 2, adaptation changes the severe-seen drop by $+0.488$\ci{0.245}{0.727} on clean inputs and by $-0.421$\ci{-0.657}{-0.186} under augmentation, so the interaction of \cref{tab:f_contrasts}, augmented minus clean adaptation effects, is $-0.910$\ci{-1.214}{-0.602}. Across the balanced 2$\times$2 IL/GRPO-seed design, the interaction is $-0.953$\ci{-1.174}{-0.728} on the severe-seen drop and $+0.080$\ci{0.016}{0.146} on the ten-condition mean, versus $-0.99$ $[-1.24, -0.75]$ on severe seen for GRPO seed 1 in main-paper Sec.~\ref{sec:sota}.

\paragraph{Broader Augmentation.}
An extension adds motion blur, JPEG quality 10 and noise 25, three corruptions of the evaluation suite, to the augmentation of the three additional IL epochs, once for \dbplus{} and once for the equal-update control on frozen features; both are followed by the same GRPO stage with seed 1, for IL seeds 1 and 2. Over both IL seeds, adaptation under the broader augmentation lowers the drop under the stronger motion blur 25, JPEG quality 5 and noise 50 by 0.46 $[0.28, 0.64]$ relative to the frozen path, the drop under the severe seen corruptions by 0.65 $[0.43, 0.87]$ and the weather drop by 0.23 $[0.10, 0.37]$, although weather is not augmented. It raises the ten-condition mean by 0.24 $[0.16, 0.33]$ at an unchanged clean score, $+0.02$ $[-0.11, 0.15]$; with IL seed 1 alone, the corresponding reductions are 0.47 and 0.27 and the gain is 0.28, as listed in \cref{tab:x4}. On frozen features, the same augmentation gains no robustness under these corruptions: over both IL seeds, its drop there changes by $-0.05$ $[-0.16, 0.06]$ against the equal-update control, and its ten-condition mean by $+0.05$ $[-0.01, 0.10]$. Against \dbplus{} with its original augmentation, the broader version raises the ten-condition mean by 0.19 $[0.11, 0.27]$ over both IL seeds, lowers the drop under the stronger motion blur, JPEG and noise by 0.45 $[0.26, 0.63]$ and the weather drop by 0.31 $[0.15, 0.47]$, and raises extended comfort by 0.62 $[0.22, 1.00]$, all at the same clean score. Before RL, at the end of the three additional IL epochs, adaptation under the broader augmentation already lowers the drop under the stronger corruptions by 0.24 $[0.01, 0.45]$ over both IL seeds.

\begin{table}[htbp]
  \setlength{\tabcolsep}{4pt}
  \renewcommand{\arraystretch}{1.06}
  \centering
  \caption{\textbf{Broader augmentation in the additional IL epochs} on NAVSIM navtest, IL seed 1 and GRPO seed 1. Broader augmentation adds motion blur, JPEG quality 10 and noise 25 to \tname{blur4} and \tname{dim050}. Drops are clean minus corrupted PDMS; Unseen drop: motion blur, JPEG quality 10 and noise 25, outside the original augmentation; Sev. seen: blur radius 6 and brightness 0.25; Sev. unseen: motion blur 25, JPEG quality 5 and noise 50; Weather: fog, rain, snow and overexposure. Paired 95\% log-bootstrap intervals.}
  \label{tab:x4}
  \small
  \setlength{\tabcolsep}{3pt}
  \textbf{(a) Clean, Ten-cond., Seen drop, and Unseen drop}\\[2pt]
  \begin{tabular}{@{}lcccc@{}}
    \toprule
    Contrast & Clean & Ten-cond. & Seen drop & Unseen drop \\
    \midrule
    Adapted $-$ frozen, broader aug. & \vci{0.00}{-0.18}{0.21} & \vci{+0.28}{0.18}{0.38} & \vci{-0.22}{-0.45}{0.03} & \vci{-0.41}{-0.63}{-0.20} \\
    Broader $-$ original aug., frozen & \vci{+0.05}{-0.07}{0.18} & \vci{+0.03}{-0.05}{0.11} & \vci{+0.08}{-0.07}{0.23} & \vci{+0.08}{-0.06}{0.22} \\
    Broader $-$ original aug., \dbplus{} & \vci{+0.06}{-0.10}{0.21} & \vci{+0.16}{0.07}{0.25} & \vci{+0.19}{-0.02}{0.39} & \vci{-0.21}{-0.38}{-0.03} \\
    \bottomrule
  \end{tabular}
  \par
  \vspace{8pt}
  \textbf{(b) Sev. seen, Sev. unseen, Weather, and EC}\\[2pt]
  \begin{tabular}{@{}lcccc@{}}
    \toprule
    Contrast & Sev. seen & Sev. unseen & Weather & EC \\
    \midrule
    Adapted $-$ frozen, broader aug. & \vci{-0.56}{-0.88}{-0.26} & \vci{-0.47}{-0.72}{-0.20} & \vci{-0.27}{-0.47}{-0.07} & \vci{-0.38}{-1.03}{0.25} \\
    Broader $-$ original aug., frozen & \vci{-0.02}{-0.18}{0.13} & \vci{-0.06}{-0.20}{0.09} & \vci{-0.05}{-0.20}{0.11} & \vci{+1.11}{0.51}{1.71} \\
    Broader $-$ original aug., \dbplus{} & \vci{+0.14}{-0.09}{0.37} & \vci{-0.28}{-0.51}{-0.06} & \vci{-0.20}{-0.41}{0.00} & \vci{+0.69}{0.12}{1.25} \\
    \bottomrule
  \end{tabular}
\end{table}

\paragraph{Representation Stability.}
We measure, for every navtest token, how far the planner input features move when the image is corrupted, for the original 8B VLM, the clean-input adaptation and \dbplus{}, under the nine corruptions and twelve further strengths. As listed in \cref{tab:x3}, adaptation under augmentation makes the features move less under the augmented families, by 3.7\% for blur and 5.1\% for dimming averaged over all strengths, while clean-input adaptation reduces the change by 2.4\% for blur and leaves dimming unchanged; in both families, the features of \dbplus{} move less than those of clean-input adaptation. The reduction for blur stays nearly constant across radii, and the one for dimming peaks at factor 0.25. The per-condition intervals agree with these averages: in cosine distance times 100, \dbplus{} moves less than the original VLM at every blur and dimming strength, with all intervals excluding zero, and its advantage over clean-input adaptation grows with strength in both families, from 0.07 $[0.04, 0.10]$ at blur radius 2 to 0.78 $[0.74, 0.82]$ at radius 8 and from 0.15 $[0.14, 0.17]$ at brightness 0.75 to 2.16 $[2.11, 2.22]$ at 0.25.

\begin{table}[htbp]
  \setlength{\tabcolsep}{4pt}
  \renewcommand{\arraystretch}{1.06}
  \centering
  \caption{\textbf{Representation stability under corruption} on NAVSIM navtest: cosine distance between the planner input features of the clean and of the corrupted image, $\times 100$, averaged over 12,146 tokens and over the listed strengths; lower is more stable. Adapted features average two IL seeds; $\Delta$: relative change from the original 8B VLM in \%. Per-condition 95\% intervals are given in the text.}
  \label{tab:x3}
  \small
  \setlength{\tabcolsep}{4pt}
  \begin{tabular}{@{}lccccc@{}}
    \toprule
    & Original & \multicolumn{2}{c}{Clean-input adaptation} & \multicolumn{2}{c}{\dbplus{}} \\
    \cmidrule(lr){3-4}
    \cmidrule(lr){5-6}
    Corruptions & 8B VLM & Distance & $\Delta$ & Distance & $\Delta$ \\
    \midrule
    Blur, radius 2--8 & 37.30 & 36.40 & $-2.4$ & 35.91 & $-3.7$ \\
    Dimming, factor 0.75--0.125 & 27.32 & 27.36 & $+0.1$ & 25.93 & $-5.1$ \\
    Motion blur 15, 25 & 37.61 & 36.70 & $-2.4$ & 36.58 & $-2.7$ \\
    JPEG quality 10, 5 & 45.39 & 43.80 & $-3.5$ & 45.35 & $-0.1$ \\
    Noise 25, 50 & 40.63 & 39.30 & $-3.3$ & 40.39 & $-0.6$ \\
    Fog, rain, snow, overexp. & 33.87 & 33.30 & $-1.7$ & 33.39 & $-1.4$ \\
    \bottomrule
  \end{tabular}
\end{table}

\section{Additional IL and GRPO Seeds}
\label{sec:thirdseed}
The main comparisons keep their sign across IL and GRPO seeds.

We add IL seed 3 for Base-IL and Aug-IL and run the same GRPO on each, which gives three IL seeds for Base-IL + GRPO, Aug-IL and DriveBase. As listed in \cref{tab:thirdseed}, the main comparisons hold. Averaged over the three IL seeds, DriveBase keeps the clean score of Base-IL + GRPO, with a difference of $+0.06$ PDMS and an interval of $[-0.13, 0.24]$. It reduces the seen and unseen drops by 1.52 and 0.72 PDMS, with intervals of $[1.24, 1.83]$ and $[0.47, 0.98]$. RL from Aug-IL again erodes the seen robustness, by 0.47 PDMS with an interval of $[0.14, 0.75]$, and the overall criterion holds for each IL seed and for their mean. The intervals reflect the sampling of driving logs and not the variance across further training runs.

\begin{table}[htbp]
  \setlength{\tabcolsep}{4pt}
  \renewcommand{\arraystretch}{1.06}
  \centering
  \caption{\textbf{Three IL seeds} on NAVSIM navtest. Mean drop over the seen and unseen corruptions, and mean PDMS over the five corruptions.}
  \label{tab:thirdseed}
  \small
  \setlength{\tabcolsep}{4pt}
  \begin{tabular}{@{}llcccc@{}}
    \toprule
    Model & IL seed & Clean$\uparrow$ & Seen drop$\downarrow$ & Unseen drop$\downarrow$ & Mean$\uparrow$ \\
    \midrule
    \multirow{4}{*}{Base-IL + GRPO} & 1 & 91.10 & 2.70 & 1.85 & 88.91 \\
    & 2 & 90.95 & 2.15 & 1.68 & 89.08 \\
    & 3 & 91.28 & 2.32 & 1.94 & 89.18 \\
    & mean & 91.11 & 2.39 & 1.83 & 89.06 \\
    \midrule
    \multirow{4}{*}{Aug-IL} & 1 & 86.92 & 0.50 & 1.00 & 86.12 \\
    & 2 & 86.71 & 0.26 & 0.61 & 86.24 \\
    & 3 & 87.03 & 0.43 & 1.02 & 86.24 \\
    & mean & 86.89 & 0.40 & 0.88 & 86.20 \\
    \midrule
    \multirow{4}{*}{DriveBase} & 1 & 91.18 & 0.97 & 1.04 & 90.17 \\
    & 2 & 90.93 & 0.72 & 1.10 & 89.98 \\
    & 3 & 91.39 & 0.92 & 1.17 & 90.32 \\
    & mean & \best{91.17} & 0.87 & 1.10 & \best{90.16} \\
    \bottomrule
  \end{tabular}
\end{table}

\paragraph{Variation across Seeds.}
The log-bootstrap intervals condition on the trained models and do not contain training variation. Across the three IL seeds of \cref{tab:thirdseed}, the standard deviation of the clean PDMS is 0.16 to 0.23 and that of the five-corruption mean 0.07 to 0.17. Changing only the GRPO seed moves the ten-condition mean on the development split by 0.04, as detailed in \cref{sec:posthoc}, and the evaluation sampling seed moves the navtest mean by at most 0.06 to 0.18, as listed in \cref{tab:inflation}, compared with 0.90 to 1.34 PDMS gained by selecting the best of six samples with ground truth. These magnitudes set the resolution of NAVSIM comparisons, so alongside each interval we report whether a difference holds in every IL seed.

For \dbplus{} and DriveBase-8B, we trained a second GRPO seed for both IL seeds and evaluated all eight runs on navtest with the final protocol, as listed in \cref{tab:grposeed,tab:grposeed_paired}. Scores are averaged per token over the IL seeds and then over the GRPO seeds. For \dbplus{}, the second seed changes the ten-condition mean by $-0.10$ $[-0.20, 0.00]$, by the same amount for both IL seeds, while the clean score, the seen drop and the unseen drop change by $-0.05$ $[-0.19, 0.08]$, $+0.05$ $[-0.09, 0.19]$ and $+0.03$ $[-0.09, 0.16]$. Extended comfort rises by 2.26 $[1.46, 3.08]$ and $R$ by 1.12 $[0.69, 1.58]$. DriveBase-8B moves in the same direction, by $-0.14$ $[-0.23, -0.05]$ on the ten-condition mean and by $+2.29$ $[1.60, 3.04]$ in extended comfort. A second GRPO seed shifts both recipes alike, so their paired difference is stable even though the shift of each recipe is as large as the adaptation gain at the training severities, as detailed in \cref{sec:attribution}.

Over the 2$\times$2 runs, \dbplus{} exceeds DriveBase-8B by 0.13 $[0.05, 0.20]$ on the ten-condition mean and by 0.19 $[0.10, 0.28]$ on the five-corruption mean, and its seen drop is 0.17 $[0.03, 0.29]$ lower. The ten-condition gain is 0.11 $[0.02, 0.19]$ with the first GRPO seed and 0.15 $[0.07, 0.22]$ with the second. Averaged over the GRPO seeds, it is 0.15 $[0.05, 0.25]$ for IL seed 1 and 0.11 $[0.02, 0.20]$ for IL seed 2, as listed in \cref{tab:attr_navtest_paired}. The difference is positive in all four single runs, with intervals above zero in three of them. \dbplus{} keeps the clean score, $+0.07$ $[-0.03, 0.17]$, and relies more on the camera, with $R$ higher by 0.32 $[0.11, 0.54]$. Over the 2$\times$2 runs, \dbplus{} averages 91.24 clean, 90.74 five-corruption and 90.03 ten-condition PDMS, against 91.17, 90.55 and 89.90 for DriveBase-8B. The main table uses GRPO seed 1.

\begin{table}[!htb]
  \setlength{\tabcolsep}{4pt}
  \renewcommand{\arraystretch}{1.06}
  \centering
  \caption{\textbf{Two GRPO seeds of \dbplus{} and DriveBase-8B} on NAVSIM navtest. Mean: five primary corruptions. Ten: clean input and nine corruptions. EC uses the 12{,}104 tokens valid in all eight runs.}
  \label{tab:grposeed}
  \small
  \begin{tabular}{@{}llccccccc@{}}
    \toprule
    IL seed & GRPO seed & Clean$\uparrow$ & Mean$\uparrow$ & Ten$\uparrow$ & Seen drop$\downarrow$ & Unseen drop$\downarrow$ & EPDMS$\uparrow$ & EC$\uparrow$ \\
    \midrule
    \multicolumn{9}{@{}l}{\textit{DriveBase-8B}} \\
    1 & 1 & 91.30 & 90.64 & 90.08 & 0.58 & 0.71 & 81.36 & 22.74 \\
    2 & 1 & 91.18 & 90.60 & 89.87 & 0.46 & 0.66 & 81.22 & 23.50 \\
    1 & 2 & 91.15 & 90.49 & 90.00 & 0.48 & 0.79 & 81.45 & 25.65 \\
    2 & 2 & 91.07 & 90.47 & 89.67 & 0.50 & 0.66 & 81.31 & 25.17 \\
    \multicolumn{2}{@{}l}{2$\times$2 mean} & 91.17 & 90.55 & 89.90 & 0.51 & 0.70 & 81.33 & 24.26 \\
    \midrule
    \multicolumn{9}{@{}l}{\textit{\dbplus{}}} \\
    1 & 1 & 91.30 & 90.84 & 90.23 & 0.23 & 0.61 & 81.22 & 22.47 \\
    2 & 1 & 91.24 & 90.73 & 89.93 & 0.40 & 0.59 & 81.26 & 23.03 \\
    1 & 2 & 91.22 & 90.70 & 90.13 & 0.32 & 0.65 & 81.38 & 24.71 \\
    2 & 2 & 91.22 & 90.69 & 89.83 & 0.41 & 0.61 & 81.47 & 25.31 \\
    \multicolumn{2}{@{}l}{2$\times$2 mean} & 91.24 & 90.74 & 90.03 & 0.34 & 0.61 & 81.33 & 23.88 \\
    \bottomrule
  \end{tabular}
\end{table}

\begin{table}[!htb]
  \setlength{\tabcolsep}{4pt}
  \renewcommand{\arraystretch}{1.06}
  \centering
  \caption{\textbf{Paired GRPO-seed contrasts on NAVSIM navtest.} Left minus right with paired 95\% log-bootstrap intervals on navtest. Scores are averaged per token over the two IL seeds, and over the two GRPO seeds in the 2$\times$2 row; drop differences are negative when the left side degrades less. The GRPO seed 1 row repeats the first contrast of \cref{tab:attr_navtest_paired}.}
  \label{tab:grposeed_paired}
  \small
  \setlength{\tabcolsep}{4pt}
  \begin{tabular}{@{}lccccccc@{}}
    \toprule
    & $\Delta$Clean & $\Delta$Mean & $\Delta$Ten & $\Delta$Seen drop & $\Delta$Unseen drop & $\Delta$EPDMS & $\Delta$EC \\
    \midrule
    \multicolumn{8}{@{}l}{\textit{\dbplus{} $-$ DriveBase-8B}} \\
    GRPO seed 1 & \vci{+0.03}{-0.09}{0.16} & \vci{+0.16}{0.05}{0.27} & \vci{+0.11}{0.02}{0.19} & \vci{-0.21}{-0.36}{-0.04} & \vci{-0.08}{-0.22}{0.05} & \vci{-0.05}{-0.17}{0.08} & \vci{-0.36}{-0.81}{0.10} \\
    GRPO seed 2 & \vci{+0.11}{-0.01}{0.24} & \vci{+0.22}{0.13}{0.31} & \vci{+0.15}{0.07}{0.22} & \vci{-0.13}{-0.28}{0.03} & \vci{-0.09}{-0.23}{0.04} & \vci{+0.04}{-0.09}{0.19} & \vci{-0.40}{-0.81}{0.03} \\
    2$\times$2 & \vci{+0.07}{-0.03}{0.17} & \vci{+0.19}{0.10}{0.28} & \vci{+0.13}{0.05}{0.20} & \vci{-0.17}{-0.29}{-0.03} & \vci{-0.09}{-0.20}{0.02} & \vci{0.00}{-0.10}{0.11} & \vci{-0.38}{-0.70}{-0.04} \\
    \midrule
    \multicolumn{8}{@{}l}{\textit{GRPO seed 2 $-$ GRPO seed 1}} \\
    DriveBase-8B & \vci{-0.13}{-0.27}{0.00} & \vci{-0.15}{-0.25}{-0.05} & \vci{-0.14}{-0.23}{-0.05} & \vci{-0.03}{-0.16}{0.11} & \vci{+0.04}{-0.09}{0.17} & \vci{+0.09}{-0.10}{0.28} & \vci{+2.29}{1.60}{3.04} \\
    \dbplus{} & \vci{-0.05}{-0.19}{0.08} & \vci{-0.09}{-0.20}{0.02} & \vci{-0.10}{-0.20}{0.00} & \vci{+0.05}{-0.09}{0.19} & \vci{+0.03}{-0.09}{0.16} & \vci{+0.18}{0.01}{0.36} & \vci{+2.26}{1.46}{3.08} \\
    \bottomrule
  \end{tabular}
\end{table}

\paragraph{Equal-update Control across GRPO Seeds.}
\Cref{tab:equal_update_grposeed} repeats main-paper Tab.~\ref{tab:adaptation-seeds}. Averaging IL seeds 1 and 2 per token, then both GRPO seeds, \dbplus{} gains $+0.157$\ci{0.088}{0.229} in ten-condition mean and reduces severe-seen drop by $0.600$\ci{0.438}{0.759}; both contrasts exclude zero for each GRPO seed. Main-paper Tab.~\ref{tab:main} uses GRPO seed 1.

\begin{table}[!htb]
  \setlength{\tabcolsep}{4pt}
  \renewcommand{\arraystretch}{1.06}
  \centering
  \caption{\textbf{\dbplus{} minus equal-update control, 2$\times$2 IL/GRPO seeds.} Ten: clean plus nine corruptions. Severe seen: clean minus mean PDMS at blur radius 6 and brightness 0.25. Paired 95\% log-bootstrap intervals; negative drop contrasts favour \dbplus{}.}
  \label{tab:equal_update_grposeed}
  \small
  \begin{tabular}{@{}lcc@{}}
    \toprule
    GRPO seeds & $\Delta$Ten-condition & $\Delta$Severe-seen drop \\
    \midrule
    1 & \vci{+0.101}{0.020}{0.179} & \vci{-0.660}{-0.831}{-0.479} \\
    2 & \vci{+0.214}{0.132}{0.298} & \vci{-0.539}{-0.729}{-0.352} \\
    1 and 2, pooled & \vci{+0.157}{0.088}{0.229} & \vci{-0.600}{-0.759}{-0.438} \\
    \bottomrule
  \end{tabular}
\end{table}

\paragraph{Controls with GRPO Seed 2.}
\Cref{tab:grposeed_controls} reports the control endpoints with GRPO seed 2, including clean-input adaptation and its frozen counterpart for both IL seeds. Their five-corruption means range from 90.39 to 90.59, below the 90.69 to 90.84 of the four \dbplus{} runs in \cref{tab:grposeed}. With IL seed 2, clean-input adaptation raises the severe-seen drop over its frozen counterpart by $+0.258$\ci{-0.003}{0.518}, while it adds $+0.229$\ci{0.112}{0.350} to the ten-condition mean, so clean-input adaptation costs robustness only under strong corruptions of the augmented families.

\begin{table}[htbp]
  \setlength{\tabcolsep}{4pt}
  \renewcommand{\arraystretch}{1.06}
  \centering
  \caption{\textbf{Control endpoints with GRPO seed 2} on NAVSIM navtest. Mean: five primary corruptions. Ten: clean input and nine corruptions. Severe drops use the conditions of \cref{tab:f_endpoints}. Point estimates over 12{,}146 tokens.}
  \label{tab:grposeed_controls}
  \small
  \setlength{\tabcolsep}{4pt}
  \begin{tabular}{@{}llccccc@{}}
    \toprule
    Endpoint & IL seed & Clean$\uparrow$ & Mean$\uparrow$ & Ten$\uparrow$ & Severe seen$\downarrow$ & Severe unseen$\downarrow$ \\
    \midrule
    Equal-update control & 1 & 91.08 & 90.47 & 89.96 & 1.86 & 1.92 \\
    Equal-update control & 2 & 91.06 & 90.41 & 89.58 & 1.97 & 1.93 \\
    Clean frozen & 1 & 91.19 & 90.47 & 89.95 & 2.00 & 2.09 \\
    Clean frozen & 2 & 91.06 & 90.39 & 89.58 & 2.10 & 1.97 \\
    Clean-input adaptation & 1 & 91.20 & 90.46 & 89.97 & 2.49 & 2.20 \\
    Clean-input adaptation & 2 & 91.29 & 90.59 & 89.81 & 2.36 & 1.99 \\
    \bottomrule
  \end{tabular}
\end{table}

\section{Severity Sweep}
\label{sec:severity}
Stronger corruptions widen the lead of \dbplus{} over other model families, as Sec.~\ref{sec:sota} states.

\begin{figure}[htbp]
  \centering
  \begin{tikzpicture}
    \begin{axis}[width=0.47\textwidth, height=5.0cm, xlabel={Gaussian blur radius (px)}, ylabel={PDMS drop}, xtick={2,4,6}, grid=both, grid style={line width=.1pt, draw=gray!25}, tick label style={font=\scriptsize}, label style={font=\small}, legend style={font=\scriptsize, at={(0.03,0.97)}, anchor=north west}]
    \addplot+[blue, mark=o, thick, error bars/.cd, y dir=both, y explicit] coordinates {(2,0.721) +- (0,0.333) (4,4.854) +- (0,0.787) (6,12.375) +- (0,1.206)};
    \addlegendentry{Base-IL + GRPO}
    \addplot+[blue, mark=*, thick, error bars/.cd, y dir=both, y explicit] coordinates {(2,-0.088) +- (0,0.251) (4,0.434) +- (0,0.349) (6,2.958) +- (0,0.634)};
    \addlegendentry{Base-IL + RA-GRPO}
    \addplot+[red, mark=triangle, thick, error bars/.cd, y dir=both, y explicit] coordinates {(2,0.089) +- (0,0.281) (4,1.604) +- (0,0.495) (6,5.454) +- (0,0.876)};
    \addlegendentry{Aug-IL + GRPO}
    \addplot+[red, mark=triangle*, thick, error bars/.cd, y dir=both, y explicit] coordinates {(2,0.039) +- (0,0.186) (4,0.304) +- (0,0.314) (6,2.305) +- (0,0.626)};
    \addlegendentry{Aug-IL + RA-GRPO}
    \addplot+[gray, mark=square*, thick, error bars/.cd, y dir=both, y explicit] coordinates {(2,0.353) +- (0,0.319) (4,3.365) +- (0,0.674) (6,10.043) +- (0,1.168)};
    \addlegendentry{ReCogDrive-2B (RL)}
    \end{axis}
  \end{tikzpicture}
  \hfill
  \begin{tikzpicture}
    \begin{axis}[x dir=reverse, width=0.47\textwidth, height=5.0cm, xlabel={Brightness factor}, ylabel={PDMS drop}, xtick={0.25,0.5,0.75}, grid=both, grid style={line width=.1pt, draw=gray!25}, tick label style={font=\scriptsize}, label style={font=\small}]
    \addplot+[blue, mark=o, thick, error bars/.cd, y dir=both, y explicit] coordinates {(0.75,0.076) +- (0,0.153) (0.5,0.539) +- (0,0.281) (0.25,1.991) +- (0,0.557)};
    \addplot+[blue, mark=*, thick, error bars/.cd, y dir=both, y explicit] coordinates {(0.75,-0.019) +- (0,0.155) (0.5,0.131) +- (0,0.224) (0.25,1.169) +- (0,0.390)};
    \addplot+[red, mark=triangle, thick, error bars/.cd, y dir=both, y explicit] coordinates {(0.75,0.101) +- (0,0.151) (0.5,0.327) +- (0,0.224) (0.25,1.564) +- (0,0.516)};
    \addplot+[red, mark=triangle*, thick, error bars/.cd, y dir=both, y explicit] coordinates {(0.75,0.131) +- (0,0.146) (0.5,0.191) +- (0,0.224) (0.25,1.174) +- (0,0.426)};
    \addplot+[gray, mark=square*, thick, error bars/.cd, y dir=both, y explicit] coordinates {(0.75,0.096) +- (0,0.183) (0.5,0.525) +- (0,0.307) (0.25,2.236) +- (0,0.651)};
    \end{axis}
  \end{tikzpicture}
  \caption{\textbf{Severity sweep on NAVSIM navtest} for IL seed 1. Aug-IL was trained at radius 4 and factor 0.5 only. Error bars: 95\% log-clustered intervals, symmetrized to the larger side.}
  \label{fig:severity}
\end{figure}
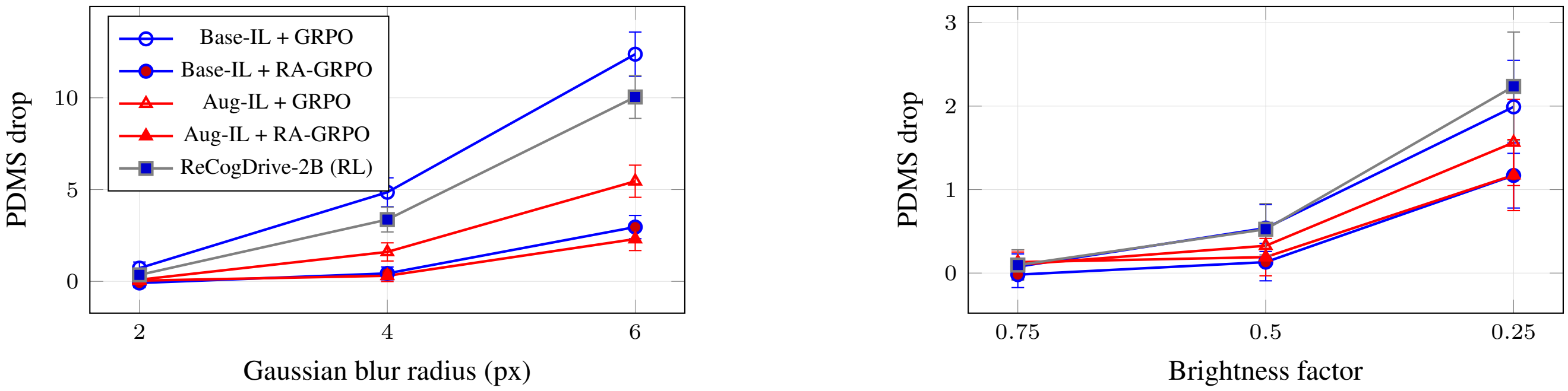

\Cref{fig:severity} extends the two seen corruptions to three severities each: Gaussian blur with radius 2, 4 and 6 pixels, and brightness scaled by 0.75, 0.5 and 0.25. Degradation grows with severity for every model, and the ordering of the models is largely stable. The three models that saw corrupted inputs during training, Aug-IL + GRPO, Base-IL + RA-GRPO and Aug-IL + RA-GRPO, degrade least at all other severities. There are two exceptions. Under the mildest dimming, factor 0.75, Base-IL + GRPO loses 0.08 PDMS against 0.10 for Aug-IL + GRPO and 0.13 for Aug-IL + RA-GRPO. Under the strongest dimming, the released ReCogDrive-2B loses 2.24 PDMS against 1.99 for Base-IL + GRPO. Aug-IL was trained only at radius 4 and factor 0.5. Yet at radius 6, Aug-IL + GRPO loses 5.45 PDMS, against 12.38 for Base-IL + GRPO and 10.04 for the released ReCogDrive-2B, and at factor 0.25 it loses 1.56, against 1.99 and 2.24. The robustness from augmentation extends beyond the training strengths. \Cref{sec:attribution} evaluates the 8B paths at these and further strengths, with DriveBase averaged over two IL seeds.

\paragraph{Severe Corruptions across Methods.}
\Cref{tab:x1} evaluates WAM-Flow, OneVL and ReCogDrive-2B RL under the five severe corruptions of \cref{sec:attribution}, each with its released weights. On clean inputs, \dbplus{} leads them by 0.56 to 2.34 PDMS. Under blur radius 6 and brightness 0.25, its lead grows to 4.30 to 5.39 PDMS, and under the three severe unseen corruptions to 2.79 to 3.44 PDMS; all these intervals lie above zero for both IL seeds. The three methods also change order. ReCogDrive-2B RL ranks first of the three on clean inputs and last under blur radius 6, whereas WAM-Flow ranks last under brightness 0.25, so the clean ranking does not carry over to strong corruptions.

\begin{table}[htbp]
  \setlength{\tabcolsep}{4pt}
  \renewcommand{\arraystretch}{1.06}
  \centering
  \caption{\textbf{Released methods under severe corruptions on NAVSIM navtest.} PDMS of each method with its released weights and the inference settings of Tab.~\ref{tab:main}. Severe seen: blur radius 6 and brightness 0.25. Severe unseen: motion blur 25, JPEG quality 5 and noise 50. $\Delta$: \dbplus{} minus the method, with paired 95\% log-bootstrap intervals; \dbplus{} averages two IL seeds.}
  \label{tab:x1}
  \small
  \setlength{\tabcolsep}{4pt}
  \begin{tabular}{@{}lccccccccc@{}}
    \toprule
    Method & Clean & Blur 6 & Dim 0.25 & Motion 25 & JPEG 5 & Noise 50 & $\Delta$Severe seen & $\Delta$Severe unseen \\
    \midrule
    WAM-Flow & 90.35 & 86.77 & 82.85 & 86.56 & 85.97 & 87.71 & \vci{+5.15}{4.43}{5.87} & \vci{+2.79}{2.10}{3.50} \\
    OneVL & 88.93 & 84.98 & 86.33 & 87.55 & 85.92 & 84.82 & \vci{+4.30}{3.58}{4.98} & \vci{+3.44}{2.74}{4.08} \\
    ReCogDrive-2B RL & 90.71 & 80.66 & 88.47 & 84.94 & 85.78 & 87.76 & \vci{+5.39}{4.65}{6.16} & \vci{+3.37}{2.74}{3.99} \\
    \midrule
    \dbplus{} & 91.27 & 89.93 & 89.98 & 89.99 & 88.66 & 89.95 & -- & -- \\
    \bottomrule
  \end{tabular}
\end{table}

\paragraph{Fewer Zero Scores.}
The higher severe-corruption PDMS is accompanied by fewer zero scores. In \cref{tab:x2_zero}, \dbplus{} reduces the PDMS-zero rate relative to ReCogDrive-8B RL by 3.11 percentage points under blur radius 6 and by 1.05 points under brightness 0.25; both paired intervals exclude zero. Zero PDMS flags composite-score failure, not its cause.

\begin{table}[htbp]
  \setlength{\tabcolsep}{4pt}
  \renewcommand{\arraystretch}{1.06}
  \centering
  \caption{\textbf{Severe-seen PDMS-zero rates} on 12{,}146 NAVSIM navtest tokens. Rates in \%; $\Delta$ in percentage points: \dbplus{} minus ReCogDrive-8B RL. 95\% log-bootstrap intervals; \dbplus{} averages IL-seed zero-score indicators per token, then aggregates.}
  \label{tab:x2_zero}
  \small
  \begin{tabular}{@{}lccc@{}}
    \toprule
    Condition & ReCogDrive-8B RL (\%) & \dbplus{} (\%) & $\Delta$ (pp) \\
    \midrule
    Blur radius 6 & \vci{7.80}{6.67}{8.94} & \vci{4.69}{3.94}{5.49} & \vci{-3.11}{-3.85}{-2.39} \\
    Brightness 0.25 & \vci{5.67}{4.72}{6.69} & \vci{4.63}{3.92}{5.35} & \vci{-1.05}{-1.51}{-0.58} \\
    \bottomrule
  \end{tabular}
\end{table}

\section{Qualitative Results}
\label{sec:qual}
The examples follow fixed selection rules and show \dbplus{} keeping collision-free plans where ReCogDrive-8B RL collides.

\begin{figure}[!htb]
  \centering
  \includegraphics[width=\textwidth]{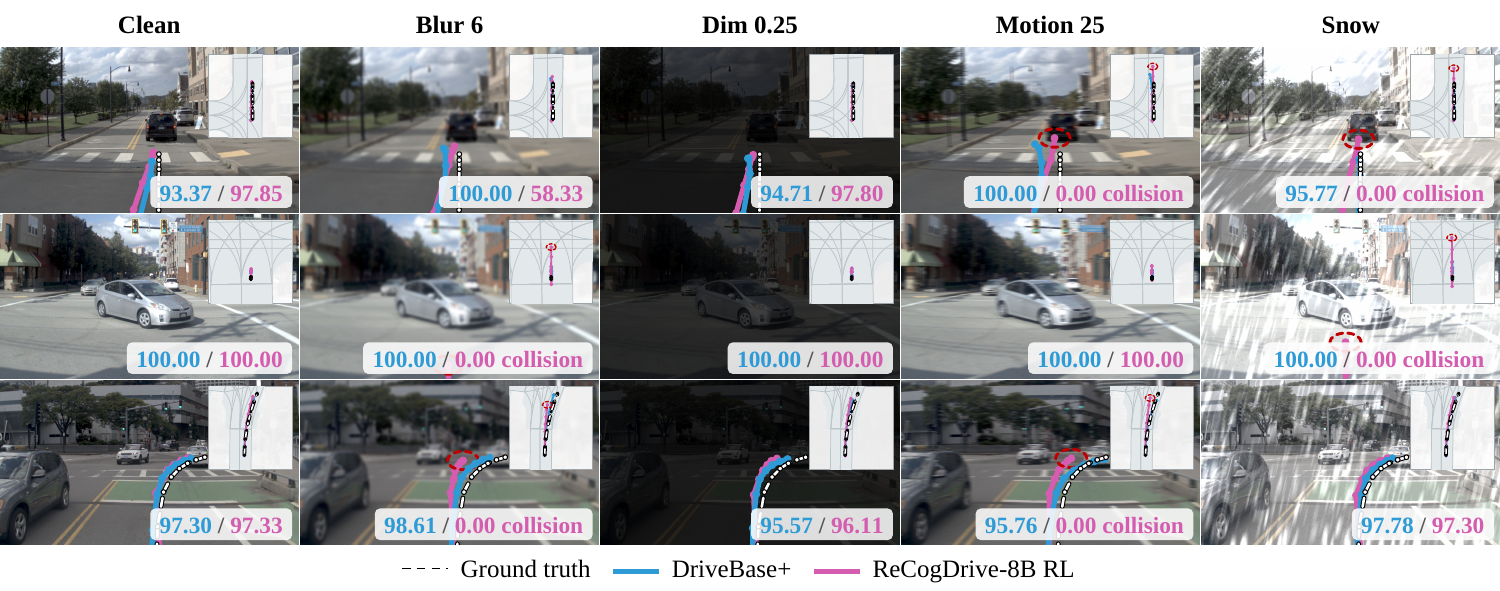}
  \caption{\textbf{\dbplus{} and ReCogDrive-8B RL across corruptions.} Plans projected onto the front camera, with a bird's-eye inset; scores are the PDMS of \dbplus{} and ReCogDrive-8B RL, each with its Tab.~\ref{tab:main} inference settings; labels name the term that zeros the score.}
  \label{fig:q1}
\end{figure}

\paragraph{Robust Planning across Corruptions.}
\Cref{fig:q1} compares \dbplus{} with ReCogDrive-8B RL on three scenes chosen by a fixed rule: both IL seeds of \dbplus{} score at least 90 on the clean input and under the four corruptions, while ReCogDrive-8B RL scores at least 90 on the clean input and at most 50 under at least two corruptions, with a plan at least 2\,m away wherever it fails. The two scenes of Fig.~\ref{fig:m1} come from this set. In all six failures shown, ReCogDrive-8B RL plans a collision, while \dbplus{} keeps a collision-free plan within the drivable area.

\begin{figure}[htbp]
  \centering
  \includegraphics[width=\textwidth]{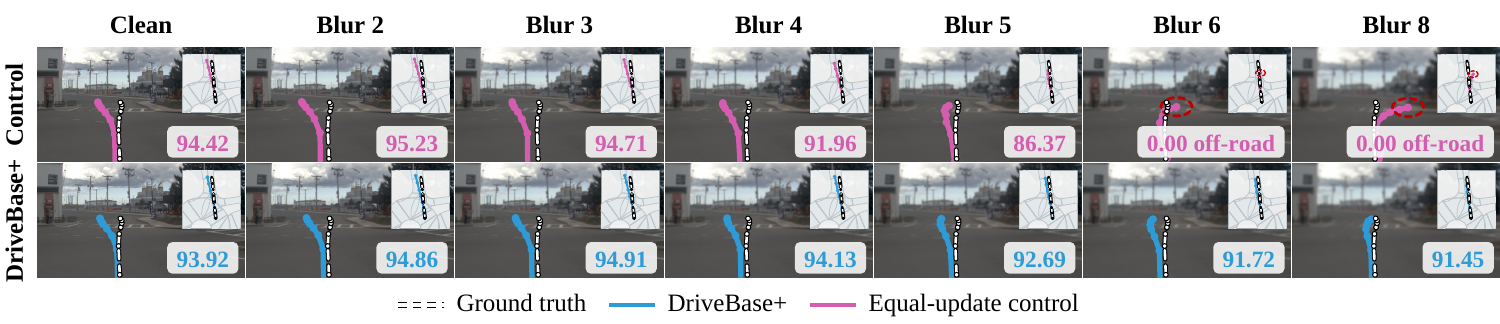}
  \caption{\textbf{One scene across blur strengths.} Top: the equal-update control; bottom: \dbplus{}. The control fails from radius 6 on, where \dbplus{} keeps following the lane.}
  \label{fig:q2}
\end{figure}

\paragraph{Severity.}
\Cref{fig:q2} follows one scene across blur strengths. The equal-update control declines from 94.42 PDMS on the clean input to 86.37 at radius 5 and leaves the drivable area at radius 6 and 8, while \dbplus{} stays between 91.45 and 94.91. The scene follows a fixed rule: both models score at least 90 on the clean input and at radius 2, \dbplus{} at least 90 at every strength, and the control at most 50 at radius 6 and 8.

\begin{figure}[htbp]
  \centering
  \includegraphics[width=0.86\textwidth]{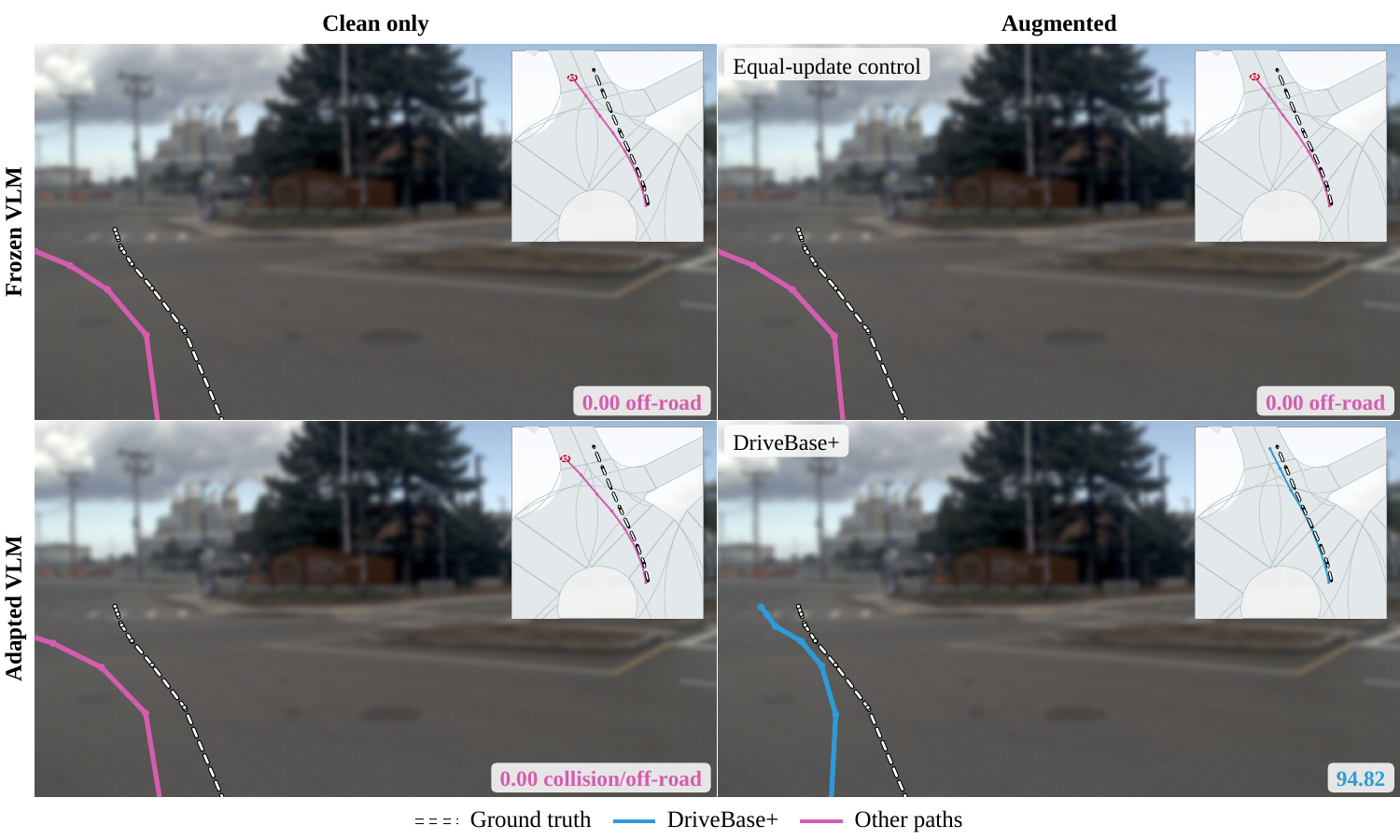}
  \caption{\textbf{The four paths of the adaptation 2$\times$2 under blur radius 6.} Rows: VLM frozen or adapted in the three additional IL epochs; columns: clean or augmented inputs in those epochs.}
  \label{fig:q3}
\end{figure}

\paragraph{Adaptation and Augmentation.}
\Cref{fig:q3} shows the four paths of Fig.~\ref{fig:factorial} on one scene under blur radius 6. On this scene, both IL seeds of \dbplus{} score at least 90 and those of the three other paths at most 50, with all three plans at least 2\,m from that of \dbplus{}. All three cut the corner of the turn and leave the drivable area, while \dbplus{} follows the turn with 94.82 PDMS.

\section{Transfer of the Augmentation Recipe}
\label{sec:transferapp}
The augmentation transfers to two other planners and benchmarks, as Sec.~\ref{sec:transfer} states.

\paragraph{AutoVLA on NAVSIM.}
We train AutoVLA in \fw{} with supervised training without chain-of-thought for 2{,}690 updates, with and without the \tname{blur4} and \tname{dim050} augmentation of Aug-IL, for two training seeds. All models start from the released base model, and the lower clean score than that of the released AutoVLA reflects the absence of reasoning and RL. \Cref{tab:transfer_autovla} averages the two seeds per token. Augmentation leaves the clean score unchanged, with a difference of $+0.13$ PDMS and an interval of $[-0.46, 0.70]$. It lowers the seen drop by 1.31 PDMS and the unseen drop by 0.37 PDMS, both with intervals that exclude zero.

\begin{table}[htbp]
  \setlength{\tabcolsep}{4pt}
  \renewcommand{\arraystretch}{1.06}
  \centering
  \caption{\textbf{Augmentation for AutoVLA} on NAVSIM navtest, scores averaged over two training seeds; $\Delta$: with minus without augmentation.}
  \label{tab:transfer_autovla}
  \small
  \setlength{\tabcolsep}{2.6pt}
  \begin{tabular}{@{}lcccccccc@{}}
    \toprule
    Model & Clean & \tname{blur4} & \tname{dim050} & motion & JPEG & noise & Seen drop & Unseen drop \\
    \midrule
    SFT & 76.65 & 75.44 & 73.80 & 76.06 & 74.95 & 76.38 & 2.03 & 0.85 \\
    SFT + aug. & \best{76.79} & \best{76.12} & \best{76.01} & \best{76.52} & \best{75.69} & \best{76.69} & \best{0.72} & \best{0.49} \\
    \midrule
    $\Delta$ aug. & $+0.13$ & $+0.68$ & $+2.21$ & $+0.46$ & $+0.74$ & $+0.31$ & $-1.31$ & $-0.37$ \\
    \bottomrule
  \end{tabular}
\end{table}

\paragraph{OpenDriveVLA on nuScenes.}
We fine-tune OpenDriveVLA-0.5B from its released weights for one epoch with two seeds, with and without the augmentation, and evaluate the average L2 error under the ST-P3 protocol on all 6{,}019 validation samples. \Cref{tab:transfer_nusc} averages the two seeds per sample and resamples the 150 scenes. Augmentation lowers the increase of L2 under the seen corruptions by 0.53\,cm, with an interval of $[0.17, 0.85]$, mostly under \tname{blur4}, without a significant change on clean inputs. On real night scenes of the validation set, the augmented model has a lower L2 for both seeds.

\begin{table}[htbp]
  \setlength{\tabcolsep}{4pt}
  \renewcommand{\arraystretch}{1.06}
  \centering
  \caption{\textbf{OpenDriveVLA augmentation} on nuScenes; mean L2 (cm), lower is better; $\Delta$: fine-tuned with minus without augmentation.}
  \label{tab:transfer_nusc}
  \small
  \setlength{\tabcolsep}{2.6pt}
  \begin{tabular}{@{}lccccccc@{}}
    \toprule
    Model & Clean & \tname{blur4} & \tname{dim050} & motion & JPEG & noise & Seen incr. \\
    \midrule
    Released weights & 34.80 & 36.40 & 35.40 & 35.70 & 35.70 & 35.70 & 1.10 \\
    Fine-tuned & 33.30 & 34.90 & 33.70 & 33.70 & 33.70 & 33.90 & 1.00 \\
    Fine-tuned + aug. & \best{33.20} & \best{34.00} & \best{33.40} & \best{33.40} & \best{33.50} & \best{33.60} & \best{0.50} \\
    \midrule
    $\Delta$ aug. & $-0.10$ & $-0.90$ & $-0.30$ & $-0.30$ & $-0.20$ & $-0.30$ & $-0.53$ \\
    \bottomrule
  \end{tabular}
\end{table}

\FloatBarrier
\supppart{Part V. Action-Head Analyses}
\section{Sensitivity of the Action-Head Comparison}
\label{sec:headsens}
The ranking of action heads in Tab.~\ref{tab:ablation} holds when the alternative heads are retuned.

\paragraph{Tuning of the Alternative Heads.}
We retrain the regression head and the discrete vocabulary of Tab.~\ref{tab:ablation} with IL seed 1 under four further settings on the same cached features and schedule for 128 epochs: the MLP head with a learning rate of $3\times10^{-4}$, and vocabularies of 1{,}024, 4{,}096 and 8{,}192 actions with learning rates of $10^{-4}$ or $3\times10^{-4}$. As \cref{tab:headsens} lists, no setting closes the gap. Against Base-IL with the same seed, clean PDMS stays 6.79 to 15.38 lower and the drop under \tname{blur4} 7.53 to 23.22 higher, all with intervals that exclude zero. The largest gain over the original head, $+2.29$ clean PDMS for the 1{,}024-action vocabulary, still leaves it 11.48 behind Base-IL. The ranking of the action heads does not come from their tuning. Intervals reflect evaluation uncertainty for one training seed.

\begin{table}[htbp]
  \setlength{\tabcolsep}{4pt}
  \renewcommand{\arraystretch}{1.06}
  \centering
  \caption{\textbf{Retuned alternative action heads} on NAVSIM navtest, IL seed 1. Drop: clean minus \tname{blur4} PDMS. $\Delta$: paired change with 95\% interval against Base-IL with the same seed, which reaches 86.96 clean PDMS and a drop of 4.42, or against the original head of Tab.~\ref{tab:ablation}.}
  \label{tab:headsens}
  \small
  \setlength{\tabcolsep}{3.2pt}
  \begin{tabular}{@{}lccccccc@{}}
    \toprule
    & & & & \multicolumn{2}{c}{Against Base-IL} & Against original head \\
    \cmidrule(lr){5-6}
    \cmidrule(lr){7-7}
    Setting & Clean & \tname{blur4} & Drop & $\Delta$Clean & $\Delta$Drop & $\Delta$Clean \\
    \midrule
    MLP, lr $3\times10^{-4}$ & 80.17 & 68.22 & 11.95 & \vci{-6.79}{-8.02}{-5.55} & \vci{+7.53}{5.58}{9.40} & \vci{-0.55}{-1.22}{0.17} \\
    Vocabulary 1{,}024, lr $10^{-4}$ & 75.48 & 61.54 & 13.94 & \vci{-11.48}{-12.79}{-10.13} & \vci{+9.52}{8.11}{10.90} & \vci{+2.29}{1.38}{3.23} \\
    Vocabulary 4{,}096, lr $3\times10^{-4}$ & 72.34 & 49.87 & 22.47 & \vci{-14.62}{-16.06}{-13.17} & \vci{+18.05}{15.45}{20.70} & \vci{-0.85}{-1.60}{-0.13} \\
    Vocabulary 8{,}192, lr $10^{-4}$ & 71.58 & 43.95 & 27.64 & \vci{-15.38}{-16.99}{-13.66} & \vci{+23.22}{20.39}{25.86} & \vci{-1.61}{-2.48}{-0.71} \\
    \bottomrule
  \end{tabular}
\end{table}

\paragraph{Visual Context of the Planner.}
Replacing the sequence of VLM hidden states by its average at inference lowers Base-IL to between 64.18 and 64.26 PDMS under every condition we test, clean inputs, \tname{blur4}, blur radius 6, brightness 0.25, motion blur 25, JPEG quality 5 and noise 50, a loss of 22.77 $[21.34, 24.14]$ PDMS on clean inputs. The planner draws its visual information from the token sequence, and the corruptions act through it.

\section{Full-Precision Values of the Ablation}
\label{sec:table3fp}
Tab.~\ref{tab:ablation} of the main paper rounds scores to two decimals and computes every change before rounding, so a printed change can differ by 0.01 from the difference of two printed scores. \Cref{tab:table3fp} lists the same results with three decimals together with the 95\% intervals behind its significance marks, from the log-level bootstrap with 5{,}000 resamples and seed 20271003. Removing the navigation command is the closest case, with an interval that ends at 0.001.

\begin{table}[htbp]
  \setlength{\tabcolsep}{4pt}
  \renewcommand{\arraystretch}{1.06}
  \centering
  \caption{\textbf{Full-precision values of Tab.~\ref{tab:ablation}}, means over three IL seeds on NAVSIM navtest. $\Delta$Clean: paired change from Base-IL. Drop red.: reduction of the clean-to-\tname{blur4} drop.}
  \label{tab:table3fp}
  \small
  \setlength{\tabcolsep}{4pt}
  \begin{tabular}{@{}lccccc@{}}
    \toprule
    Variant & Clean & $\Delta$Clean & 95\% interval & Drop red. & 95\% interval \\
    \midrule
    Base-IL: diffusion, 5 steps & 86.827 & -- & -- & -- & -- \\
    flow matching & 86.524 & $-0.303$ & $[-0.553, -0.063]$ & $-0.134$ & $[-0.438, 0.170]$ \\
    MLP regression & 80.942 & $-5.885$ & $[-6.935, -4.848]$ & $-7.719$ & $[-9.269, -6.125]$ \\
    discrete vocabulary & 73.361 & $-13.466$ & $[-14.866, -12.013]$ & $-23.282$ & $[-25.387, -21.117]$ \\
    no navigation command & 86.623 & $-0.204$ & $[-0.426, 0.001]$ & $+0.200$ & $[-0.084, 0.482]$ \\
    no ego state & 86.198 & $-0.629$ & $[-0.913, -0.362]$ & $-0.609$ & $[-0.967, -0.246]$ \\
    3 denoising steps & 86.834 & $+0.007$ & $[-0.123, 0.146]$ & $+0.049$ & $[-0.143, 0.229]$ \\
    \bottomrule
  \end{tabular}
\end{table}

\FloatBarrier

\end{document}